%% file: main.tex
\documentclass{article}

\usepackage{paper_style}
\usepackage[utf8]{inputenc}
\usepackage[T1]{fontenc}
\usepackage{xcolor}
\usepackage{url}
\usepackage{booktabs}
\usepackage{amsfonts}
\usepackage{microtype}
\usepackage{graphicx}
\usepackage[numbers,sort&compress]{natbib}
\usepackage{amsmath}
\usepackage{amssymb}
\usepackage{amsthm}
\usepackage{bm}
\usepackage{xspace}
\usepackage{enumitem}
\usepackage{tabularx}
\usepackage{multirow}
\usepackage{makecell}
\usepackage{longtable}
\usepackage{colortbl}
\usepackage{caption}
\usepackage{float}
\usepackage{subcaption}
\usepackage{algorithm}
\usepackage[noend]{algpseudocode}
\usepackage{tikz}
\usetikzlibrary{arrows.meta,positioning,calc,fit,backgrounds,shapes.geometric,shapes.misc,decorations.pathreplacing,shadows}
\usepackage{pgfplots}
\pgfplotsset{compat=1.18}
\usepackage{hyperref}

\definecolor{careink}{HTML}{1F2A37}
\definecolor{careteal}{HTML}{00616B}
\definecolor{careblue}{HTML}{2F6DB5}
\definecolor{careorange}{HTML}{D9822B}
\definecolor{caregreen}{HTML}{3C8D5A}
\definecolor{carered}{HTML}{C0392B}
\definecolor{carepurple}{HTML}{6B4FA0}
\definecolor{caregray}{HTML}{6B7280}
\definecolor{carelight}{HTML}{F3F5F7}
\colorlet{midnightgreen}{careteal}

\hypersetup{
    colorlinks=true,
    linkcolor=careteal,
    urlcolor=careteal,
    citecolor=careteal,
    pdftitle={CARE: Grounded Recursive Self-Improvement for Production Customer-Service Agents},
    pdfauthor={Yanbei Jiang, Chao Lei, Rena Gao, Xiaochen Zhao, Yuchen Ge,
      Haohang Liu, Runze Nie, Qitian Chen, Qi An, Haolin Song, Yufei Sun,
      Yihao Ding, Jey Han Lau, Krista A. Ehinger, Xingjun Ma, Weijia Chen, Ruiyuan Wu,
      Zhen Luo, Ke Xing, Yongjun Yin},
}

\setlist[itemize]{leftmargin=*}
\setlist[enumerate]{leftmargin=*}
\setlist[description]{leftmargin=*}
\newcommand{\care}{CARE\xspace}
\newcommand{\caremodel}{CARE-31B\xspace}
\newcommand{\pistar}{\pi^{\star}}
\newcommand{\pith}{\pi_{\theta}}
\newcommand{\ctx}{c}
\newcommand{\Ctx}{\mathcal{C}}

\newcommand{\sg}{\operatorname{sg}}
\newcommand{\E}{\mathbb{E}}
\theoremstyle{definition}
\newtheorem{definition}{Definition}

\newif\ifdraft
\drafttrue

\newcolumntype{Y}{>{\raggedright\arraybackslash}X}
\newcolumntype{C}{>{\centering\arraybackslash}X}

\title{CARE: Grounded Recursive Self-Improvement\\
for Production Customer-Service Agents}
\newcommand{\careauthor}[2]{\mbox{#1\textsuperscript{#2}}}
\author{%
  \parbox[t]{\dimexpr\textwidth-2\tabcolsep\relax}{\centering\normalfont
    \careauthor{Yanbei Jiang}{1,2},
    \careauthor{Chao Lei}{2},
    \careauthor{Rena Gao}{2},
    \careauthor{Xiaochen Zhao}{1},
    \careauthor{Yuchen Ge}{1},
    \careauthor{Haohang Liu}{1,3},
    \careauthor{Runze Nie}{1},
    \careauthor{Qitian Chen}{1},
    \careauthor{Qi An}{1},
    \careauthor{Haolin Song}{1},
    \careauthor{Yufei Sun}{1},
    \careauthor{Yihao Ding}{2},
    \careauthor{Jey Han Lau}{2},
    \careauthor{Krista A. Ehinger}{2},
    \careauthor{Xingjun Ma}{3},
    \careauthor{Weijia Chen}{1},
    \careauthor{Ruiyuan Wu}{1},
    \careauthor{Zhen Luo}{1},
    \careauthor{Ke Xing}{1},
    \careauthor{Yongjun Yin}{1}\\[6pt]
    {\small \textsuperscript{1}AI and Data Team, Keeta, Meituan\qquad
    \textsuperscript{2}University of Melbourne\qquad
    \textsuperscript{3}Fudan University}%
  }%
}
\date{}

\renewcommand{\headeright}{Technical Report}
\begin{document}

\maketitle
\vspace{-6pt}
\begin{center}
\small Technical Report $\cdot$ October 2026
\end{center}

\begin{abstract}
\input{Technical_Report/sections/00_abstract}
\end{abstract}

\clearpage
\begingroup
\small
\setlength{\parskip}{0pt}
\tableofcontents
\endgroup
\clearpage

\input{Technical_Report/sections/01_introduction}

\input{Technical_Report/sections/02_related_work}
\input{Technical_Report/sections/03_problem}
\input{Technical_Report/sections/04_cold_start}
\input{Technical_Report/sections/05_retrospection}
\input{Technical_Report/sections/06_evolve}
\input{Technical_Report/sections/07_evaluation}
\input{Technical_Report/sections/08_experiments}
\input{Technical_Report/sections/09_discussion}
\input{Technical_Report/sections/10_conclusion}

\clearpage
\bibliographystyle{unsrtnat}
\bibliography{references}

\clearpage
\appendix
\input{Technical_Report/sections/appendix}

\end{document}

%% file: Technical_Report/sections/00_abstract.tex
Recursive self-improvement (RSI) in production agents involves transforming accumulated experience into reliable model updates while adapting to evolving business policies. We present \textbf{CARE} (\textbf{C}ustomer-service
\textbf{A}gent \textbf{R}etrospective and \textbf{E}volve), a grounded RSI
framework that operationalizes this process in customer service. Its goal is
to replace a frontier proprietary incumbent with an in-house mid-sized model
that continues learning after deployment.
After a one-time cold start on governed production traces, the current model
serves production traffic alongside the incumbent, and its own failures drive
the next update. A 24/7 inspection process establishes failures through
evidence-blinded, independently reviewed judgments. Attribution separates model
errors from defects in Skills, tools, or the harness, routing component
failures to repair. Model-attributable trajectories are truncated at the first
erroneous decision and paired with verified repairs. On-policy
self-distillation (OPSD) then updates the model using the same model as
teacher, with the diagnosis and repair supplied as privileged information
unavailable to the student.
To support continued evolution as business requirements change, training
mixes failures from different Skill/SOP versions, each judged against the
policy in force when it occurred. Candidates run an executable simulation
test compiled from the latest SOP and then frozen, followed by offline
inspection of the resulting traces. Promotion requires a combined
simulation-and-inspection score at least at parity with the deployed version,
confirmation through an online comparison with the incumbent, and human
approval.
In a multilingual food-delivery deployment, five RSI iterations reduce the
inspection Bad Case rate of our in-house 31B model, \caremodel, from 6.42\%
after cold start to 1.97\%, below the incumbent GPT-5.4's 4.60\%, raise its
simulation accuracy from 93.0\% to 94.1\%, and bring its online user
satisfaction from 4.6 points below the incumbent to parity (1.7 points above
it in the latest period). \caremodel also lowers median reply latency by
27.7\%, and replacing the incumbent completely lowers the serving cost per
service order by 70.5\%.

%% file: Technical_Report/sections/01_introduction.tex
\section{Introduction}
\label{sec:intro}

Large language model (LLM) agents now resolve a substantial share of
customer-service conversations end to end. They interpret user requests,
consult business state through tools, follow a Skill or standard operating
procedure (SOP), and take actions such as refunds, compensation, or handoff to
a human. In many deployments, a frontier proprietary model powers the agent.
This \emph{incumbent} delivers strong quality, but serving it at scale is
expensive, its inference adds latency to every turn, and the operator has
limited control over its behavior, update schedule, and data boundaries.
These constraints motivate post-training an \emph{in-house} mid-sized model
that can replace the incumbent in production.

This replacement requires more than one-time distillation for three reasons.
First, post-training on historical traces leaves a residual quality gap
concentrated in long-tail situations. These failures emerge only when the
model is deployed and encounters the state distribution induced by its own
behavior. Second, the production environment is non-stationary: Skills and
SOPs are revised, tool schemas change, and prompts are hot-fixed. Behavior
that is correct under one configuration may become incorrect under the next.
Third, once the incumbent is retired its production traces stop, so fresh
experience can only come from the replacement's own behavior. The replacement must therefore continue learning
from its own experience while adapting to a changing environment. We frame
this requirement as \emph{recursive self-improvement} (RSI).

The idea of a system that improves the process by which it improves dates back
to \citet{good1965} and was formalized in self-referential agents such as the
G\"odel machine~\citep{godelmachine}. In the LLM era, RSI usually denotes a loop
in which a system turns its own experience and feedback into persistent updates
to its components, including weights, prompts, memory, tools, or harness
code~\citep{surveysevolving,modernsurvey,rsi_taxonomy}. Recent analyses
identify the grounding of feedback as a central constraint on such loops.
When the same model generates and evaluates outputs, correlated errors can
reinforce confident mistakes. Recursive training on self-generated data can
also erode the tails of the distribution~\citep{collapse},
and LLM judges can be gamed~\citep{judgeguard,fragility}. Production
customer service compounds these difficulties: there is no verifier that
certifies an entire conversation as correct; a failure may originate in the model, a
Skill, a tool, the harness, the simulator, the judge, or the infrastructure;
and experimentation on real users is constrained.

\input{Technical_Report/figs/care_loop}

We present \care (Customer-service Agent Retrospective and Evolve), a grounded
RSI framework for replacing an incumbent model in production
(Figure~\ref{fig:care_loop}). A one-time \textbf{cold start} (Phase~I) builds
the initial policy from governed production traces of the incumbent system,
using a data engine that admits, allocates, and refines conversations.

Thereafter, the improvement loop runs on the in-house model's own experience
(Phase~II). The current model serves part of the production traffic, and a 24/7 \textbf{online
inspection} stage examines its conversations through evidence-blinded,
independently reviewed judgments. For each confirmed failure, an
\textbf{attribution} stage determines whether responsibility lies with the
model or a non-model component. Component failures are repaired outside the
model. For each model-attributable failure, the trajectory is truncated at the
first erroneous decision and a verified repair is constructed. The model then
\textbf{evolves} through on-policy self-distillation (OPSD): the student
samples its own decision at that point, while the same model serves as the
teacher with access to the diagnosis and verified repair as privileged
information.

A separate \textbf{evaluation system} determines whether the candidate is
ready to replace the deployed version. The candidate runs an executable
simulation test compiled from the latest Skill/SOP and then frozen; offline
inspection evaluates the resulting simulation traces. Its combined score must
be at least at parity with the deployed version before an online comparison
with the incumbent. Failures used for learning and cases used for evaluation
come from separate sources and are never mixed. After cold start, the model's
own failures supply the training data, and the model itself supplies the
teacher.

The loop is not free of external intervention. An external frontier model,
not the in-house model, drives online inspection, attribution, and repair
construction; component owners repair non-model failures; and a person approves
each release. The external model's output reaches training only as privileged
information for the teacher, so \care is self-improving in the source of its
experience and of its teacher, not in the sense of running without outside help.
The operating context also changes independently of the loop: Skills and SOPs
can be revised at any time, including while a version is serving. Until the next
update the deployed model has to follow a revision from its input alone, which
is why training mixes several policy versions, and each simulation test is
compiled from the current version. Section~\ref{sec:loop_scope} sets out both
points.

In Phase~II, the reliance on self-generated training data and teacher supervision introduces a risk of error amplification, when model-generated outputs are evaluated without independent verification. Three checks outside the model therefore govern the loop. Reviewed
evidence, rather than a single judge score, establishes whether a production
conversation contains a failure. Attribution determines whether that failure
warrants a model update or a component repair. In simulation, executed
business state establishes whether an action such as a refund or cancellation
actually occurred, independently of the model's description of that action.
Attribution also supplies privileged information for self-distillation: the
teacher's information advantage comes from being told what went wrong, not
from critiquing its own output.

Our contributions are as follows.
\begin{enumerate}[itemsep=3pt]
  \item \textbf{Problem formulation.} We formalize production model replacement
  as grounded RSI, with an explicit replacement criterion and an iteration in
  which the model supplies both its experience and its teacher
  (Section~\ref{sec:problem}).
  \item \textbf{Grounded retrospection and evaluation.} A 24/7, layered,
  evidence-blinded online inspection process and a failure-attribution stage
  separate model errors from component errors before learning
  (Section~\ref{sec:retrospection}). A separate evaluation system assesses
  every candidate through an executable, partially observable simulation test
  and offline inspection of the resulting traces (Section~\ref{sec:evaluation}).
\item \textbf{Attribution-guided on-policy self-distillation.} We truncate each attributed failure trajectory at its first erroneous decision, construct a verified minimal repair using hindsight information, and train the model on its own outputs at the identified erroneous decision under the supervision of a hindsight-informed self-teacher. Hindsight information is excluded from the student's input. Training incorporates failures from different Skill/SOP versions, each evaluated against its corresponding SOP, to improve the model's ability to make decisions consistent with the provided policy (Section~\ref{sec:evolve}).
  \item \textbf{Production study.} We report results from a multilingual
  food-delivery customer-service deployment covering seven regions, including
  comparisons with frontier models, online A/B comparisons, five RSI
  iterations that produce six model versions, and ablations of the learning
  signal and of the OPSD loss
  (Section~\ref{sec:experiments}).
\end{enumerate}

Over five iterations, the inspection Bad Case rate of our in-house 31B model, \caremodel, decreases from 6.42\% after cold start to 1.97\%, less than half of the incumbent's 4.60\% and the lowest rate among all evaluated models. Simulation accuracy improves from 93.0\% to 94.1\%, behind the incumbent and Gemini-3.5-Flash, and the combined evaluation score reaches 96.07 against the incumbent's 96.10. In the online deployment across all seven regions, user satisfaction improves from 4.6 percentage points below the incumbent at cold start to comparable performance in the final two versions (+0.3 and +1.7 points).

These results highlight the limitations of cold start alone. Although the cold-start model completes substantially more simulated cases than the untuned base model, it produces more confirmed failures than both the base model and the incumbent. Recursive self-improvement addresses this gap through iterative refinement.

The ablations show why the learning signal matters at this data scale. On a small increment of repairs, supervised fine-tuning collapses during training, and preference optimization collapses into immediate handoff, passing 15.1\% of simulated cases. OPSD keeps a simulation accuracy of 94.1\%. Among its loss configurations, a forward KL divergence truncated to the teacher's top 32 tokens performs best overall and is the one deployed.

Section~\ref{sec:related} positions \care within the RSI literature.
Section~\ref{sec:problem} formalizes the setting.
Sections~\ref{sec:cold_start}--\ref{sec:evaluation} describe cold start,
retrospection, evolution, and evaluation. Section~\ref{sec:experiments} reports
the experiments, and Section~\ref{sec:discussion} discusses limitations and
risks.

%% file: Technical_Report/figs/care_loop.tex
\begin{figure}[t]
\centering
\hyphenpenalty=10000
\resizebox{\linewidth}{!}{%
\begin{tikzpicture}[
  font=\small,
  stage/.style={rectangle, rounded corners=3pt, draw=#1!80!black, fill=#1!9,
    line width=0.8pt, align=center, text width=2.75cm, minimum height=1.25cm,
    inner sep=3pt},
  flow/.style={-{Stealth[length=2.6mm]}, line width=1.0pt, draw=careink!75},
  aux/.style={-{Stealth[length=2.2mm]}, line width=0.8pt, draw=carered!70,
    dashed},
  lbl/.style={font=\scriptsize\itshape, text=careink!80, align=center,
    inner sep=1.5pt},
  phase/.style={rectangle, rounded corners=7pt, draw=#1!55, fill=#1!3,
    line width=0.7pt, dash pattern=on 3pt off 2pt},
  phaselabel/.style={font=\small\bfseries, text=#1!75!black},
]
\node[stage=caregray] (traces) at (-9.0,3.1)
  {\textbf{Production traces}\\[1pt]\scriptsize incumbent $\pistar$ + linked\\
   human conversations};
\node[stage=caregreen] (engine) at (-9.0,1.55)
  {\textbf{Data engine}\\[1pt]\scriptsize admit $\cdot$ allocate $\cdot$ refine};
\node[stage=caregreen] (sft) at (-9.0,0)
  {\textbf{Cold-start SFT}\\[1pt]\scriptsize $\theta_{\mathrm{base}}\to\theta_0$};

\node[stage=careblue] (deploy) at (-4.3,0)
  {\textbf{Deploy}\\[1pt]\scriptsize $\pi_{\theta_k}$ serves part of\\
   production traffic};
\node[stage=careorange] (retro) at (-2.15,2.35)
  {\textbf{Online inspection}\\[1pt]\scriptsize 24/7, evidence-blinded\\
   $\to$ confirmed failures};
\node[stage=careorange] (attr) at (2.15,2.35)
  {\textbf{Attribute}\\[1pt]\scriptsize model vs.\ component\\ $\to$ diagnosis $d$};
\node[stage=careteal] (construct) at (4.3,0)
  {\textbf{Construct}\\[1pt]\scriptsize verified repairs at the\\
   first model error};
\node[stage=careteal] (evolve) at (2.15,-2.35)
  {\textbf{Evolve (OPSD)}\\[1pt]\scriptsize self-teacher sees hindsight;\\
   student does not};
\node[stage=carepurple] (gate) at (-2.15,-2.35)
  {\textbf{Evaluate \& promote}\\[1pt]\scriptsize offline tests, then online\\
   comparison with $\pistar$};
\node[stage=carered] (repair) at (2.15,4.65)
  {\textbf{Component repair}\\[1pt]\scriptsize Skill/SOP $\cdot$ tool $\cdot$ harness\\
   judge $\cdot$ infra};
\node[stage=caregray] (context) at (-4.3,4.65)
  {\textbf{Operating context} $\ctx$\\[1pt]\scriptsize Skill/SOP, tools, prompt\\
   revised at any time};

\foreach \n/\a/\c in {retro/north east/south east,attr/north west/south west,construct/north east/south east}
  \node[font=\tiny\itshape, text=carered!85!black, anchor=\a, inner sep=1pt]
    at (\n.\c) {external model};

\node[stage=carepurple, fill=carepurple!4] (simtest) at (-3.65,-4.95)
  {\textbf{Simulation test}\\[1pt]\scriptsize executable environment\\
   $\to$ simulation accuracy};
\node[stage=carepurple, fill=carepurple!4] (offinsp) at (0.55,-4.95)
  {\textbf{Offline inspection}\\[1pt]\scriptsize of the simulation traces\\
   $\to$ Bad Case rate};
\node[align=center] at (0,0)
  {{\large $\pi_{\theta_k}\;\longrightarrow\;\pi_{\theta_{k+1}}$}\\[4pt]
   {\scriptsize\color{caregray} grounded by}\\
   {\scriptsize\color{careink} executed state $\cdot$ reviewed evidence}\\
   {\scriptsize\color{careink} causal attribution}};

\draw[flow] (traces) -- (engine);
\draw[flow] (engine) -- (sft);
\draw[flow] (sft) -- node[lbl,above] {$\theta_0$} (deploy);

\draw[flow] (deploy) to[bend left=24] node[lbl,right=3pt] {own\\trajectories} (retro);
\draw[flow] (retro) to[bend left=18] node[lbl,above=1pt] {confirmed\\failures} (attr);
\draw[flow] (attr) to[bend left=24] node[lbl,above right] {model-\\attributable} (construct);
\draw[flow] (construct) to[bend left=24] node[lbl,below right] {$\Delta_k$} (evolve);
\draw[flow] (evolve) to[bend left=18] node[lbl,below=1pt] {candidate $\theta'$} (gate);
\draw[flow] (gate) to[bend left=24] node[lbl,below left] {promote\\$\theta_{k+1}$} (deploy);

\draw[flow,draw=carered!80] (attr) -- node[lbl,right] {non-model\\cause} (repair);
\draw[aux] (repair.west) -- node[lbl,above] {revised Skill/SOP and tools} (context.east);
\draw[aux] ([xshift=-0.85cm]context.south) -- node[lbl,right,pos=0.4] {through\\the input}
  ([xshift=-0.85cm]deploy.north);
\draw[aux] (context.west) -- ++(-0.25,0) |- node[lbl,right,pos=0.37] {test\\recompiled}
  (simtest.west);

\coordinate (evalbottom) at ([yshift=-8pt]simtest.south);
\begin{scope}[on background layer]
\node[phase=caregreen, fit=(traces)(engine)(sft), inner sep=9pt,
  label={[phaselabel=caregreen]above:Phase I $\cdot$ cold start (once)}] {};
\node[phase=careteal, fit=(deploy)(retro)(attr)(construct)(evolve)(gate)(repair)(context)
  (evalbottom)(offinsp),
  inner xsep=12pt, inner ysep=15pt,
  label={[phaselabel=careteal]above:Phase II $\cdot$ inspect, evolve, evaluate
  (recurring, $k=0,1,2,\dots$)}] {};
\node[rectangle, rounded corners=5pt, draw=carepurple!60, fill=carepurple!7,
  line width=0.7pt, fit=(simtest)(offinsp), inner xsep=8pt, inner ysep=6pt,
  label={[font=\scriptsize\bfseries, text=carepurple!80!black, inner sep=2pt]
  above:Evaluation system $\cdot$ frozen test compiled from the latest Skill/SOP}]
  (evalsys) {};
\end{scope}
\draw[flow, draw=carepurple!85!black] ([yshift=13pt]evalsys.north -| gate) -- node[lbl,right] {metrics} (gate.south);
\draw[flow, draw=carepurple!85!black] (simtest) -- node[lbl,above=1pt] {traces} (offinsp);
\end{tikzpicture}}
\caption{Overview of \care. Phase~I bootstraps an in-house policy
$\pi_{\theta_0}$ from governed production traces of the incumbent system.
Phase~II is the recursive loop. The current model $\pi_{\theta_k}$ serves part
of the production traffic; 24/7 online inspection turns its own conversations
into reviewed failures; attribution separates model-attributable failures from
component failures, which are repaired outside the model; model-attributable
failures are cut at the first erroneous decision and repaired, and the model
distills from itself on that decision, with the hindsight visible only to the
teacher. The candidate is measured by the evaluation system, where it runs a
simulation test compiled from the latest Skill/SOP and then frozen, and its
simulation traces are inspected offline; it is promoted only if the combined
score of the two is at least at parity with the deployed version and an online
comparison with the incumbent confirms it. After cold
start, training data come from the model's own failures and the teacher is the
model itself. Stages tagged in red are driven by an external frontier model rather
than by the in-house model. The operating context is revised independently of the
loop: a revision reaches the deployed model through its input and the evaluation
system through a recompiled test (Section~\ref{sec:loop_scope}).}
\label{fig:care_loop}
\end{figure}
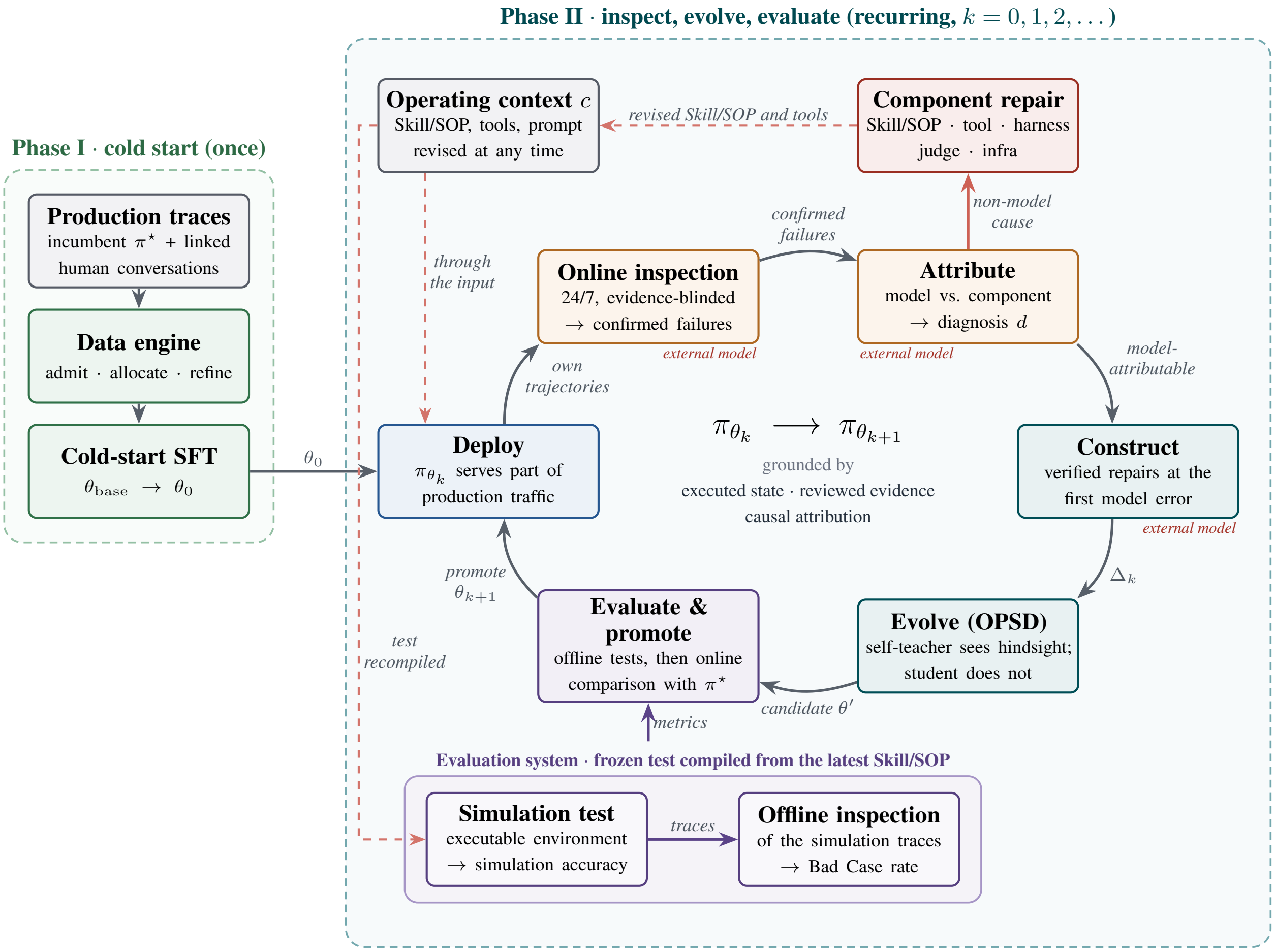

%% file: Technical_Report/sections/02_related_work.tex
\section{Background and Related Work}
\label{sec:related}

\subsection{Recursive Self-Improvement}
\label{sec:related_rsi}

\citet{good1965} introduced the concept of an ``intelligence explosion,'' in which a machine improves the design of its successors. The G\"odel machine~\citep{godelmachine} formalized a self-referential agent capable of modifying its own code upon proving the benefits of such modifications. Contemporary recursive self-improvement (RSI) extends these ideas to systems that iteratively update their components through accumulated experience and feedback~\citep{surveysevolving,modernsurvey,rsi_taxonomy,rsi_workshop}. Existing approaches differ in their update mechanisms and supervision signals. STaR improves reasoning through self-generated rationales leading to correct answers~\citep{star}, SPIN learns from comparisons between model-generated responses and human demonstrations~\citep{spin}, and Self-Rewarding Language Models use model-generated judgments to improve both the policy and its reward model~\citep{selfreward}. Beyond model parameters, Reflexion incorporates verbal feedback into subsequent attempts~\citep{reflexion}, Voyager accumulates executable skills through environment interaction~\citep{voyager}, and the Darwin G\"odel Machine evolves agent code through empirical evaluation~\citep{dgm}.

The reliability of self-generated feedback remains a central challenge in RSI. Shared biases between generators and evaluators can reinforce existing errors, while recursive training on generated data may lead to model collapse~\citep{collapse}. Previous studies also report high variance, task-order sensitivity, and vulnerabilities in LLM-based evaluation~\citep{fragility,judgeguard}. Formal verification and execution feedback provide more direct evidence of correctness than model-generated judgments alone~\citep{rsi_taxonomy}. A recent taxonomy~\citep{rsi_taxonomy} distinguishes \emph{bounded} self-improvement, where updates are evaluated against fixed external criteria, from \emph{open-ended} RSI, which can also modify the mechanisms governing improvement. It further categorizes improvement loops according to human involvement: \emph{human-in-the-loop} systems require human review of individual updates, \emph{human-on-the-loop} systems automate improvement while retaining human oversight and deployment control, and \emph{closed-loop} systems operate without human intervention. Within this taxonomy, \care adopts a bounded, human-on-the-loop approach in which retrospection, error attribution, and model training operate automatically, while promotion to production remains subject to evaluation and deployment gates. Learning signals are grounded in simulated business-state transitions, reviewed inspections, and verified repairs, while failures attributed to non-model components are directed to component repair.

\subsection{On-Policy and Privileged-Information Distillation}
\label{sec:related_distillation}

On-policy learning addresses compounding errors arising from the mismatch between training distributions and states encountered during execution~\citep{dagger}. Generalized Knowledge Distillation (GKD) extends this principle to language models by sampling sequences from the student and minimizing divergence from teacher predictions~\citep{gkd}. Recent research further examines the training dynamics and practical considerations of on-policy distillation for LLMs~\citep{opd}. Learning Using Privileged Information (LUPI) introduces supervision from information available during training but inaccessible during inference~\citep{lupi}. Privileged Information Distillation extends this principle to multi-turn agents, where a teacher conditioned on privileged actions guides a student without access to those actions~\citep{privileged}. On-Policy Self-Distillation (OPSD) enables a single model to serve as both teacher and student, using verified reasoning traces in the teacher context to minimize token-level divergence on student-generated trajectories~\citep{opsd_reasoner}.

\care adopts the self-distillation framework of OPSD while introducing failure-specific hindsight as privileged information. Unlike approaches using reference solutions or privileged actions, the self-teacher receives a reviewed, causally attributed diagnosis and a verified minimal repair at the first erroneous decision. The student learns from its own decision outputs under hindsight-informed supervision without access to hindsight in its input. This enables recursive model improvement through verified supervision derived from the model's own production failures.

\subsection{Interactive Evaluation and Continual Adaptation}
\label{sec:related_evaluation}

Interactive agent evaluation examines decision-making through multi-turn interactions, tool execution, and environment-state transitions. $\tau$-bench evaluates tool--agent--user interactions under domain-specific policies~\citep{taubench}, while $\tau^2$-bench extends this setting to environments where both users and agents can modify shared state~\citep{tau2bench}. These benchmarks motivate the explicit state transitions and outcome verification adopted by the \care simulator. Agent Lightning and DataEnvGym further investigate the separation of agent execution from training and data-generation infrastructure~\citep{agentlightning,dataenvgym}.

Continual learning introduces challenges including catastrophic forgetting, distribution shift, and evaluation leakage during successive model updates~\citep{cl_survey}. Production agents face an additional challenge as business policies evolve independently of model updates. \care addresses this setting by incorporating failure trajectories from different Skill/SOP versions, with each trajectory evaluated against the SOP applicable during its execution. This enables the model to learn decisions consistent with the provided policy rather than relying on fixed policy-specific behaviors.

Table~\ref{tab:related} compares representative self-improvement and distillation methods in terms of updated components, experience sources, grounding signals, teacher configurations, and adaptation to changing policy contexts.

\begin{table}[t]
\caption{Positioning of \care among representative self-improvement and distillation methods. ``Grounding'' denotes the evidence or supervision used to guide improvement, while ``Contexts'' indicates whether the method explicitly addresses changing policy requirements during iterative improvement. A dash indicates that the corresponding feature is not explicitly addressed.}
\label{tab:related}
\centering
\footnotesize
\setlength{\tabcolsep}{4pt}
\begin{tabularx}{\linewidth}{@{}l>{\raggedright\arraybackslash}p{2.15cm}YYYc@{}}
\toprule
Method & Updated component & Experience source & Grounding & Teacher & Contexts \\
\midrule
STaR~\citep{star} & weights & own rationales & answer correctness & -- & -- \\
SPIN~\citep{spin} & weights & own generations & human demonstrations & previous self & -- \\
Self-Rewarding~\citep{selfreward} & weights & own responses & self-judgment & self & -- \\
Reflexion~\citep{reflexion} & prompt / memory & own attempts & env.\ signal + self-reflection & -- & -- \\
Voyager~\citep{voyager} & skill library & own episodes & environment execution & -- & -- \\
DGM~\citep{dgm} & agent code & own code edits & benchmark evaluation & -- & -- \\
GKD~\citep{gkd} & weights & own rollouts & teacher likelihood & external model & -- \\
$\pi$-Distill~\citep{privileged} & weights & own rollouts & privileged actions & PI-conditioned teacher & -- \\
OPSD~\citep{opsd_reasoner} & weights & own rollouts & verified solutions & self + PI & -- \\
\midrule
\textbf{\care} & weights; routed component repair & own failures in simulation and production & executed state, reviewed inspection, attribution, verified repair & self + hindsight & \checkmark \\
\bottomrule
\end{tabularx}
\end{table}

%% file: Technical_Report/sections/03_problem.tex
\section{Problem Formulation}
\label{sec:problem}

\subsection{Agents, policies in context, and the incumbent}
A customer-service agent operates under business rules specified through its input rather than fixed within the model parameters. We define the \emph{operating context} as
\begin{equation}
  \ctx=(s,\,w,\,p),
  \label{eq:context}
\end{equation}
where $s$ denotes the Skill/SOP governing agent behavior, $w$ the available tools and their schemas, and $p$ the prompting instructions. The correct action for a given request depends on the operating context, as different Skills/SOPs and tool configurations may impose different requirements and constraints. As Skills and SOPs evolve during operation, the operating context changes accordingly.

An interaction under operating context $\ctx$ generates a trajectory
\begin{equation}
    \tau = (x_0, o_1, a_1, y_1, \ldots, o_T, a_T, y_T, z),
    \label{eq:trajectory}
\end{equation}
where $x_0$ denotes the initial business state (e.g., order, delivery, or after-sales status), $o_t$ the agent-visible observation, $a_t$ the agent's action (e.g., a response, tool query, or business operation), $y_t$ the corresponding tool or environment result, and $z$ the terminal business outcome. The agent follows a policy $\pi(a_t\mid h_t,\ctx)$, which determines the next action given the agent-visible interaction history $h_t$ and operating context $\ctx$. We denote the incumbent model serving production by $\pistar$ and the in-house model by $\pith$, initialized from an open-weight base model with parameters $\theta_{\mathrm{base}}$.

The incumbent serves three roles: its production trajectories provide the initial training data for cold start, its measured performance establishes the replacement benchmark, and it remains available as a fallback during model promotion. Beyond these roles, the incumbent is not used to generate training labels during recursive self-improvement.

\subsection{Replacement and improvement criteria}

Let $Q(\pi,\ctx)$ denote the expected business and service quality under
operating context $\ctx$, $H(\pi,\ctx)$ the rate of hard violations
(e.g., prohibited actions or invalid state transitions), and $K(\pi)$
the average serving cost per interaction. We define the replacement
criteria separately for each operating context rather than aggregating
performance across contexts.

\begin{definition}[Replacement]
\label{def:replacement}
An in-house policy $\pith$ satisfies the replacement criteria relative
to the incumbent $\pistar$ over a context set $\Ctx$, with quality
tolerance $\delta \geq 0$, if, for every $\ctx \in \Ctx$,
\begin{equation}
  Q(\pith,\ctx) \geq Q(\pistar,\ctx)-\delta,
  \qquad
  H(\pith,\ctx) \leq H(\pistar,\ctx),
  \qquad
  K(\pith) < K(\pistar).
  \label{eq:replacement}
\end{equation}
\end{definition}

These criteria define the model replacement objective rather than
the evaluation procedure for individual updates. Continued improvement
remains necessary as new failures emerge and business policies evolve
during deployment.

\begin{definition}[Grounded RSI iteration]
\label{def:rsi}
Given the current model parameters $\theta_k$, an iteration of \care
generates experience $\mathcal T_k \sim \pi_{\theta_k}$ and produces
a candidate model through
\begin{equation}
\begin{aligned}
  \mathcal D_k &=
  (\mathrm{Construct}\circ\mathrm{Attribute}\circ
  \mathrm{Retrospect})(\mathcal T_k),\\
  \theta'_k &= \mathrm{Evolve}(\theta_k;\mathcal D_k),\\
  \theta_{k+1} &=
  \begin{cases}
    \theta'_k, & \text{if } \mathrm{Gate}(\theta'_k,\theta_k)=1,\\
    \theta_k, & \text{otherwise}.
  \end{cases}
\end{aligned}
\label{eq:rsi}
\end{equation}
The current model generates the training experience and serves as
the self-teacher during evolution. External information provides
grounding through executed business states, reviewed evidence, and
verified repairs, with hindsight available only to the teacher.
The binary promotion gate accepts a candidate if its combined offline
score meets or exceeds that of the deployed model, an online comparison
with the incumbent confirms its suitability, and human approval is
obtained. The combined score incorporates simulation accuracy on a
test set compiled from the latest Skill/SOP and frozen for candidate
evaluation, together with offline inspection of the resulting
simulation trajectories (Section~\ref{sec:evaluation},
Eq.~\eqref{eq:gate}).
\end{definition}

Definition~\ref{def:rsi} specifies what ``self'' means in \care:
the model supplies both the training experience and the self-teacher,
while external evidence and verification establish the correctness
of the learning signals. Each promoted iteration is subject to
independent evaluation and controlled deployment.

\subsection{What the loop depends on}
\label{sec:loop_scope}

\paragraph{External intervention.}
Definition~\ref{def:rsi} is narrower than a loop that improves without external
intervention. The in-house model supplies the experience and the teacher, but
$\mathrm{Retrospect}$, $\mathrm{Attribute}$, and $\mathrm{Construct}$ in
Eq.~\eqref{eq:rsi} are driven by an external frontier model, such as GPT, rather
than by $\pith$. People also take part: component owners and engineers repair
non-model failures, cases that cannot be confirmed may go to manual review, and a
person approves each promotion. The external model's output reaches training only
as privileged information for the self-teacher, and the student never sees it
(Section~\ref{sec:opsd}). We keep these stages outside the in-house model because
a model that identifies and diagnoses its own failures would reinforce its own
errors; independent judgment is what grounds the loop. \care is therefore
self-improving in its weights and training experience, not autonomous.

\paragraph{Changes to the operating context.}
A change of $\ctx$ is not a step of the loop. Business owners revise Skills, SOPs,
tools, and prompts on their own schedule, and the component repairs that
attribution triggers change them too, so a revision can land at any stage of an
iteration. It reaches the loop in three places. While a version is serving, a
revision reaches the model through its input and the weights stay the same, so
until the next iteration the deployed model has to adapt by following the policy
it is given. If it fails, online inspection picks the failure up and judges it
against the revised policy, so the next iteration learns from it. During
construction and training, each failure keeps the context in which it occurred,
and a training set mixes several policy versions so that the model learns to
follow the policy it is given and to generalize across revisions
(Section~\ref{sec:replay}). During evaluation, the simulation test is compiled
automatically from the current Skill/SOP and then frozen, so a candidate is
measured against the policy in force (Section~\ref{sec:evaluation}).

\subsection{Requirements of grounded RSI in production}
\label{sec:requirements}

Reliable improvement under Eq.~\eqref{eq:rsi} depends on the
validity of the feedback, attribution, learning, and evaluation
processes. We identify four requirements for grounded RSI
in production, each addressed by a component of \care.

\paragraph{R1: Grounded failure identification.}
Customer-service interactions lack a universal correctness
verifier, making failure identification dependent on executed
business states and independently reviewed evidence.
Reliable retrospection requires distinguishing confirmed
failures from inconclusive judgments while preventing
evaluation shortcuts arising from model identities, prior
quality labels, or other source metadata. Independence
between training failures and evaluation cases is further
required to avoid evaluation leakage
(Sections~\ref{sec:retrospection} and~\ref{sec:evaluation}).

\paragraph{R2: Causal attribution before learning.}
A failed interaction may originate from model behavior,
defective Skills or prompts, tool or routing errors,
simulator artifacts, judge errors, or infrastructure
failures. These causes require different corrective
interventions, and treating them uniformly as model
errors can encourage the model to compensate for
defective components. Reliable learning therefore
depends on attributing failures to their underlying
causes and restricting model updates to failures
attributable to model behavior
(Section~\ref{sec:attribution}).

\paragraph{R3: Learning from self-generated failures without reinforcing errors.}
Confirmed failures provide evidence of incorrect execution, whereas effective model improvement depends on identifying the first erroneous decision and constructing a verified repair. Directly incorporating failed trajectories into training risks propagating the errors embedded in those trajectories. Reliable on-policy learning therefore requires supervision grounded in verified repairs, applied to the model's own generated decisions, with privileged hindsight accessible exclusively to the self-teacher (Section~\ref{sec:opsd}).

\paragraph{R4: Policy-conditioned learning and safe promotion.}
Evolving Skills and SOPs introduce policy variations across trajectories collected during successive iterations, such that similar interactions may require different actions under different policy versions. Learning without accounting for these variations risks reinforcing decisions that are inconsistent with the operating context. Policy-conditioned learning therefore preserves the Skill/SOP associated with each failure and verifies its corresponding repair against the policy in force at the time, enabling training across multiple policy versions (Section~\ref{sec:replay}). Reliable promotion further requires evaluating candidate models on a simulation test compiled from the latest Skill/SOP and frozen for candidate evaluation, followed by offline inspection of the resulting trajectories, online comparison with the incumbent, and human approval (Section~\ref{sec:evaluation}).

%% file: Technical_Report/sections/04_cold_start.tex
\section{Phase I: Cold Start from Production Traces}
\label{sec:cold_start}

Cold start produces the initial policy $\pi_{\theta_0}$ that enters the
recursive loop. Its raw material is the logged production behavior of the
incumbent system: complete agent conversations with tool definitions, calls,
results, and Skill content, together with conversations handled by human
support staff. These logs are abundant but uneven. They contain execution
defects, language errors, and scenario imbalance, and an incumbent trace that
\emph{looks} fluent may still take an action that contradicts the business
state. We therefore treat cold start as a data-governance problem with three
separate decisions (Figure~\ref{fig:data_engine}): which conversations are
admissible as positive demonstrations, how a fixed training budget is allocated
among them, and whether their wording can be improved without changing their
behavior. For raw candidates $D$, admitted candidates $D_s$, selected members
$S$, and retained output $F$, the engine enforces
\begin{equation}
  \operatorname{IDs}(F)\subseteq\operatorname{IDs}(S)\subseteq
  \operatorname{IDs}(D_s)\subseteq\operatorname{IDs}(D),
  \label{eq:membership}
\end{equation}
so later stages can only remove members, never introduce new ones.

\input{Technical_Report/figs/data_engine}

\subsection{Admission}
\label{sec:admission}
The learning unit is a complete conversation rather than an isolated reply,
because a reply that is appropriate in one business state can be wrong in
another. Exports may contain several cumulative snapshots of one conversation;
the engine keeps the most complete snapshot, deduplicates by conversation
identity, normalizes region, scenario, and provenance, and attaches the context
key $\ctx$ of Eq.~\eqref{eq:context}. 

\paragraph{Language consistency.}
Multilingual customer service introduces language-related defects,
including unintended script mixing, word-internal concatenation,
and intrusion of unrelated languages. A rule-based detector
identifies potentially anomalous spans in assistant messages,
which are subsequently evaluated by a model using the surrounding
conversational context. Protocol fields, URLs, placeholders,
brand names, and user-initiated code-switching are excluded from
defect classification. Conversations with confirmed or unresolved
language issues are excluded from the admitted pool.

\paragraph{Structural quality.}
Deterministic checks then filter scenario scope, validate action chains, and
compute context length with the target tokenizer and chat template, including
tool definitions. Over-length conversations are excluded rather than
truncated, since truncation silently changes the decision problem. Known risk
patterns, such as inappropriate handoff placement, formulaic online-presence
confirmations, claims of real-time processing, and placeholder images, are
removed at this stage.

\paragraph{Evidence-anchored semantic screening.}
Structural validity alone provides insufficient evidence of behavioral
correctness. A screening model examines every structurally qualified
conversation, including tool execution results and structured business
facts, to identify execution defects supported by identifiable evidence.
Representative cases include applying a missing-item procedure to a
whole-order non-receipt request and claiming courier contact without a
corresponding executed action. Weak indicators, including low customer
satisfaction, prolonged interactions, and requests for human assistance,
provide insufficient grounds for confirming a defect. Each conversation
$x$ receives a verdict
$V(x)\in\{\texttt{no\_problem},\texttt{problem},\texttt{inconclusive}\}$,
where \texttt{inconclusive} denotes cases whose classification depends
on unavailable runtime evidence. Admission is restricted to
conversations classified as \texttt{no\_problem}:
\begin{equation}
  D_s = \{x \in D_q : V(x) = \texttt{no\_problem}\},
  \label{eq:screening}
\end{equation}
where $D_q$ denotes the structurally qualified candidate pool.
Semantic screening is applied to the complete candidate pool before
allocation, ensuring that training quotas are determined from admitted
conversations. The screening procedure prioritizes precision in
identifying \texttt{problem} cases, treating the configured
problem-rate limit as a ceiling rather than a target.

\subsection{Hierarchical, capacity-constrained allocation}
\label{sec:allocation}

Production traffic is dominated by a few regions and scenarios; in an
early snapshot, one scenario alone accounted for 37.8\% of the
conversations. Sampling according to the empirical traffic distribution
would therefore underrepresent low-frequency regions and scenarios.
Allocation assigns a fixed budget $N$ hierarchically: first across four
region groups (Saudi Arabia, Brazil, Hong Kong, and a GCC4 group of the
United Arab Emirates, Bahrain, Kuwait, and Qatar), then across the
scenarios present in each group, and finally across the individual
regions within each group--scenario cell, limiting the dominance of
high-volume categories.

\paragraph{Cells and capacities.}
Every admitted conversation $x\in D_s$ carries a cell key
$\kappa(x)=(g,\sigma,r)$: its region group
$g\in\mathcal G=\{\mathrm{SA},\mathrm{BR},\mathrm{HK},\mathrm{GCC4}\}$, its
scenario $\sigma$, and its region $r\in\mathcal R_g$, where
$\mathcal R_{\mathrm{GCC4}}=\{\mathrm{AE},\mathrm{BH},\mathrm{KW},\mathrm{QA}\}$
and every other group contains a single region. The admitted capacity
of each cell and its marginals are
\begin{equation}
  c_{g\sigma r}=\bigl|\{x\in D_s:\kappa(x)=(g,\sigma,r)\}\bigr|,\qquad
  c_{g\sigma}=\sum_{r\in\mathcal R_g}c_{g\sigma r},\qquad
  c_g=\sum_{\sigma}c_{g\sigma}.
  \label{eq:capacity}
\end{equation}

\paragraph{Capacity-constrained equal split.}
The same capacity-constrained allocation principle is applied at each
hierarchical level. For a budget $M$ and capacities
$\mathbf c=(c_1,\dots,c_K)$, the continuous water-filling solution
$\mathbf n^*=(n_1^*,\dots,n_K^*)$ satisfies
\begin{equation}
  n_j^*=\min(c_j,\lambda),\qquad
  0\le n_j^*\le c_j,\qquad
  \sum_{j=1}^{K}n_j^*=M,
  \label{eq:waterfill}
\end{equation}
where the water level $\lambda$ solves
$\sum_j\min(c_j,\lambda)=M$. A scarce category contributes its entire
available capacity, and the remaining budget is distributed equally
among categories with residual capacity. The corresponding integer
allocation operator $\mathrm{WF}(M;\mathbf c)$ implements this
principle through the sequential assignment procedure described below.
The operator is applied hierarchically as follows:
\begin{equation}
  (n_g)_{g}=\mathrm{WF}\bigl(N;(c_g)_{g}\bigr),\qquad
  (n_{g\sigma})_{\sigma}=\mathrm{WF}\bigl(n_g;(c_{g\sigma})_{\sigma}\bigr),\qquad
  (n_{g\sigma r})_{r}=\mathrm{WF}\bigl(n_{g\sigma};(c_{g\sigma r})_{r}\bigr).
  \label{eq:hier}
\end{equation}
$\mathrm{WF}(M;\mathbf c)$ is feasible if and only if
$M\le\sum_j c_j$. Each parent capacity is the sum of its children's
capacities (Eq.~\eqref{eq:capacity}), and the constraints
$n_g\le c_g$ and $n_{g\sigma}\le c_{g\sigma}$ ensure the feasibility
of every subsequent allocation. Consequently, the overall selection
is feasible if and only if $|D_s|\ge N$. If $|D_s|<N$, selection fails
rather than reducing the specified training budget. A category with
$c_j=0$ receives zero allocation, reflecting the coverage limitations
of the admitted pool.

\paragraph{Integer assignment.}
The implementation assigns slots sequentially, prioritizing categories
with remaining capacity and the smallest current allocation. The
resulting integer quotas satisfy
\begin{equation}
  n_k-n_j\ge 2\;\Longrightarrow\;n_j=c_j,
  \qquad \forall j,k.
  \label{eq:maxmin}
\end{equation}
A category can therefore trail another by two or more slots only
after exhausting its available capacity, while categories with
remaining capacity differ by at most one slot. When category $k$
received its last slot, any category $j$ with fewer assigned slots
and unused capacity would have been selected first. This establishes
max--min fairness under capacity constraints. Proportional sampling,
$n_j=M c_j/\sum_k c_k$, would instead reproduce the traffic skew,
assigning a scenario that accounts for 37.8\% of conversations the
same proportion of the training budget.

\paragraph{Member selection.}
The allocation determines the number of conversations contributed
by each cell, while secondary criteria determine the selected
members:
\begin{equation}
  S=\bigcup_{g,\sigma,r}S_{g\sigma r},\qquad
  S_{g\sigma r}\subseteq\{x\in D_s:\kappa(x)=(g,\sigma,r)\},\qquad
  |S_{g\sigma r}|=n_{g\sigma r},
  \label{eq:selection}
\end{equation}
so $|S|=N$. Records are selected without duplication
($n_{g\sigma r}\le c_{g\sigma r}$), and quarantined records are
excluded from backfilling ($S\subseteq D_s$). Secondary criteria,
including handoff stage, data source, and multimodal content,
determine membership within each cell while preserving the
allocated quotas. The availability of linked human references and
eligibility for subsequent expression refinement are excluded from
selection criteria to prevent refinement opportunities from
influencing training exposure. The selected set $S$ is then frozen
as the training membership.

\subsection{State-aligned expression refinement}
\label{sec:refinement}

Incumbent responses generally follow the intended business
procedures, although their linguistic expression can be formulaic,
particularly when addressing customer frustration. Human support
conversations provide more natural expressions of empathy and
reassurance, while differing in operational permissions and
workflow states. We therefore use human conversations exclusively
as references for linguistic expression, preserving the business
facts and actions of the original agent trajectories. Refinement
operates on the frozen selected set $S$.

\paragraph{Representation.}
We represent a selected conversation as
$\tau=\bigl((z_1,u_1),\dots,(z_L,u_L)\bigr)$, where $z_i$
contains the protected fields of message $i$, including roles,
actions, FAQ codes, responsibility assignments, action extensions,
and tool calls and results, while $u_i$ denotes its natural-language
content. The projection onto protected fields,
$P(\tau)=(z_1,\dots,z_L)$, additionally preserves the number
and order of messages.

\paragraph{Reference labeling.}
Let $\tau^{\mathrm h}$ denote a human support conversation linked
to $\tau$ through an explicit conversation key. Ambiguous links
are excluded from reference matching, while conversations
without valid references retain their original expressions.
For an assistant message $i$ in $\tau$ and a human response
$v\in\tau^{\mathrm h}$, a labeling model determines their
alignment in business state and action,
$\alpha(v,i;\tau)\in\{0,1\}$. The eligible references are
\begin{equation}
  R_i(\tau)=\bigl\{v\in\tau^{\mathrm h}\setminus\mathcal X:\;
  \alpha(v,i;\tau)=1\bigr\},
  \label{eq:references}
\end{equation}
where $\mathcal X$ comprises rating solicitations, closing and
repetitive templates, empty apologies, self-introductions, and
private content. An alignment label $\alpha=1$ indicates
eligibility as a linguistic style reference, with business
facts retained exclusively from the original agent trajectory.

\paragraph{Eligible targets.}
Refinement is restricted to existing assistant messages with
at least one eligible reference:
\begin{equation}
  T(\tau)=\bigl\{i:\;R_i(\tau)\neq\varnothing\bigr\}.
  \label{eq:targets}
\end{equation}

\paragraph{Constrained rewrite.}
For $T(\tau)\neq\varnothing$, the rewriting model proposes
$\tau'$ through local modifications to eligible message text,
$u'_i=f\bigl(u_i;\tau,R_i(\tau)\bigr)$ for $i\in T(\tau)$.
The rewriting process preserves the original message structure,
actions, and tool interactions by enforcing
\begin{equation}
  P(\tau')=P(\tau),\qquad
  \{i:u'_i\neq u_i\}\subseteq T(\tau).
  \label{eq:protect}
\end{equation}
A validation gate $\gamma(\tau,\tau')\in\{0,1\}$ assesses
reference reliability, expected linguistic improvement, and
potential risks introduced by the rewrite. Rejected proposals
revert to the original conversation:
\begin{equation}
  \tilde\tau=
  \begin{cases}
    \tau' & \text{if } T(\tau)\neq\varnothing
            \text{ and } \gamma(\tau,\tau')=1,\\
    \tau  & \text{otherwise}.
  \end{cases}
  \label{eq:fallback}
\end{equation}

\paragraph{Final check.}
Preservation of structural fields alone is insufficient to
establish semantic equivalence. The refinement process
therefore recomputes context length and derived attributes
and repeats the language and structural checks of
Section~\ref{sec:admission}, including model-based language
adjudication where applicable. Let
$q(\tilde\tau)\in\{0,1\}$ denote the final quality check.
The retained corpus is
\begin{equation}
  F=\bigl\{\tilde\tau:\;\tau\in S,\;q(\tilde\tau)=1\bigr\},\qquad
  n^{F}_{g\sigma r}=\bigl|\{\tilde\tau\in F:
  \kappa(\tilde\tau)=(g,\sigma,r)\}\bigr|
  \le n_{g\sigma r}.
  \label{eq:finalize}
\end{equation}
Rejected rewrites retain their original conversations
(Eq.~\eqref{eq:fallback}), whereas failures in the final
quality check result in exclusion. Consequently, $F$
satisfies the membership constraint of
Eq.~\eqref{eq:membership}. The absence of resampling
allows the retained cell distribution $n^F$ to deviate
from the original allocation quotas, which are
reported separately.

\subsection{Cold-start objective}
\label{sec:cold_objective}
The retained set $F$ forms the cold-start corpus $D_0$. Starting from
$\theta_{\mathrm{base}}$, the instruction-tuned Gemma-4-31B-IT checkpoint, we
minimize the masked negative log-likelihood
\begin{equation}
  \mathcal L_{\mathrm{cold}}(\theta)=
  -\frac{1}{Z}\sum_{i\in D_0}\sum_{j} m_{ij}\,
  \log\pith\bigl(y_{ij}\mid y_{i,<j},\ctx_i\bigr),
  \qquad Z=\sum_{i,j}m_{ij},
  \label{eq:cold}
\end{equation}
where $m_{ij}$ selects assistant tokens. System-injected context cards, tool
results, and business state serve as conditioning inputs; supervision begins
with the first assistant response or tool call following the first user message.
This objective trains the model to generate actions and responses conditioned
on the supplied Skills, tool contracts, and business context, providing a
context-dependent initialization for subsequent recursive improvement.

The training configuration of this stage is provided in
Appendix~\ref{app:cold_config}, while historical revisions of the data
preparation process, including earlier approaches that concatenated or
inserted human turns, are documented in
Appendix~\ref{app:data_history}.

%% file: Technical_Report/figs/data_engine.tex
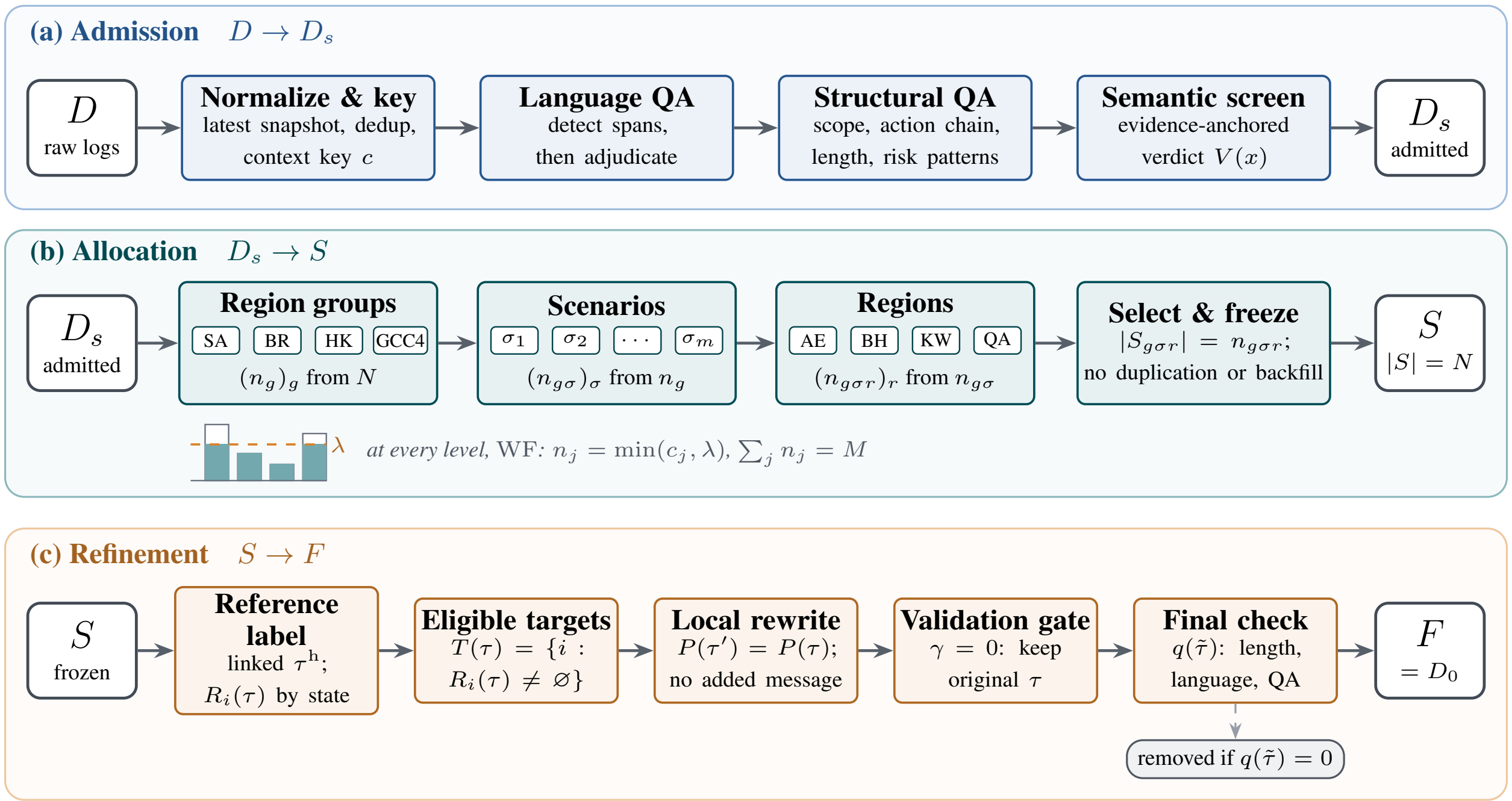
\begin{figure}[t]
\centering
\resizebox{\linewidth}{!}{%
\begin{tikzpicture}[
  font=\footnotesize,
  step/.style={rectangle, rounded corners=2.5pt, draw=#1!80!black, fill=#1!9,
    line width=0.7pt, align=center, inner sep=2pt},
  setnode/.style={rectangle, rounded corners=4pt, draw=careink!85,
    fill=white, line width=0.9pt, align=center, minimum width=1.2cm,
    minimum height=1.05cm, inner sep=2pt},
  chip/.style={rectangle, rounded corners=1.5pt, draw=careteal!70!black,
    fill=white, line width=0.45pt, font=\tiny, minimum width=0.52cm,
    minimum height=0.3cm, inner sep=0.8pt},
  note/.style={rectangle, rounded corners=5pt, draw=#1!75!black, fill=#1!10,
    line width=0.6pt, align=center, font=\scriptsize, minimum height=0.42cm,
    inner sep=2pt},
  flow/.style={-{Stealth[length=2.2mm]}, line width=0.9pt, draw=careink!75},
  drop/.style={-{Stealth[length=1.8mm]}, line width=0.6pt, draw=#1!70,
    dashed},
  lbl/.style={font=\scriptsize\itshape, text=careink!80, align=center,
    inner sep=1.2pt},
  lane/.style={rounded corners=6pt, draw=#1!45, fill=#1!3, line width=0.6pt},
  ltitle/.style={font=\small\bfseries, text=#1!75!black, anchor=west},
]
\def\xin{0.75}\def\xout{15.65}
\def\xa{3.25}\def\xb{6.55}\def\xc{9.85}\def\xd{13.15}

\begin{scope}[on background layer]
\draw[lane=careblue]   (-0.1,-1.62) rectangle (16.5,0.63);
\draw[lane=careteal]   (-0.1,-4.8)  rectangle (16.5,-1.85);
\draw[lane=careorange] (-0.1,-8.15) rectangle (16.5,-5.15);
\end{scope}
\node[ltitle=careblue]   at (0.05,0.33)  {(a) Admission\quad$D\to D_s$};
\node[ltitle=careteal]   at (0.05,-2.1) {(b) Allocation\quad$D_s\to S$};
\node[ltitle=careorange] at (0.05,-5.45) {(c) Refinement\quad$S\to F$};

\def\y{-0.72}
\node[setnode] (aD) at (\xin,\y) {{\large$D$}\\{\scriptsize raw logs}};
\node[step=careblue, text width=2.65cm, minimum height=1.15cm] (a1) at (\xa,\y)
  {\textbf{Normalize \& key}\\{\scriptsize latest snapshot, dedup,\\context key $\ctx$}};
\node[step=careblue, text width=2.65cm, minimum height=1.15cm] (a2) at (\xb,\y)
  {\textbf{Language QA}\\{\scriptsize detect spans,\\then adjudicate}};
\node[step=careblue, text width=2.65cm, minimum height=1.15cm] (a3) at (\xc,\y)
  {\textbf{Structural QA}\\{\scriptsize scope, action chain,\\length, risk patterns}};
\node[step=careblue, text width=2.65cm, minimum height=1.15cm] (a4) at (\xd,\y)
  {\textbf{Semantic screen}\\{\scriptsize evidence-anchored\\verdict $V(x)$}};
\node[setnode] (aDs) at (\xout,\y) {{\large$D_s$}\\{\scriptsize admitted}};
\foreach \u/\v in {aD/a1,a1/a2,a2/a3,a3/a4,a4/aDs} \draw[flow] (\u) -- (\v);

\def\y{-3.1}
\node[setnode] (bDs) at (\xin,\y) {{\large$D_s$}\\{\scriptsize admitted}};
\foreach \x/\n in {\xa/b1,\xb/b2,\xc/b3}
  \node[step=careteal, minimum width=2.85cm, minimum height=1.35cm] (\n) at (\x,\y) {};
\node[step=careteal, text width=2.65cm, minimum height=1.35cm] (b4) at (\xd,\y)
  {\textbf{Select \& freeze}\\{\scriptsize $|S_{g\sigma r}|=n_{g\sigma r}$;\\no duplication or backfill}};
\node[setnode] (bS) at (\xout,\y) {{\large$S$}\\{\scriptsize $|S|=N$}};

\node[font=\footnotesize\bfseries] at (\xa,\y+0.42) {Region groups};
\node[font=\footnotesize\bfseries] at (\xb,\y+0.42) {Scenarios};
\node[font=\footnotesize\bfseries] at (\xc,\y+0.42) {Regions};
\foreach \n [count=\i] in {SA,BR,HK,GCC4}
  \node[chip] at ({\xa+0.68*(\i-2.5)},\y+0.03) {\n};
\foreach \n [count=\i] in {$\sigma_1$,$\sigma_2$,$\cdots$,$\sigma_m$}
  \node[chip] at ({\xb+0.68*(\i-2.5)},\y+0.03) {\n};
\foreach \n [count=\i] in {AE,BH,KW,QA}
  \node[chip] at ({\xc+0.68*(\i-2.5)},\y+0.03) {\n};
\node[font=\scriptsize] at (\xa,\y-0.4) {$(n_g)_g$ from $N$};
\node[font=\scriptsize] at (\xb,\y-0.4) {$(n_{g\sigma})_\sigma$ from $n_g$};
\node[font=\scriptsize] at (\xc,\y-0.4) {$(n_{g\sigma r})_r$ from $n_{g\sigma}$};
\foreach \u/\v in {bDs/b1,b1/b2,b2/b3,b3/b4,b4/bS} \draw[flow] (\u) -- (\v);

\def\base{-4.62}
\foreach \c [count=\i] in {0.62,0.3,0.18,0.52} {
  \pgfmathsetmacro{\bx}{1.75+0.36*\i}
  \draw[draw=caregray, line width=0.45pt, fill=white] (\bx,\base) rectangle ++(0.26,\c);
  \pgfmathsetmacro{\fh}{min(\c,0.4)}
  \fill[careteal!55] (\bx,\base) rectangle ++(0.26,\fh);
}
\draw[caregray, line width=0.45pt] (1.95,\base) -- (3.45,\base);
\draw[careorange, line width=0.7pt, dashed] (1.95,\base+0.4) -- (3.45,\base+0.4)
  node[right, font=\scriptsize, text=careorange!80!black, inner sep=1pt] {$\lambda$};
\node[lbl, anchor=west, align=left] at (3.85,\base+0.3)
  {at every level, $\mathrm{WF}$: $n_j=\min(c_j,\lambda)$, $\sum_j n_j=M$};

\def\y{-6.5}
\node[setnode] (cS) at (\xin,\y) {{\large$S$}\\{\scriptsize frozen}};
\foreach \x/\n/\txt in {
  2.9/c1/{\textbf{Reference label}\\{\scriptsize linked $\tau^{\mathrm h}$;\\$R_i(\tau)$ by state}},
  5.55/c2/{\textbf{Eligible targets}\\{\scriptsize $T(\tau)=\{i:$\\$R_i(\tau)\neq\varnothing\}$}},
  8.2/c3/{\textbf{Local rewrite}\\{\scriptsize $P(\tau')=P(\tau)$;\\no added message}},
  10.85/c4/{\textbf{Validation gate}\\{\scriptsize $\gamma=0$: keep\\original $\tau$}},
  13.5/c5/{\textbf{Final check}\\{\scriptsize $q(\tilde\tau)$: length,\\language, QA}}}
  \node[step=careorange, text width=2.1cm, minimum height=1.15cm] (\n) at (\x,\y) {\txt};
\node[setnode] (cF) at (\xout,\y) {{\large$F$}\\{\scriptsize $=D_0$}};
\foreach \u/\v in {cS/c1,c1/c2,c2/c3,c3/c4,c4/c5,c5/cF} \draw[flow] (\u) -- (\v);

\node[note=caregray, minimum width=2.4cm] (removed) at (13.5,-7.7)
  {removed if $q(\tilde\tau)=0$};
\draw[drop=caregray] (c5.south) -- (removed.north);
\end{tikzpicture}}
\caption{The cold-start data engine. (a) Admission keeps complete,
context-keyed conversations that pass language, structural, and
evidence-anchored semantic checks; only \texttt{no\_problem} conversations
enter $D_s$, and rejected or unresolved records are never backfilled.
(b) Allocation splits a fixed budget $N$ over region groups, the scenarios of
each group, and the regions of GCC4, applying the capacity-constrained equal
split of Eq.~\ref{eq:waterfill} at every level, and then freezes the selected
members $S$. (c) Refinement borrows tone from linked human conversations for
eligible turns while preserving every protected field; a rejected rewrite
keeps the original conversation, and only a failed final check removes it.}
\label{fig:data_engine}
\end{figure}

%% file: Technical_Report/sections/05_retrospection.tex
\section{Retrospection: Online Inspection and Attribution}
\label{sec:retrospection}

Retrospection turns the deployed model's own production conversations into
failure evidence that is reliable enough to learn from (requirement R1), and
decides whose failure each one is before any of them can influence training
(R2). It runs continuously, around the clock, on the traffic served by the
in-house model. Its screening, analysis, review, and attribution steps are run by
an external frontier model rather than by the in-house model
(Section~\ref{sec:loop_scope}), so the deployed model never judges its own
conversations. It is deliberately separate from the evaluation system of
Section~\ref{sec:evaluation}: online inspection discovers failures to learn
from, whereas the evaluation system measures candidate versions on fixed test
sets. The two use the same inspection procedure but never share cases or
denominators.

\subsection{Continuous online inspection}
\label{sec:inspection}
Online inspection examines the production conversations handled by the
in-house model as they mature, 24 hours a day. Because no verifier exists,
inspection is organized as a sequence of separable decisions rather than a
single judge score (Figure~\ref{fig:retrospection}). Three separations run
through the design: discovering a suspicious conversation is separate from
confirming a Bad Case, a model's judgment is separate from the act of
publishing it, and local evidence is separate from production writes.

\input{Technical_Report/figs/retrospection}

\paragraph{Freezing and evidence blinding.}
Each inspection run first freezes its input: the source window or report, the
case identities, and the evidence view. We write $\mathcal Q_t$ for the frozen
case set of window $t$; online it holds the mature conversations served by
$\pi_{\theta_k}$ in that window, and offline it is a fixed test set. Every case
$x\in\mathcal Q_t$ carries an evidence view $\xi(x)$ (dialogue, tool calls and
results, order state, trajectory, and the applicable policy knowledge) and
metadata $\mu(x)$ (source quality labels and scores, report and model
identities, experiment labels, and release-side fields). The metadata are kept
for auditing but never enter a judgment prompt, so every judgment stage $h$
(screening, analysis, review) is counterfactually invariant to them:
\begin{equation}
  h\bigl(\xi(x),\mu\bigr)=h\bigl(\xi(x),\mu'\bigr)
  \qquad\text{for all metadata }\mu,\mu'.
  \label{eq:blinding}
\end{equation}
Scenario identity is taken from stable fields (test contract, region, business
group, scenario definition) rather than from the model's reply. Without
Eq.~\eqref{eq:blinding}, a judge that knew the source system had failed a trace,
or which version produced it, could learn a label shortcut instead of judging
the behavior.

\paragraph{Screening as evidence grading.}
A screening model compresses $\mathcal Q_t$ into a ranked candidate pool. It is
an evidence grader and ranker, not the final judge. For each case it returns a
suspicion score $s(x)\in[0,1]$ together with an evidence profile: whether the
defect is visible, the number of locatable evidence anchors $a(x)$, the
likelihood $u(x)$ that the case survives to publication, and the degree
$\rho(x)$ to which the verdict depends on runtime evidence missing from the
log. Only cases with strong evidence are eligible for high-confidence
admission,
\begin{equation}
  \mathcal Q^{\mathrm{el}}_t=\bigl\{x\in\mathcal Q_t:\;
  \text{defect visible},\;a(x)\ge a_{\min},\;u(x)\ge u_{\min},\;
  \rho(x)\le\rho_{\max}\bigr\}.
  \label{eq:eligible}
\end{equation}
Weak signals such as low satisfaction, long dialogue, or repeated requests for
a human may raise $s(x)$ but cannot establish eligibility on their own. A case
on which screening fails is not routed onward by any fallback, so an
infrastructure fault is never converted into an apparent business risk.

\paragraph{Capacity-aware admission.}
Deep analysis, review, and gating are expensive, and the choice of which cases
receive them shapes what the system can conclude. Admission therefore fills a
rolling capacity $B_t$ from three strata,
\begin{equation}
  \mathcal Q^{\mathrm{adm}}_t=
  \mathcal Q^{\mathrm{hi}}_t\cup\mathcal Q^{\mathrm{bd}}_t\cup
  \mathcal Q^{\mathrm{ex}}_t,\qquad
  \bigl|\mathcal Q^{\mathrm{adm}}_t\bigr|\le B_t,
  \label{eq:admission}
\end{equation}
where $\mathcal Q^{\mathrm{hi}}_t$ takes the highest-scoring cases of
$\mathcal Q^{\mathrm{el}}_t$ to maximize immediate discovery,
$\mathcal Q^{\mathrm{bd}}_t$ samples the boundary band
$\{x:|s(x)-\eta|\le\beta\}$ around the screening threshold $\eta$ to measure
where the screener is uncertain, and $\mathcal Q^{\mathrm{ex}}_t$ samples
low-score or rare strata to keep new failure families discoverable. Pure
top-$K$ admission would entrench the screener's historical bias, and pure
random admission would spend expensive capacity on uninformative cases. Every
admitted case is committed to reach a terminal state; a suspicious case outside
$\mathcal Q^{\mathrm{adm}}_t$ carries no such commitment.

\paragraph{Deep analysis.}
Each admitted case receives a structured case record that answers six
questions: what the user actually requested; which intent, FAQ, Skill, or SOP
path the system selected; which actions or promises the agent made at the key
turns; whether those violated regional, business-line, compensation,
risk-control, or language rules; whether the runtime evidence suffices to
support that judgment; and, if a defect exists, its type, likely cause, and a
suggested fix. A claim ledger links every conclusion of the record to anchors
in $\xi(x)$, and mechanical contract checks verify the record's structure.
These auxiliary checks improve consistency but never short-circuit the
semantic review and gate.

\paragraph{Review, reconciliation, and gate.}
An independent reviewer, distinct from the analyst, challenges the record
against the policy knowledge base and returns a verdict
$\omega(x)\in\{\textsc{bad},\textsc{expected},\textsc{inconclusive}\}$. Evidence
gaps, malformed records, or conflicting conclusions send the record back for
reconciliation before the verdict is final. A publication gate $G(x)\in\{0,1\}$
then requires a unique anomalous turn, a complete evidence package, a valid
problem classification, and an explicit verdict. The terminal state
$\psi(x)$ of an admitted case is
\begin{equation}
  \psi(x)=
  \begin{cases}
    \textsc{bad} & \text{if } \omega(x)=\textsc{bad}\text{ and }G(x)=1,\\
    \textsc{expected} & \text{if } \omega(x)=\textsc{expected},\\
    \textsc{fail} & \text{if review or infrastructure fails},\\
    \textsc{inconclusive} & \text{otherwise}.
  \end{cases}
  \label{eq:terminal}
\end{equation}
The rule is fail-closed: a case that cannot be confirmed is recorded as
inconclusive or sent for manual review, and a review or gate pass is never
assumed. Only $\psi(x)=\textsc{bad}$ can reach learning.

\paragraph{Audit and the confirmed-bad rate.}
An audit checks that every admitted case reached a terminal state, that every
confirmed case holds both a review and a gate pass, that expected and
inconclusive verdicts are backed by an independent review, that no source label
leaked into a prompt, and that denominators are consistent. With
$\mathcal B_t=\{x\in\mathcal Q^{\mathrm{adm}}_t:\psi(x)=\textsc{bad}\}$, the
headline metric is the confirmed-bad rate over the complete frozen set,
\begin{equation}
  \widehat R_t=\frac{|\mathcal B_t|}{|\mathcal Q_t|}.
  \label{eq:risk}
\end{equation}
Cases that were never admitted, and inconclusive cases, count neither as
failures nor as passes. If confirmation is precise, that is,
$\mathcal B_t\subseteq\mathcal B^{\star}_t$ for the set $\mathcal B^{\star}_t$
of truly defective cases, then
$\widehat R_t\le|\mathcal B^{\star}_t|/|\mathcal Q_t|$: $\widehat R_t$ is a
conservative lower bound on the defect rate rather than a calibrated estimate
of it. Online, it is a monitoring signal over shifting traffic; versions are
compared instead with the offline inspection test of
Section~\ref{sec:offline_inspection}, which applies the same procedure to a
fixed case set.

\paragraph{Feedback to the screener.}
Terminal states also recalibrate the inspection system itself. Writing
$\mathcal F^{+}_t$ and $\mathcal F^{-}_t$ for the screener's positive and
negative examples,
\begin{equation}
  \mathcal F^{+}_{t+1}=\mathcal F^{+}_t\cup\mathcal B_t,\qquad
  \mathcal F^{-}_{t+1}=\mathcal F^{-}_t\cup
  \{x\in\mathcal Q^{\mathrm{adm}}_t:\psi(x)=\textsc{expected}\}.
  \label{eq:feedback}
\end{equation}
Inconclusive and failed cases enter neither set: a review failure is
semantically undecided, and an infrastructure failure says nothing about model
quality, so a broken evaluator is never learned as evidence of safe behavior.
The boundary and exploration strata of Eq.~\eqref{eq:admission} supply the
cases from which new failure families enter $\mathcal F^{+}$. Inspection modes
and the stage contract are summarized in Appendix~\ref{app:inspection}, and
Appendix~\ref{app:inspection_cases} follows two production conversations
through the procedure, one judged expected behavior and one confirmed as a Bad
Case.

\subsection{Failure attribution}
\label{sec:attribution}
A confirmed failure says that something went wrong, not that the model should
change. Attribution assigns responsibility before learning. It assembles an
evidence package for each confirmed case: the complete trajectory, the executed
tool results, the Skill/SOP, prompt, and tool definitions the model was given
and their recent changes, the training coverage of similar situations, and the
version of the judge involved. Candidate causes and their repair routes
are listed in Table~\ref{tab:attribution}.

\begin{table}[t]
\caption{Attribution categories and repair routes. Only the model-behavior
category produces training data for Phase~II; every other category is repaired
outside the model.}
\label{tab:attribution}
\centering
\small
\begin{tabularx}{\linewidth}{@{}l Y Y@{}}
\toprule
Cause & Evidence sought & Route \\
\midrule
Model behavior & Same context and valid tools; the decision error is reproducible, including in situations that training covered poorly & Incremental task and attribution-guided OPSD \\
Skill/SOP or prompt & The policy or prompt itself is wrong, ambiguous, or inconsistent & Revise the Skill/SOP or prompt \\
Tool, harness, or router & Schema, dispatch, projection, or route mismatch & Runtime repair \\
Judge or inspection & Unsupported verdict or review disagreement & Recalibrate the evaluator; discard the label \\
Infrastructure & Timeout, missing artifact, failed write & Operational recovery; no quality label \\
\bottomrule
\end{tabularx}
\end{table}

Co-occurrence is not causation: a concurrent Skill change does not by itself
make a failure the Skill's fault, and the absence of similar examples from
training does not by itself explain a model error. Where feasible, attribution performs controlled replay that varies
the model, the Skill/SOP and prompt, and the tool or harness contract
independently, and it retains a multi-cause or unknown label when the cause
cannot be isolated. This conservatism prevents the loop from teaching the model
to work around a broken tool or a miscalibrated judge, which would be a
self-confirming failure of exactly the kind RSI must avoid.

For a model-attributable case, attribution emits a diagnosis $d$ alongside the
route. The diagnosis records the supported cause, including whether similar
situations were scarce in training, the policy and context under
which the correct behavior is defined, the evidence anchors, the decision point
at which the model went wrong, and corrective guidance. Recording the decision
point matters for learning: the guidance must be actionable with the
information available at that point, so that the model is not taught to assert
outcomes it could not have known. Because the correct action depends on the
SOP in force, the diagnosis is defined relative to the Skill/SOP the model was
given when the failure occurred.

%% file: Technical_Report/figs/retrospection.tex
\begin{figure}[t]
\centering
\resizebox{\linewidth}{!}{%
\begin{tikzpicture}[
  font=\footnotesize,
  src/.style={rectangle, rounded corners=2.5pt, draw=careblue!80!black,
    fill=careblue!9, line width=0.7pt, align=center, text width=1.95cm,
    minimum height=1.5cm, inner sep=2pt},
  stage/.style={rectangle, rounded corners=2.5pt, draw=careorange!80!black,
    fill=careorange!#1, line width=0.7pt, align=center, text width=1.95cm,
    minimum height=1.5cm, inner sep=2pt},
  chip/.style={rectangle, rounded corners=1.5pt, draw=#1!80!black, fill=#1!18,
    line width=0.5pt, font=\tiny, minimum width=0.5cm, minimum height=0.3cm,
    inner sep=0.5pt},
  pill/.style={rectangle, rounded corners=6pt, draw=#1!80!black, fill=#1!12,
    line width=0.6pt, align=center, minimum height=0.6cm, inner sep=3pt,
    font=\scriptsize},
  route/.style={rectangle, rounded corners=2.5pt, draw=#1!80!black, fill=#1!9,
    line width=0.7pt, align=center, text width=3.5cm, minimum height=0.95cm,
    inner sep=2pt},
  flow/.style={-{Stealth[length=2.2mm]}, line width=0.9pt, draw=careink!75},
  back/.style={-{Stealth[length=2mm]}, line width=0.8pt, draw=caregray,
    dashed},
  bus/.style={line width=0.8pt, draw=caregray},
  lbl/.style={font=\scriptsize\itshape, text=careink!80, align=center,
    inner sep=1.2pt},
]
\fill[rounded corners=5pt, caregray!9] (3.55,-0.98) rectangle (13.5,1.62);
\node[lbl, anchor=north west, font=\scriptsize] at (3.65,1.56)
  {\textbf{evidence-blinded:} $h(\xi(x),\mu)=h(\xi(x),\mu')$};

\node[src] (prod) at (-0.4,0)
  {\textbf{Production}\\{\scriptsize conversations of\\$\pi_{\theta_k}$, 24/7}};
\node[stage=6] (freeze) at (2.15,0)
  {\textbf{Freeze}\\{\scriptsize cases $\mathcal Q_t$,\\evidence $\xi(x)$}};
\node[stage=9] (screen) at (4.7,0)
  {\textbf{Screen}\\{\scriptsize score $s(x)$,\\evidence profile}};
\node[stage=12] (admit) at (7.25,0) {};
\node[font=\footnotesize\bfseries] at (7.25,0.42) {Admit};
\node[font=\scriptsize] at (7.25,0.04) {$|\mathcal Q^{\mathrm{adm}}_t|\le B_t$};
\node[chip=carered]    at (6.65,-0.42) {hi};
\node[chip=careorange] at (7.25,-0.42) {bd};
\node[chip=careblue]   at (7.85,-0.42) {ex};
\node[stage=15] (deep) at (9.8,0)
  {\textbf{Deep analysis}\\{\scriptsize case record,\\claim ledger}};
\node[stage=18] (review) at (12.35,0)
  {\textbf{Review}\\{\scriptsize independent\\verdict $\omega(x)$}};
\node[stage=24] (audit) at (14.9,0)
  {\textbf{Gate \& audit}\\{\scriptsize gate $G(x)$,\\terminal $\psi(x)$}};

\draw[flow] (prod) -- (freeze);
\foreach \a/\b in {freeze/screen,screen/admit,admit/deep,deep/review,review/audit}
  \draw[flow] (\a) -- (\b);

\draw[back] (review.north) to[out=125, in=55]
  node[midway, above=2pt, lbl] {reconcile} (deep.north);

\node[pill=carered]    (bad)   at (14.9,-1.85)  {confirmed bad};
\node[pill=caregreen]  (exp)   at (12.35,-1.85) {expected};
\node[pill=caregray]   (inc)   at (9.8,-1.85)   {inconclusive};
\node[pill=carepurple] (infra) at (7.25,-1.85)  {review / infra failure};
\draw[bus] (audit.south) -- (14.9,-1.3);
\draw[bus] (14.9,-1.3) -- (7.25,-1.3);
\draw[flow] (14.9,-1.3) -- (bad.north);
\foreach \n in {exp,inc,infra}
  \draw[flow, draw=caregray] (\n.north |- 0,-1.3) -- (\n.north);
\node[lbl] at (9.8,-2.42) {not routed to model learning};

\draw[back] (7.25,-1.3) -- (4.7,-1.3) -- (screen.south);
\node[lbl, anchor=north east] at (4.55,-1.25)
  {feedback to screener\\$\mathcal F^{+}\!\leftarrow$ bad,
   $\mathcal F^{-}\!\leftarrow$ expected};

\node[rectangle, rounded corners=3pt, draw=careorange!80!black, fill=careorange!20,
  line width=0.9pt, align=center, text width=3.0cm, minimum height=1.25cm]
  (attr) at (14.9,-4.15)
  {\textbf{Attribution}\\[1pt]\scriptsize controlled replay,\\context and tool diffs};
\draw[flow] (bad) -- (attr);
\node[route=careteal] (model) at (10.35,-3.45)
  {\textbf{model-attributable}\\{\scriptsize $(x,\ctx,d)\to\Delta_k$, Section~\ref{sec:evolve}}};
\node[route=carered] (comp) at (10.35,-4.85)
  {\textbf{component}\\{\scriptsize Skill/SOP, tool, harness, judge}};
\node[route=caregray, text width=2.9cm] (unk) at (5.65,-4.15)
  {\textbf{multi-cause / unknown}\\{\scriptsize held for more evidence}};
\draw[flow] (attr.west) -- ++(-0.5,0) |- (model.east);
\draw[flow] (attr.west) -- ++(-0.5,0) |- (comp.east);
\draw[flow, draw=caregray] (attr.west) -- (unk.east);
\end{tikzpicture}}
\caption{Online inspection and attribution. Production conversations of the
in-house model are frozen into a case set $\mathcal Q_t$ and pass an
evidence-blinded inspection (Eq.~\ref{eq:blinding}) that separates ranking,
capacity-aware admission from three strata (Eq.~\ref{eq:admission}), analysis,
and independent review with reconciliation, and ends in one of four terminal
states (Eq.~\ref{eq:terminal}). Terminal states feed back to the screener
(Eq.~\ref{eq:feedback}). Only confirmed failures reach attribution, which routes
model-attributable cases, together with their diagnosis $d$, to incremental task
construction, routes component failures to repair, and holds cases whose cause
cannot be isolated.}
\label{fig:retrospection}
\end{figure}
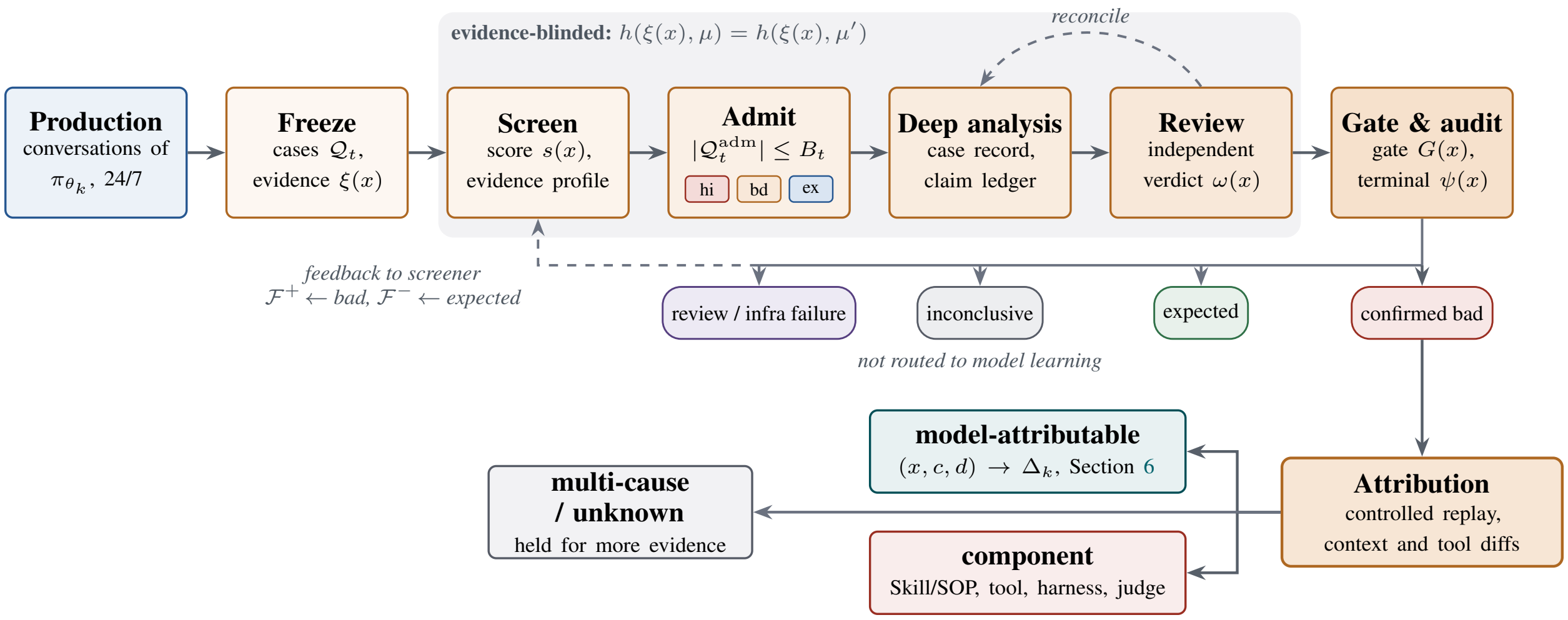

%% file: Technical_Report/sections/06_evolve.tex
\section{Phase II: Evolve by Attribution-Guided On-Policy Self-Distillation}
\label{sec:evolve}

The evolve stage converts model-attributable failures into a weight update
(requirement R3). It has three parts. \emph{Retrospective repair construction}
turns each failure into a verified repair at the first decision where the model
went wrong. \emph{On-policy self-distillation} trains the model on its own
outputs at that decision against a teacher that is the model itself, given the
hindsight as privileged information. \emph{Data selection} balances the training
signal and deliberately mixes policy versions (R4). The resulting candidate is
then evaluated and promoted as described in Section~\ref{sec:evaluation};
Algorithm~\ref{alg:care} summarizes one full iteration.

\subsection{From failures to verified repair tuples}
\label{sec:construct}
A confirmed failure arrives as a long multi-turn trajectory, logged across
several streams (the requests sent to the model, the agent's outputs and tool
calls, state persistence, and what the user saw), together with a retrospective
diagnosis. It cannot be used for training as it stands, because it records the
whole conversation, whereas the model made each decision from only what
preceded it. Everything from the first error onward is therefore hindsight: the
later turns are conditioned on a state that a corrected policy would never
reach, so their correction is ill-defined, and the outcome and inspection that
exposed the failure were not available when the model erred. Construction
accordingly works at the single decision where the model first went wrong, and
separates the input the model had there from what was learned afterwards
(Figure~\ref{fig:construction}).

Let $h_t$ denote the exact input of the model at step $t$: the message history,
the tool schemas, the Skill content, and the generation settings of that request.
It is the agent's information set at the moment of decision. Everything that
retrospection adds beyond $h_t$ is \emph{hindsight}.

\input{Technical_Report/figs/construction}

\paragraph{Step 1: localize the first model error.}
Let $\mathcal A^{\star}(h_t,\ctx)$ be the set of acceptable actions at step $t$
under context $\ctx$. Construction anchors the earliest defective decision,
\begin{equation}
  t^{\star}=\min\bigl\{t:\;a_t\notin\mathcal A^{\star}(h_t,\ctx)\bigr\},
  \label{eq:first_error}
\end{equation}
and proceeds only if attribution assigns this defect to the model. If the
earliest defect belongs to a tool, a state resolver, or an upstream component,
the case is routed to component repair even when later model errors exist, since
those later decisions were made on a corrupted state. The anchor is identified
by the request identity and the exact logged output rather than by ordinal
position, and a case without a unique anchor is held as pending.

\paragraph{Step 2: causal truncation.}
The training input is the exact request at the anchored decision, $x=h_{t^\star}$:
the message history up to but excluding $a_{t^\star}$, with every other request
field kept verbatim. The erroneous output $a^{-}=a_{t^\star}$, everything after
it, the outcome $z$, and the diagnosis $d$ form the hindsight
$\mathcal H=(a^{-},h_{>t^\star},z,d)$, which may enter training only on the
teacher side.

\paragraph{Step 3: minimal counterfactual repair.}
A proposer model receives $x$, $a^{-}$, the incident-time policy, and the
diagnosis $d$ (marked as invisible to the student), and proposes a replacement
$a^{+}$ for the single erroneous decision. Writing $\mathcal M(x)$ for the
actions whose asserted facts are supported by $x$ alone, the proposer
approximates the smallest admissible edit,
\begin{equation}
  a^{+}\in\arg\min_{a\,\in\,\mathcal A^{\star}(x,\ctx)\,\cap\,\mathcal M(x)}
  \;\delta\bigl(a,a^{-}\bigr),
  \label{eq:repair}
\end{equation}
where $\delta$ measures the extent of the change. Four constraints make this
precise. \emph{Locality}: only the erroneous decision changes; the prefix is
identical, and nothing after the decision (tool results, user replies,
follow-up messages) is generated. \emph{Feasibility}: $a^{+}$ uses only the tools,
arguments, and response fields available in $x$ under context $\ctx$.
\emph{Non-anticipation}: hindsight may tell the proposer \emph{that} $a^{-}$ was
wrong, but it may not make $a^{+}$ more specific than $x$ allows. For example, if
a status tool returned only ``not picked up'' and the original reply asserted
that the food was still being prepared, the repair may remove the unsupported
claim but may not substitute an order status observed later. \emph{Compliance}:
$a^{+}$ follows the policy in force at the time of the incident, not the current
one. When $x$ contains images, the proposer must consume the images themselves.

\paragraph{Step 4: independent verification.}
Deterministic checks first confirm prefix identity, a single in-place
difference, and schema validity of tool calls and response fields. An
independent verifier then judges $(x,a^{-},a^{+},d)$ along separate axes: whether
the target is a genuine model fault, whether the inspected symptom is corrected,
whether the repair stays at the decision boundary, whether it complies with the
incident-time policy, and whether it respects the tool and format contract.
A repair is accepted only if every axis passes,
$\mathrm{accept}=\bigwedge_{j}V_j(x,a^{-},a^{+},d)$. Acceptance certifies that
$a^{+}$ is an acceptable decision at $x$; it does not certify that a tool call
would succeed or that the conversation would end well.

\paragraph{Repair tuples and their training views.}
Accepted cases form the incremental set
\begin{equation}
  \Delta_k=\bigl\{(x_i,\,\ctx_i,\,a^{-}_i,\,a^{+}_i,\,d_i,\,\ell_i)\bigr\},
  \label{eq:delta}
\end{equation}
where $\ell_i$ links each tuple to its source case. One tuple supports three
training views (Table~\ref{tab:views}), which we compare in
Section~\ref{sec:exp_objective}. Supervised fine-tuning imitates $a^{+}$, and direct
preference optimization~\citep{dpo} contrasts $a^{+}$ with $a^{-}$; both treat a
text written by the proposer as the target. OPSD instead keeps the student on
its own outputs and lets the reference reach it only through the teacher.
The construction resembles hindsight relabeling~\citep{her} in that it reuses
information that becomes available only after the episode, but the hindsight is
used to correct the decision under the original goal, and it never reaches the
student.

\begin{table}[t]
\caption{Training views of a repair tuple $(x,\ctx,a^{-},a^{+},d)$.
$I=\mathrm{enc}(d,a^{-},a^{+})$ is the privileged information.}
\label{tab:views}
\centering
\small
\begin{tabularx}{\linewidth}{@{}l l Y@{}}
\toprule
View & Student input & Learning signal \\
\midrule
SFT & $x$ & imitate the reference repair $a^{+}$ \\
DPO & $x$ & prefer $a^{+}$ over $a^{-}$ \\
OPSD & $x$ & match a self-teacher that sees $(x,I)$, on the student's own sampled output \\
\bottomrule
\end{tabularx}
\end{table}

\paragraph{Accounting.}
Every case ends in exactly one terminal state: \emph{admitted}, \emph{routed}
(the first defect is not the model's), \emph{rejected} by the verifier, or
\emph{pending} (no unique anchor, a failed construction, or unverified visual
evidence). Pending cases are never relabeled as rejected or routed, and cases
that are not admitted are not counted as failures the model cannot learn.
Before a tuple is used for training, four further checks apply: the incident
model is replayed at $x$ to test whether $a^{-}$ reproduces; the tuple is rendered
with the target tokenizer, chat template, tool schemas, and loss mask; the
collator is verified to keep $I$ out of the student input; and a minimal training
run is executed. Implementation details are given in
Appendix~\ref{app:construction}.

\subsection{Hindsight as privileged information}
\label{sec:opsd}
For a tuple $(x,\ctx,a^{-},a^{+},d)\in\Delta_k$, the student samples its own
decision at the anchored point, $\hat y\sim\pith(\cdot\mid x)$: a single
assistant turn, which may contain a tool call; $x$ already encodes the context
$\ctx$. A \emph{self-teacher}, a frozen copy $\bar\theta_k$ of the model at the
start of the iteration, reads the same input with the privileged information
$I=\mathrm{enc}(d,a^{-},a^{+})$ added to its system message, which states what
went wrong, the SOP in force at the incident, and the verified repair. The
teacher then follows the student's rollout token by token
(Figure~\ref{fig:opsd}), so that at each position both models give a
distribution over the whole vocabulary from the same prefix,
\begin{equation}
  p_{\theta,t}=\pith\bigl(\cdot\mid x,\hat y_{<t}\bigr),
  \qquad
  q_t=\sg\!\left[\pi_{\bar\theta_k}\bigl(\cdot\mid x,I,\hat y_{<t}\bigr)\right],
  \label{eq:opsd_dists}
\end{equation}
where $\sg$ denotes stop-gradient. The objective accumulates a divergence
$\ell_t$ between the two over the student's own samples,
\begin{equation}
  \mathcal L_{\mathrm{OPSD}}(\theta)=
  \E_{(x,I)\sim\Delta_k}\;
  \E_{\hat y\sim\pith(\cdot\mid x)}
  \Bigl[\sum_{t} m_t\,\ell_t(\theta)\Bigr],
  \label{eq:opsd}
\end{equation}
where $m_t$ selects the generated tokens of the student's turn and excludes the
prompt and padding. The rollout fixes the prefixes at which the two models are
compared and is then treated as data: the loss is backpropagated directly
through $p_{\theta,t}$, without a policy-gradient estimator over the sampling
step.

\paragraph{A family of truncated KL losses.}
We define $\ell_t$ as a Kullback--Leibler (KL) divergence restricted to the
tokens that matter to the teacher. Let $\mathcal S^{K}_t$ be the $K$ tokens to
which $q_t$ assigns the highest probability. The family leaves three choices
open: the direction of the divergence, the size $K$, and a floor on negative
token losses. Section~\ref{sec:exp_loss} ablates all three. The configuration
that we deploy uses the forward direction with $K=32$ and the floor,
\begin{equation}
  u^{K}_t(\theta)=\sum_{v\in\mathcal S^{K}_t} q_t(v)\,
  \bigl[\log q_t(v)-\log p_{\theta,t}(v)\bigr],
  \qquad
  \ell_t(\theta)=\max\bigl(u^{K}_t(\theta),\,0\bigr).
  \label{eq:opsd_forward}
\end{equation}
Both distributions keep their full-vocabulary normalization: the $K$ terms are
not renormalized, and no smoothing or lower bound is added to the probabilities.

\emph{Direction.} The forward direction weights each token by the teacher's
probability. For $u^{K}_t>0$ its gradient is
$-\sum_{v\in\mathcal S^{K}_t}q_t(v)\,\nabla_\theta\log p_{\theta,t}(v)$, a soft
cross-entropy: the student is pulled toward every token that the informed
teacher finds plausible, in proportion to the teacher's confidence, instead of
toward one fixed token. The alternative is the reverse direction over
$\mathcal S^{K}_t$ and one cell that merges all remaining tokens,
\begin{equation}
  \ell^{\mathrm{rev}}_t(\theta)=
  \sum_{c\,\in\,\mathcal S^{K}_t\cup\{\mathrm{tail}\}}
  p_{\theta,t}(c)\,\bigl[\log p_{\theta,t}(c)-\log q_t(c)\bigr],
  \qquad
  r(\mathrm{tail})=\sum_{v\notin\mathcal S^{K}_t}r(v)
  \;\text{ for }r\in\{p_{\theta,t},q_t\},
  \label{eq:opsd_reverse}
\end{equation}
which is a proper KL divergence between the two distributions coarsened to
$K+1$ cells, and hence nonnegative. It is mode-seeking: it penalizes the student
for placing mass where the teacher does not, rather than for missing mass where
the teacher places it.

\emph{Floor.} Without a tail cell, $u^{K}_t$ is only a partial sum of the forward
KL, and it can be negative although the full divergence cannot. By the log-sum
inequality, $u^{K}_t\ge Q_t\log(Q_t/P_t)$, where $Q_t$ and $P_t$ are the total
probabilities that the teacher and the student place on $\mathcal S^{K}_t$, so
it is negative only when the student already places more mass on the set than
the teacher does. Driving it lower would only concentrate the student further:
the minimizer of $u^{K}_t$ is $q_t/Q_t$ on the set, with zero probability on
every other token. The floor stops the update of such a token at zero. It acts
on the sum at each token, before masking and averaging, not on individual terms
or on the batch mean, and it bounds a loss, not a probability.

\emph{Batch loss.} Over a batch, the loss is the mean over all generated tokens,
\begin{equation}
  \hat{\mathcal L}(\theta)=\frac{1}{N_{\mathrm{tok}}}
  \sum_{i,t}m_{i,t}\,\ell_{i,t}(\theta),
  \qquad N_{\mathrm{tok}}=\sum_{i,t}m_{i,t},
  \label{eq:opsd_batch}
\end{equation}
so every generated token of the batch, tool-call syntax included, has equal
weight. No task reward, policy-gradient term, or auxiliary SFT or DPO loss
enters the update, and the forward objective needs from the teacher only its $K$
most likely tokens and their probabilities at each position. The configuration
is listed in Appendix~\ref{app:opsd_config}.

\input{Technical_Report/figs/opsd}

Three properties make this objective a mechanism for grounded self-improvement
rather than for compression.

\paragraph{On-policy at the failing decision.}
The model is corrected on what it would do now at the decision where it failed,
not on a text written by someone else. A fresh sample is drawn for every update,
so the same tuple yields different student turns in different epochs, and the
data never contain a fixed target sequence. Supervised fine-tuning on $a^{+}$ is
off-policy: it imitates the surface form of the proposer's output, which may be
unlikely under the model's own distribution. Preference optimization contrasts
two fixed outputs, neither of which need be what the current model produces.
OPSD defines its targets on the student's own sample, so the signal stays
relevant as the model changes, in line with on-policy distillation more
broadly~\citep{dagger,gkd}.

\paragraph{A teacher that is the model itself.}
The teacher starts from the same weights as the student, so its targets lie close
to the student's distribution and the update concentrates on the decisions that
the privileged information changes. Keeping it fixed during the iteration,
rather than tracking the student as a moving average, anchors the signal: the
student cannot move its own target. This anchoring is what makes a small
increment usable. The target at every token is the model's own distribution, so
tokens that the privileged information does not affect carry little loss,
whereas a fixed target text has to be fitted everywhere; on increments of this
size SFT and DPO collapse (Section~\ref{sec:exp_objective}). The teacher's only
advantage over the student is $I$, in line with the intuition that a capable
model can rationalize verified privileged information and teach its weaker,
uninformed self~\citep{opsd_reasoner,privileged}. In our implementation the
repair proposer is a frontier model (GPT-5.6-Sol). Its output reaches training
only as a verified reference inside $I$: the student never sees it and is never
trained to copy its wording. The proposer is a replaceable component, and
letting the in-house model propose its own repairs is a natural step toward a
more closed loop (Section~\ref{sec:discussion}).

\paragraph{Grounded privileged information.}
The privileged information is neither an unverified solution nor the model's own
critique. The failure is established by retrospection, its cause by attribution,
its location by first-error localization, and the reference repair by an
independent verifier; only the articulation of the corrected behavior in the
student's own distribution is left to the model. A self-critique loop would let
the model's biases define both the problem and its correction.

The privileged information never crosses the privileged boundary. It is not
appended to student inputs, production requests, or held-out evaluation, and the
training data loader is checked to keep it out of the student context, so
measured improvement reflects what the model has internalized. Because
construction enforces non-anticipation, the teacher's targets remain actionable
from $x$ alone, and because the teacher scores the student's own tokens, a
different but acceptable decision is not forced into the wording of $a^{+}$. The
student's sample is a single turn: a tool call it produces is scored but not
executed. Letting the student continue the conversation in the simulator after
the repaired decision is a natural generalization that we leave to future work.

\paragraph{Why not reinforcement learning from outcome rewards.}
\label{sec:why_not_rl}
A natural alternative is to optimize the agent against an outcome reward, for
example with group-relative policy optimization (GRPO)~\citep{grpo}. We
deliberately do not, because no available reward is faithful to the production
objective. \care is ultimately judged by user satisfaction, and satisfaction is
a poor training reward: most conversations are never rated, ratings arrive only
at the end of a conversation, and they are confounded by factors outside the
agent's decisions, such as courier delays, refund policy, and the food itself.
They also cannot be computed for a replayed or simulated decision. Rewards that
can be computed are either unrelated to satisfaction, such as format or
tool-validity checks, or produced by LLM judges, which an optimizer can
exploit~\citep{judgeguard}; rubric rewards~\citep{rubrics} make the criteria
explicit but would still have to be calibrated against satisfaction. OPSD avoids
defining a scalar reward altogether. Its token-level signal is the divergence
from a teacher that sees a verified repair, so learning is grounded in
reviewed evidence about specific decisions rather than in a proxy for
satisfaction.

\subsection{Data selection and mixing across policy versions}
\label{sec:replay}
Production failures are unevenly distributed: a few micro-scenarios, such as
repeated handoff requests or formulaic comfort replies, recur many times with
nearly identical repairs. Training on all of them would spend most of the update
on a handful of patterns. Before training, repair tuples are therefore
deduplicated per micro-scenario under a cap: for the deployed model, repeated
handoff micro-scenarios typically keep at most two tuples and homogeneous comfort
patterns at most five, distinct compound errors are kept in full, and no further
reweighting or oversampling is applied.

The failures behind a set of repair tuples occur at different times, and Skills
and SOPs are revised in between, so a training set mixes several policy
versions. We keep this mixture by design. A tuple is a good example not because
its SOP is the newest, but because its repair follows the SOP that its own input
carries: the compliance constraint of Section~\ref{sec:construct} judges every
repair against the policy in force when the failure occurred, and that policy is
part of $x$. Training on such a mixture teaches the model to follow the SOP it is
given and to generalize across revisions, instead of memorizing one version that
will soon be replaced. The cold-start corpus follows the same principle
(Section~\ref{sec:cold_objective}). Tuples from superseded policies are
therefore neither discarded nor rewritten to the latest policy.

\begin{algorithm}[t]
\caption{One run of \care}
\label{alg:care}
\small
\begin{algorithmic}[1]
\Require base parameters $\theta_{\mathrm{base}}$; production traces $D$
\State $\theta_0\gets\textsc{ColdStart}\bigl(\theta_{\mathrm{base}},\textsc{DataEngine}(D)\bigr)$
  \Comment{Phase I, Sec.~\ref{sec:cold_start}}
\For{$k=0,1,2,\dots$}\Comment{Phase II}
  \State $\mathcal T_k\gets\textsc{Serve}(\pi_{\theta_k})$
    \Comment{part of production traffic}
  \State $B_k\gets\textsc{InspectOnline}(\mathcal T_k)$
    \Comment{24/7; confirmed failures only}
  \For{$b\in B_k$}
    \State $(\mathrm{route}_b,d_b)\gets\textsc{Attribute}(b)$
  \EndFor
  \State $\textsc{RepairComponents}\bigl(\{b:\mathrm{route}_b\neq\text{model}\}\bigr)$
    \Comment{Skill/SOP, tools, harness}
  \State $\Delta_k\gets\textsc{ConstructRepairs}\bigl(\{(b,d_b):\mathrm{route}_b=\text{model}\}\bigr)$
    \Comment{Sec.~\ref{sec:construct}}
  \State $\theta'\gets\theta_k$;\; $\bar\theta\gets\theta_k$
  \Repeat
    \State sample $(x,I)$ from $\Delta_k$;\; sample $\hat y\sim\pi_{\theta'}(\cdot\mid x)$
    \State compute $\ell_t$ on $\hat y$ from $p_{\theta',t}$ and $q_t(\cdot;\bar\theta,I)$;\;
      $\theta'\gets\textsc{Step}\bigl(\theta',\nabla_{\theta'}\hat{\mathcal L}\bigr)$
      \Comment{Eqs.~\eqref{eq:opsd_forward}--\eqref{eq:opsd_batch}}
  \Until{converged}
  \State $S(\theta')\gets\textsc{Evaluate}(\theta')$
    \Comment{simulation test, then offline inspection of its traces}
  \If{$S(\theta')\ge S(\theta_k)$ (Eq.~\ref{eq:gate}), confirmed online, and approved}
    \State $\theta_{k+1}\gets\theta'$ \Comment{promote}
  \Else
    \State $\theta_{k+1}\gets\theta_k$ \Comment{keep the deployed version}
  \EndIf
\EndFor
\end{algorithmic}
\end{algorithm}

%% file: Technical_Report/figs/construction.tex
\begin{figure}[t]
\centering
\hyphenpenalty=10000
\resizebox{\linewidth}{!}{%
\begin{tikzpicture}[
  font=\small,
  cell/.style={rectangle, rounded corners=1.5pt, draw=#1!75!black, fill=#1!12,
    line width=0.6pt, minimum width=0.92cm, minimum height=0.62cm,
    inner sep=1pt, font=\footnotesize},
  step/.style={rectangle, rounded corners=3pt, draw=#1!80!black, fill=#1!9,
    line width=0.8pt, align=center, text width=2.45cm, minimum height=1.45cm,
    inner sep=3pt},
  view/.style={rectangle, rounded corners=2.5pt, draw=#1!80!black, fill=#1!9,
    line width=0.7pt, align=center, text width=3.25cm, minimum height=0.62cm,
    inner sep=2pt, font=\scriptsize},
  flow/.style={-{Stealth[length=2.2mm]}, line width=0.9pt, draw=careink!75},
  exit/.style={-{Stealth[length=1.8mm]}, line width=0.6pt, draw=caregray,
    dashed},
  lbl/.style={font=\scriptsize\itshape, text=careink!80, align=center,
    inner sep=1.5pt},
  ptitle/.style={font=\small\bfseries, text=careink, anchor=west},
]
\node[ptitle] at (-0.9,1.55) {(a) A confirmed failure, split at the first model error};
\foreach \t/\c [count=\i] in {$u_1$/careblue,$a_1$/careteal,$y_1$/caregray,%
  $u_2$/careblue,$a_2$/careteal,$y_2$/caregray}
  \node[cell=\c] (p\i) at ({(\i-1)*1.08},0) {\t};
\node[cell=carered, line width=1pt] (err) at (7.15,0) {$a_{t^\star}$};
\node[font=\footnotesize\bfseries, text=carered] at (7.15,0.58) {$\times$};
\foreach \t [count=\i] in {$u_3$,$a_{t^\star+1}$,$\cdots$,$z$}
  \node[cell=caregray, fill=caregray!5, draw=caregray!60, text=caregray]
    (f\i) at ({7.15+\i*1.08},0) {\t};
\node[cell=careorange, minimum width=1.9cm, text width=1.8cm, align=center]
  (diag) at (13.05,0) {diagnosis $d$};

\draw[carered!80, line width=0.9pt, dash pattern=on 4pt off 2pt]
  (6.42,0.75) -- (6.42,-0.75);
\node[font=\scriptsize\bfseries, text=carered!85!black] at (6.42,0.98) {causal cut};

\draw[decorate, decoration={brace, mirror, amplitude=5pt, raise=4pt},
  line width=0.6pt, draw=careteal]
  (-0.46,-0.31) -- (5.86,-0.31)
  node[midway, below=10pt, font=\scriptsize, text=careteal!80!black, align=center]
  {decision-time input $x=h_{t^\star}$: exact request\\
   (messages, tool schemas, Skill, settings) \textbf{student-visible}};
\draw[decorate, decoration={brace, mirror, amplitude=5pt, raise=4pt},
  line width=0.6pt, draw=careorange]
  (6.69,-0.31) -- (14.0,-0.31)
  node[midway, below=10pt, font=\scriptsize, text=careorange!80!black, align=center]
  {hindsight $\mathcal H=(a^{-},h_{>t^\star},z,d)$: erroneous output, later turns,\\
   outcome, diagnosis \textbf{teacher-only}};

\node[ptitle] at (-0.9,-2.35) {(b) Retrospective repair construction};
\node[step=careorange] (s1) at (0.25,-3.6)
  {\textbf{1. Localize}\\[1pt]\scriptsize first defective\\decision, kept only\\if it is the model's};
\node[step=careteal] (s2) at (3.05,-3.6)
  {\textbf{2. Truncate}\\[1pt]\scriptsize exact request at $t^\star$\\becomes $x$; the rest\\becomes $\mathcal H$};
\node[step=careblue] (s3) at (5.85,-3.6)
  {\textbf{3. Repair}\\[1pt]\scriptsize minimal $a^{+}$: local,\\feasible, compliant,\\non-anticipating};
\node[step=carepurple] (s4) at (8.65,-3.6)
  {\textbf{4. Verify}\\[1pt]\scriptsize structural checks,\\then every verifier\\axis must pass};
\node[step=caregreen] (s5) at (11.45,-3.6)
  {\textbf{Repair tuple}\\[1pt]\scriptsize $(x,\ctx,a^{-},a^{+},d)$};
\foreach \a/\b in {s1/s2,s2/s3,s3/s4,s4/s5}
  \draw[flow] (\a) -- (\b);

\node[view=caregray] (sft) at (14.85,-2.85) {\textbf{SFT}: $x\to a^{+}$};
\node[view=caregray] (dpo) at (14.85,-3.6) {\textbf{DPO}: $a^{+}\succ a^{-}\mid x$};
\node[view=careteal, fill=careteal!16] (opsd) at (14.85,-4.35)
  {\textbf{OPSD}: student $x$; teacher $x{+}I$};
\foreach \v in {sft,dpo,opsd}
  \draw[flow] (s5.east) -- (\v.west);

\draw[exit] (s1.south) -- ++(0,-0.55) node[below, lbl] {non-model first\\defect $\to$ routed};
\draw[exit] (s2.south) -- ++(0,-0.55) node[below, lbl] {no unique anchor\\$\to$ pending};
\draw[exit] (s3.south) -- ++(0,-0.55) node[below, lbl] {no legal local repair\\$\to$ skipped or pending};
\draw[exit] (s4.south) -- ++(0,-0.55) node[below, lbl] {any axis fails\\$\to$ rejected};

\draw[flow, draw=careorange!70, dashed] (diag.east) -- ++(0.4,0) |- (8.0,-1.95)
  node[lbl, pos=0.75, below] {hindsight guides the repair, never the input}
  -| (s3.north);
\end{tikzpicture}}
\caption{Retrospective repair construction. (a) A confirmed failure is cut at
the first decision the model got wrong. Everything the model could see at that
moment becomes the student input; the erroneous output, the rest of the
conversation, the outcome, and the diagnosis become hindsight. (b) The proposer
uses hindsight to write a minimal repair that remains justified by the
decision-time input alone, an independent verifier accepts or rejects it, and
each accepted tuple can be projected into SFT, DPO, or OPSD data. Every case
ends in exactly one terminal state.}
\label{fig:construction}
\end{figure}
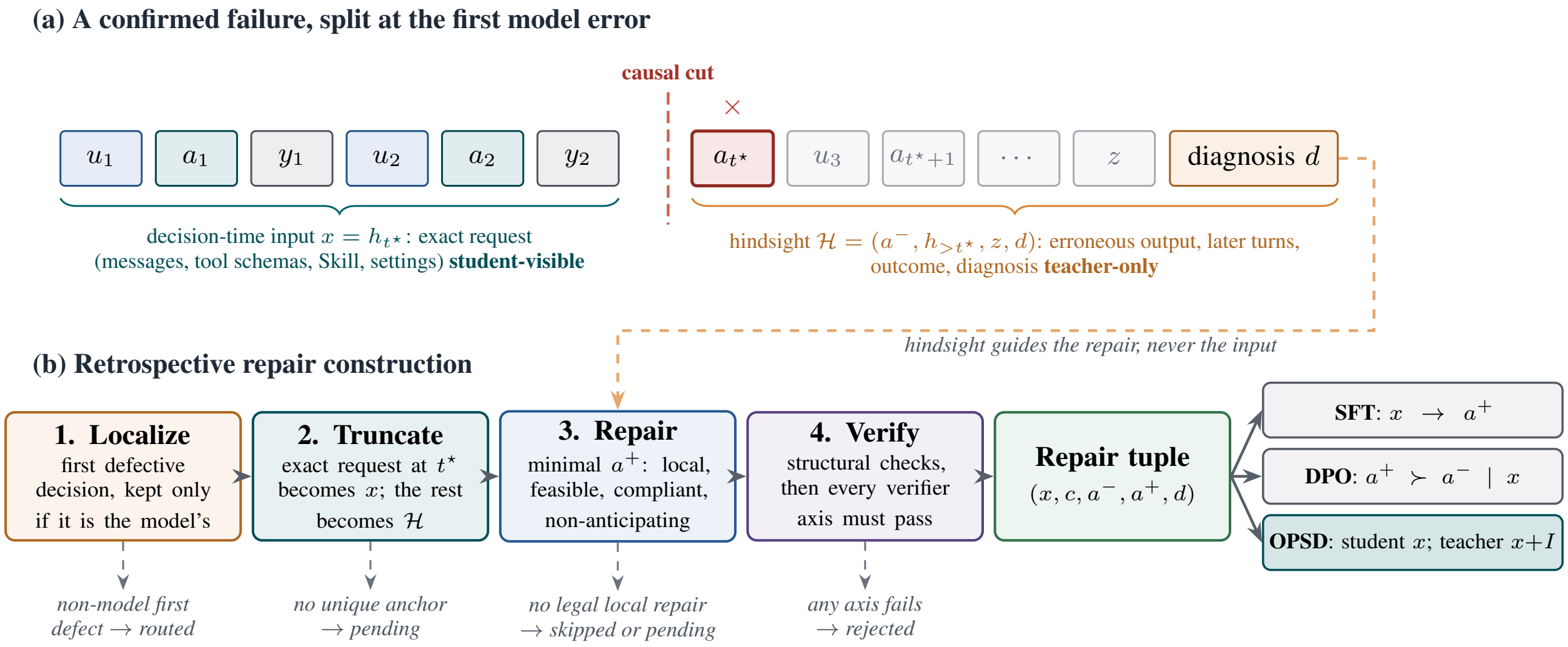

%% file: Technical_Report/figs/opsd.tex
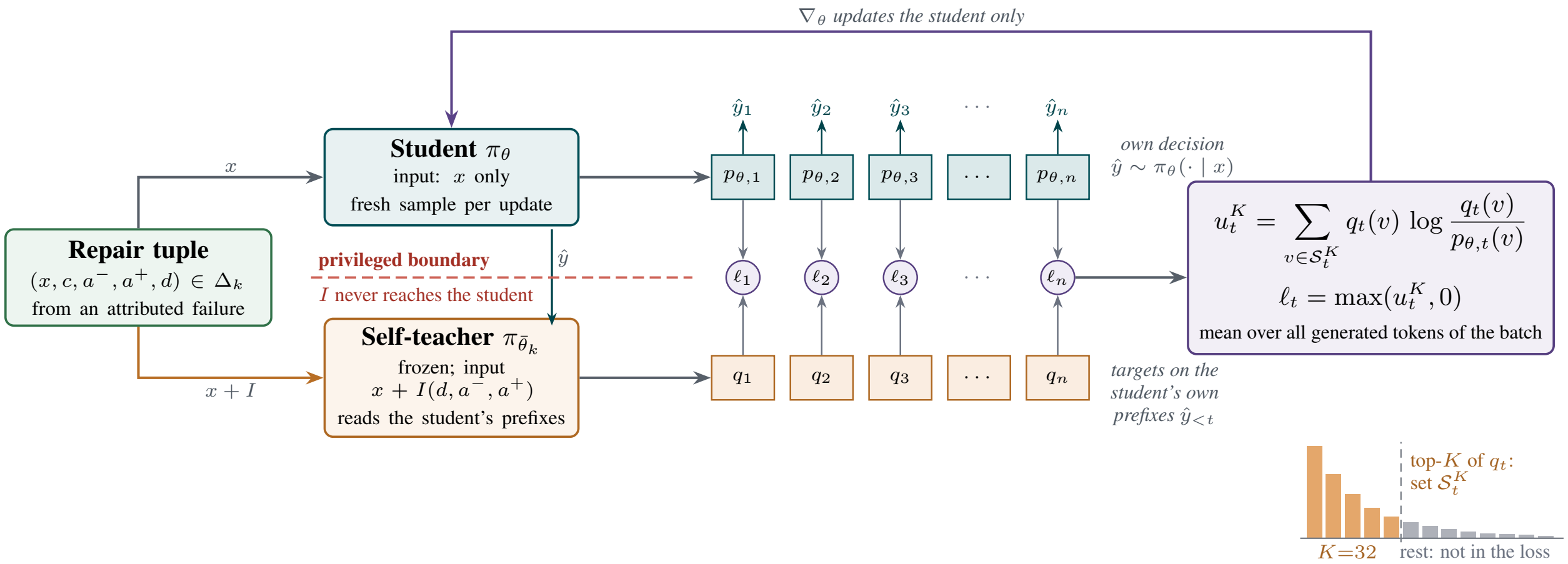
\begin{figure}[t]
\centering
\hyphenpenalty=10000
\resizebox{\linewidth}{!}{%
\begin{tikzpicture}[
  font=\small,
  box/.style={rectangle, rounded corners=3pt, draw=#1!80!black, fill=#1!9,
    line width=0.8pt, align=center, text width=2.95cm, minimum height=1.2cm,
    inner sep=3pt},
  tok/.style={rectangle, draw=careteal!80!black, fill=careteal!14,
    line width=0.55pt, minimum width=0.78cm, minimum height=0.55cm,
    inner sep=1pt, font=\scriptsize},
  dist/.style={rectangle, draw=careorange!80!black, fill=careorange!14,
    line width=0.55pt, minimum width=0.78cm, minimum height=0.55cm,
    inner sep=1pt, font=\scriptsize},
  div/.style={circle, draw=carepurple!80!black, fill=carepurple!10,
    line width=0.55pt, inner sep=0.5pt, minimum size=0.42cm,
    font=\tiny},
  flow/.style={-{Stealth[length=2.2mm]}, line width=0.9pt, draw=careink!75},
  thin/.style={-{Stealth[length=1.6mm]}, line width=0.55pt, draw=caregray},
  lbl/.style={font=\scriptsize\itshape, text=careink!80, align=center,
    inner sep=1.5pt},
]
\node[box=caregreen, text width=3.1cm] (task) at (-5.6,0.5)
  {\textbf{Repair tuple}\\[1pt]\scriptsize $(x,\ctx,a^{-},a^{+},d)\in\Delta_k$\\
   from an attributed failure};
\node[box=careteal] (stu) at (-1.7,1.75)
  {\textbf{Student} $\pi_\theta$\\[1pt]\scriptsize input: $x$ only\\ fresh sample per update};
\node[box=careorange] (tea) at (-1.7,-0.75)
  {\textbf{Self-teacher} $\pi_{\bar\theta_k}$\\[1pt]\scriptsize frozen; input $x+I(d,a^{-},a^{+})$\\ reads the student's prefixes};

\foreach \i/\t in {1/$p_{\theta,1}$,2/$p_{\theta,2}$,3/$p_{\theta,3}$,4/$\cdots$,5/$p_{\theta,n}$}
  \node[tok] (s\i) at ({0.95+0.98*\i},1.75) {\t};
\foreach \i/\t in {1/$\hat y_1$,2/$\hat y_2$,3/$\hat y_3$,5/$\hat y_{n}$} {
  \node[font=\scriptsize, text=careteal!70!black, inner sep=1pt] (y\i)
    at ({0.95+0.98*\i},2.62) {\t};
  \draw[thin, draw=careteal!70!black] (s\i) -- (y\i);
}
\node[font=\scriptsize, text=caregray] at ({0.95+0.98*4},2.62) {$\cdots$};
\foreach \i/\t in {1/$q_1$,2/$q_2$,3/$q_3$,4/$\cdots$,5/$q_{n}$}
  \node[dist] (q\i) at ({0.95+0.98*\i},-0.75) {\t};
\foreach \i/\l in {1/1,2/2,3/3,5/n} {
  \node[div] (d\i) at ({0.95+0.98*\i},0.5) {$\ell_{\l}$};
  \draw[thin] (s\i) -- (d\i);
  \draw[thin] (q\i) -- (d\i);
}
\node[font=\scriptsize, text=caregray] at ({0.95+0.98*4},0.5) {$\cdots$};

\draw[flow] (task) |- node[lbl,pos=0.75,above] {$x$} (stu);
\draw[flow, draw=careorange!85!black] (task) |- node[lbl,pos=0.75,below]
  {$x+I$} (tea);
\draw[flow] (stu) -- (s1);
\draw[flow] (tea) -- (q1);
\draw[-{Stealth[length=1.6mm]}, line width=0.7pt, draw=careteal!70!black]
  (-0.45,1.1) -- node[lbl, right, pos=0.3] {$\hat y$} (-0.45,-0.1);

\node[lbl, anchor=west] at (6.45,2.0) {own decision\\$\hat y\sim\pi_\theta(\cdot\mid x)$};
\node[lbl, anchor=west] at (6.45,-0.95) {targets on the\\student's own\\prefixes $\hat y_{<t}$};

\node[rectangle, rounded corners=3pt, draw=carepurple!80!black,
  fill=carepurple!9, line width=0.8pt, align=center, minimum height=1.2cm,
  inner sep=4pt] (loss) at (9.75,0.62)
  {$\displaystyle u^{K}_t=\sum_{v\in\mathcal S^{K}_t}q_t(v)\,
   \log\frac{q_t(v)}{p_{\theta,t}(v)}$\\[3pt]
   $\ell_t=\max(u^{K}_t,0)$\\[2pt]
   \scriptsize mean over all generated tokens of the batch};
\draw[flow] (d5) -- (loss.west |- d5);
\draw[flow, draw=carepurple!85!black] (loss.north) |- node[lbl,pos=0.75,above]
  {$\nabla_\theta$ updates the student only} ([yshift=1.2cm]stu.north) -- (stu.north);

\begin{scope}[shift={(8.95,-2.75)}]
  \foreach \i/\h in {0/1.15,1/0.8,2/0.55,3/0.38,4/0.27}
    \fill[careorange!70] ({0.24*\i},0) rectangle ++(0.19,\h);
  \foreach \i/\h in {5/0.2,6/0.15,7/0.11,8/0.08,9/0.06,10/0.05,11/0.04,12/0.03}
    \fill[caregray!55] ({0.24*\i},0) rectangle ++(0.19,\h);
  \draw[line width=0.5pt, draw=careink!60] (-0.08,0) -- (3.2,0);
  \draw[densely dashed, line width=0.5pt, draw=careink!60] (1.175,-0.05) -- (1.175,1.2);
  \node[font=\scriptsize, text=careorange!75!black, align=left, anchor=south west,
    inner sep=1pt] at (1.25,0.5) {top-$K$ of $q_t$:\\ set $\mathcal S^{K}_t$};
  \node[font=\scriptsize, text=caregray, anchor=north, inner sep=1.5pt] at (2.1,-0.02)
    {rest: not in the loss};
  \node[font=\scriptsize, text=careorange!75!black, anchor=north, inner sep=1.5pt]
    at (0.5,-0.02) {$K{=}32$};
\end{scope}

\draw[carered!80, line width=0.8pt, dash pattern=on 4pt off 2pt]
  (-3.45,0.5) -- (1.3,0.5);
\node[font=\scriptsize\bfseries, text=carered!85!black, anchor=south west, fill=white,
  inner sep=1pt] at (-3.4,0.55) {privileged boundary};
\node[font=\scriptsize, text=carered!85!black, anchor=north west, fill=white,
  inner sep=1pt] at (-3.4,0.42) {$I$ never reaches the student};
\end{tikzpicture}}
\caption{Attribution-guided on-policy self-distillation at the failing decision.
The student samples a fresh decision $\hat y$ from the decision-time input $x$. A
frozen copy of the same model, additionally given the privileged information $I$
(diagnosis, incident-time SOP, and verified repair), reads the same prefixes, so
at every token position both models give a distribution over the vocabulary. The
loss compares the two on the teacher's top-$K$ tokens, in the forward direction
with a floor at zero in the deployed configuration, and is backpropagated
directly. Gradients flow only into the student, and $I$ stays on the teacher side
of the privileged boundary. The direction of the KL divergence, $K$, and the
floor are ablated in Section~\ref{sec:exp_loss}.}
\label{fig:opsd}
\end{figure}

%% file: Technical_Report/sections/07_evaluation.tex
\section{Evaluation System and Promotion}
\label{sec:evaluation}

Before a candidate replaces the deployed version, it is evaluated by an
evaluation system with two chained offline stages, and then compared with the
incumbent online (Section~\ref{sec:online_eval}). The candidate first runs an
executable simulation test (Section~\ref{sec:simulation}), which yields its
simulation accuracy; the conversation traces it produces there are then
inspected offline (Section~\ref{sec:offline_inspection}), which yields its
inspection Bad Case rate. The simulation test is compiled automatically from the
latest Skill/SOP and then frozen, so every candidate evaluated against it sees
exactly the same cases, and both scores are computed on the same runs. This
separates evaluation from the continuous online inspection of
Section~\ref{sec:inspection}, whose cases change with the traffic and feed
learning rather than measurement.

\subsection{Executable simulation test}
\label{sec:simulation}
A question-answering test treats the user's message as a fixed input and a
single reply as the output, so it cannot separate saying the right thing from
getting the matter resolved. In customer service, information is revealed over
several turns, queries and business actions change the state of orders and
after-sales requests, and the user acts on what the agent has done. A run must
therefore answer three questions: what the agent knew, that is, whether it
obtained the facts it needed and which remained uncertain; what it did, that
is, the timing and order of its queries, explanations, confirmations, and
business actions; and what resulted, that is, how the user's situation and the
business state actually changed. The object under test is consequently a
trajectory produced under fixed conditions rather than a reply, and different
wording is acceptable whenever the process and the outcome satisfy the rules.
The simulation test realizes this as a closed loop: the conditions of a case
are fixed before the run, the simulated user, the agent, and a simulated
business world interact during it, and the verdict is decided afterwards from
what actually happened (Figure~\ref{fig:simulation}). A keyword-matching mock
cannot represent state that changes with actions and time, and executing
against production systems would introduce sensitive data and uncontrolled
side effects.

\input{Technical_Report/figs/simulation}

\paragraph{Test cases and frozen conditions.}
The $j$-th test case is the tuple
\begin{equation}
  e_j=\bigl(\phi_j,\;x_{j,0},\;\ctx,\;b_j,\;\Omega_j\bigr),
  \label{eq:testcase}
\end{equation}
where $\phi_j$ is the user's situation (why the user makes contact, what the
user knows at the start, and what the user wants resolved), $x_{j,0}$ is the
initial business state of Eq.~\eqref{eq:trajectory}, including events that are
scheduled but have not yet occurred, $\ctx$ is the operating context of
Eq.~\eqref{eq:context} under which the case is compiled, $b_j$ is the decision
branch that the case targets, and $\Omega_j$ is its acceptance oracle. All
components must refer to the same Skill/SOP version, and a case with a missing
component is not compiled rather than being completed from defaults. Freezing a
case fixes its starting point and its acceptance criteria, not the dialogue:
everything after the first turn is determined by the actions actually taken. A
frozen test $\mathcal Z=\{e_j\}_{j=1}^{N}$ is run with the same user simulator,
transition rules, tool set, run budget, and judging procedure for every
candidate, so that a difference between two candidates' results can be
attributed to the agents rather than to the conditions.

\paragraph{Branch-driven scenarios.}
The unit of coverage is a decision branch, a point at which a fact or an action
changes what the correct decision is, rather than a paraphrase of the user's
request. A user who reports that an order has not arrived does not by that
statement determine the correct reply: the order query may return that the
order is still being delivered, or that it was delivered although the user
denies receiving it, and the Skill/SOP may require different handling in the
two cases. A pair of cases that differ only in this result tests whether the
agent decides from the queried fact rather than giving one answer to every
report of a missing order. Each branch closes a five-link chain,
\begin{equation}
  \text{reachable state}\;\to\;\text{observable signal}\;\to\;
  \text{user reaction}\;\to\;\text{agent decision}\;\to\;
  \text{post-run evidence},
  \label{eq:branch_chain}
\end{equation}
and a missing link produces false coverage: a branch can be listed in the plan
while its signal never appears in a run, or the agent can make the decision
while no evidence shows on what basis and with what effect.

\paragraph{Closed-loop dynamics.}
Each case is executed in a partially observable environment with two
participants,
\begin{equation}
  \mathcal E=\bigl(\mathcal X,\,A_{\mathrm u},\,A_{\mathrm c},\,T,\,
  O_{\mathrm u},\,O_{\mathrm c},\,\Omega\bigr),
  \label{eq:env}
\end{equation}
where $\mathcal X$ is the space of business states, $A_{\mathrm u}$ the user's
utterances and operations, $A_{\mathrm c}$ the agent's replies, queries, and
business actions, $T$ the transition function, $O_{\mathrm u}$ and
$O_{\mathrm c}$ the user- and agent-visible projections of the state, and
$\Omega$ the acceptance oracle. At turn $t$ the user and the agent act on their
own histories, and the business world advances on the actions that actually
occurred and on the elapsed virtual time $\Delta_t$:
\begin{equation}
  a^{\mathrm u}_t\sim\pi_{\mathrm u}\bigl(\cdot\mid h^{\mathrm u}_t,\phi_j\bigr),
  \qquad
  a_t\sim\pi_{\theta}\bigl(\cdot\mid h_t,\ctx\bigr),
  \qquad
  x_{t+1}=T\bigl(x_t,\,a^{\mathrm u}_t,\,a_t,\,\Delta_t\bigr).
  \label{eq:dynamics}
\end{equation}
A user action states a request, supplies information, or operates a card that
has been shown; an agent action is a reply, a query, or a business action, as in
Eq.~\eqref{eq:trajectory}. The transition function is the only way the state
changes. A query returns information and leaves $x_t$ unchanged; a refund or
cancellation exists only if a registered and validated action was executed by
the system simulator; and an agent's statement that something has been done is
not a transition. Each request returns an explicit \emph{executed},
\emph{business-rejected}, or \emph{protocol-failed} outcome, which is the result
$y_t$ of Eq.~\eqref{eq:trajectory}, and a repeated request does not apply the
same change twice. Scheduled events, such as the completion of a refund review,
take effect only when the virtual clock passes them. Because $T$ and the event
schedule are deterministic, variation across repeated runs of a case comes from
the two language models and not from the environment. In
Figure~\ref{fig:simulation}, the scenario compiler produces $e_j$, the user
simulator implements $\pi_{\mathrm u}$, the system simulator holds $x_t$ and
implements $T$, and the orchestrator is the only component that sequences
turns, advances the virtual clock, and ends the run. Table~\ref{tab:sim_roles}
lists what each role may and may not do.

\begin{table}[t]
\caption{Roles in a simulation run. Each role has a fixed boundary.}
\label{tab:sim_roles}
\centering
\small
\begin{tabularx}{\linewidth}{@{}l Y Y@{}}
\toprule
Role & May & May not \\
\midrule
Scenario compiler & Fix the user situation, initial state, Skill/SOP version, branch, and oracle & Choose an action for the agent during the run \\
User simulator & Speak, confirm, or operate on what has been shown & Read $x_t$ or $\Omega_j$ \\
System simulator & Hold $x_t$, execute actions, advance virtual time & Infer a state change from the agent's wording \\
Agent under test & Query, decide, reply, and act & Observe $x_t$ other than through query results, or see $\Omega_j$ \\
Post-run assessment & Judge from $\Omega_j$ and the recorded evidence & Steer any participant during the run \\
\bottomrule
\end{tabularx}
\end{table}

\paragraph{Information boundary.}
The business state is never shown to either participant. The user and the
agent each receive their own projection of it,
\begin{equation}
  o^{\mathrm u}_t=O_{\mathrm u}\bigl(x_t;\,\text{content displayed to the user}\bigr),
  \qquad
  o_t=O_{\mathrm c}\bigl(x_t;\,\text{results returned to the agent}\bigr),
  \label{eq:projection}
\end{equation}
and the histories $h^{\mathrm u}_t$ and $h_t$ accumulate only their owner's
observations and actions. The two windows differ: that the system holds an
order status does not mean that the user has seen it, and an agent that has not
yet queried cannot answer from the hidden state. The acceptance oracle is read
only after the run, so a run depends on its case only through the starting
conditions,
\begin{equation}
  \tau_j\sim P\bigl(\cdot\mid\phi_j,\,x_{j,0},\,\ctx;\;
  \pi_{\mathrm u},\,\pi_\theta,\,T\bigr),
  \label{eq:oracle_blind}
\end{equation}
in which $\Omega_j$ does not appear, and the branch $b_j$ enters only through
the state and user situation that realize it. Eq.~\eqref{eq:oracle_blind} is
the simulation counterpart of the evidence blinding of
Eq.~\eqref{eq:blinding}. It excludes two self-confirming loops: a simulated user
that reveals a result the agent has not yet produced, so that the agent reaches
the expected answer by being told it, and an environment that advances the
state along the expected path, so that an agent passes without acting.

\paragraph{How a case reaches the modules.}
The compiled case is never handed to any module as a whole. Before the run, the
orchestrator splits $e_j$ into one view per module,
\begin{equation}
  \underbrace{\phi_j}_{\text{user simulator}},\qquad
  \underbrace{x_{j,0}}_{\text{system simulator}},\qquad
  \underbrace{\ctx}_{\text{agent}},\qquad
  \underbrace{(b_j,\,\Omega_j)}_{\text{assessment}},
  \label{eq:views}
\end{equation}
and keeps the case identity, the run budget $L$, and the termination rules for
itself. The system simulator thus knows the truth but not what is expected of
the agent; the user simulator knows the user's goal and what the user would say
in response to particular signals, but neither the business state nor the
branch under test; the agent is configured with its operating context only and
learns about the case solely through the user's messages and its own query
results; and the branch and the oracle reside with the assessment, which reads
them after the run. No module that acts during the run therefore holds
$\Omega_j$, which is what makes Eq.~\eqref{eq:oracle_blind} hold by
construction. A branch is accordingly not transmitted as one object but
realized jointly: each link of the chain in Eq.~\eqref{eq:branch_chain} is
placed in the module that owns it, and the branch is reached only if the
interaction connects them. Table~\ref{tab:branch_views} traces this
distribution for one branch.

\begin{table}[t]
\caption{How the five links of a branch (Eq.~\ref{eq:branch_chain}) are placed
in the modules, for the branch in which an order is recorded as delivered while
the user reports that it has not arrived. The example illustrates the
distribution only; the correct handling is whatever the Skill/SOP in force
specifies.}
\label{tab:branch_views}
\centering
\small
\begin{tabularx}{\linewidth}{@{}l p{4.1cm} Y@{}}
\toprule
Link & Placed in & Example \\
\midrule
Reachable state & System simulator, in $x_{j,0}$ & The order is recorded as delivered \\
Observable signal & Projections $O_{\mathrm c}$ and $O_{\mathrm u}$ & The order query returns \emph{delivered} to the agent; the user learns it only when the agent says so or a card shows it \\
User reaction & User simulator, as a gated pair $(\sigma_\ell,r_\ell)$ in $\phi_j$ & Once told that the order was delivered, the user states that nothing was received \\
Agent decision & Not given to the agent; checked by $\Omega^{\mathrm{req}}_j$ and $\Omega^{\mathrm{pro}}_j$ & Whether the agent queries before answering, and how it then handles a disputed delivery \\
Post-run evidence & Assessment, from the trace and the state log & The query and its result, the agent's statements, and any executed action with its state change \\
\bottomrule
\end{tabularx}
\end{table}

\paragraph{Information-limited user simulation.}
The user simulator is goal-directed but sees only what a real user would see:
messages, cards, and page results. Its view of the case is
\begin{equation}
  \phi_j=\bigl(g_j,\;k^0_j,\;\{(\sigma_\ell,r_\ell)\}_{\ell}\bigr),
  \label{eq:user_view}
\end{equation}
consisting of a goal $g_j$, the knowledge $k^0_j$ that the user has at the
start, and reactions $r_\ell$, each attached to a signal $\sigma_\ell$ that a
user could observe, such as being told that the order was delivered or being
shown a card. A reaction becomes available only after its signal has appeared in
the user's history, so the knowledge the simulator may use at turn $t$ is
\begin{equation}
  K_t=k^0_j\cup\bigl\{r_\ell:\;\sigma_\ell\text{ occurs in }h^{\mathrm u}_t\bigr\},
  \qquad
  \pi_{\mathrm u}\bigl(\cdot\mid h^{\mathrm u}_t,\phi_j\bigr)=
  \pi_{\mathrm u}\bigl(\cdot\mid h^{\mathrm u}_t,g_j,K_t\bigr).
  \label{eq:user_knowledge}
\end{equation}
This is how a branch reaches the user: not as a script or a target, but as a
reaction that the agent's behavior or a business event has to unlock. If the
signal never occurs, the reaction is never used and the branch is not reached
in that run. The simulator keeps a persistent session, so the user can maintain
a request across turns, remember what the agent promised, and react to new
information. Besides speaking, it has two capabilities: reading $K_t$, and
requesting an operation that has been displayed, such as a click or a
confirmation, whose effect is decided by $T$. Its wording may vary from run to
run, but its knowledge may not exceed $K_t$ and $h^{\mathrm u}_t$: it may not use
$r_\ell$ before $\sigma_\ell$ has occurred, or operate content that was never
shown.

\begin{algorithm}[t]
\caption{Running one simulation case}
\label{alg:sim_case}
\small
\begin{algorithmic}[1]
\Require frozen case $e_j$; candidate $\pi_\theta$; run budget $L$
\State load $x_{j,0}$, $\phi_j$, and $\ctx$ into their modules; hold $(b_j,\Omega_j)$
  \Comment{Eq.~\eqref{eq:views}}
\State $t\gets0$;\; $a^{\mathrm u}_0\gets$ opening message from $g_j$ and $k^0_j$
\Repeat
  \State deliver $a^{\mathrm u}_t$ to the agent
  \Repeat \Comment{one agent turn}
    \State $a\sim\pi_\theta(\cdot\mid h_t,\ctx)$
    \If{$a$ is a query or a business action}
      \State system simulator executes $a$ under $T$; append $y_t$ and $O_{\mathrm c}$ to $h_t$
    \EndIf
  \Until{$a$ is a reply to the user}
  \State deliver the reply and any card through $O_{\mathrm u}$; update $K_{t+1}$
    \Comment{Eq.~\eqref{eq:user_knowledge}}
  \State $a^{\mathrm u}_{t+1}\sim\pi_{\mathrm u}(\cdot\mid h^{\mathrm u}_{t+1},g_j,K_{t+1})$;
    an operation is executed under $T$
  \State if the conversation waits, advance the virtual clock and fire due events
  \State $t\gets t+1$
\Until{the user closes the conversation, the agent transfers it, or $t=L$}
\State assemble the trace $\tau_j$ and the state log
\State judge $Y_j$ by Eq.~\eqref{eq:sim_pass}: \textsc{pass} if $Y_j=1$, else \textsc{fail}
\end{algorithmic}
\end{algorithm}

\paragraph{Running one case.}
Algorithm~\ref{alg:sim_case} lists the steps of one run. The orchestrator first
loads each view of Eq.~\eqref{eq:views} into its module. The user simulator opens with a
message formed from $g_j$ and $k^0_j$ alone, for example that the order has not
arrived. The agent then takes a turn that may contain several steps: the
orchestrator sends each query or business action to the system simulator, which
executes it under $T$ and returns the outcome $y_t$ together with the agent's
projection, until the agent replies to the user. The orchestrator delivers the
reply and any card to the user through $O_{\mathrm u}$; if this produces a
signal $\sigma_\ell$, the corresponding reaction $r_\ell$ enters $K_t$. The user
then answers, asks, or operates a displayed control, and any operation is again
executed by the system simulator. When the conversation has to wait, for
example for a review in progress, the orchestrator advances the virtual clock
and the system simulator fires the events that fall due, whose user-visible
effects reach the user through $O_{\mathrm u}$. The loop ends when the user
closes the conversation, the agent transfers it to a human, or the budget $L$ is
exhausted, and the trace and the state log are then judged as described next. In the example of Table~\ref{tab:branch_views}, the five links
are connected in order: the agent's query exposes the delivered state, its reply
passes the signal to the user, the unlocked reaction makes the user deny
receipt, the agent's subsequent handling is the decision under test, and the
trace and the state log hold the evidence. An agent that answers without
querying never exposes the signal, so the branch is not reached, and the run is
judged on what the agent actually did, for instance asserting a delivery status
it never checked.

\paragraph{Judgment.}
Every case ends in one of two outcomes, \textsc{pass} or \textsc{fail}, decided
from the recorded trace and the final business state. The oracle is decomposed
as
$\Omega_j=(\Omega^{\mathrm{req}}_j,\Omega^{\mathrm{pro}}_j,\Omega^{\mathrm{goal}}_j)$,
and
\begin{equation}
  Y_j=\mathbf 1\bigl\{\Omega^{\mathrm{req}}_j(\tau_j)\;\wedge\;
  \neg\,\Omega^{\mathrm{pro}}_j(\tau_j)\;\wedge\;
  \Omega^{\mathrm{goal}}_j(\bar x_j)\bigr\},
  \label{eq:sim_pass}
\end{equation}
with \textsc{pass} exactly when $Y_j=1$. A run that does not meet all three
conditions is a \textsc{fail}, including one that ends without the evidence to
show that they hold. The required-step condition
$\Omega^{\mathrm{req}}_j$ checks the classification of the request and the
timing and order of the necessary steps, such as querying before asserting a
fact or obtaining confirmation before a business action; where a goal admits
several legitimate paths it accepts any of them, so it constrains paths rather
than prescribing a script. The prohibition condition $\Omega^{\mathrm{pro}}_j$
detects behavior that the Skill/SOP forbids, such as promising a compensation
the policy does not allow or reporting an action that was never executed. The
goal condition $\Omega^{\mathrm{goal}}_j$ is evaluated on the final business
state $\bar x_j$, not on the agent's account of it. Wording, politeness, and
language quality are not part of $Y_j$; they are judged by the offline
inspection of Section~\ref{sec:offline_inspection}, and a fluent reply cannot
compensate for a wrong action or an unresolved outcome.

\paragraph{Simulation accuracy.}
The simulation accuracy is the share of the frozen test that passes,
\begin{equation}
  \mathrm{Acc}_{\mathrm{sim}}(\theta)=\frac{1}{N}\sum_{j=1}^{N}Y_j(\theta),
  \label{eq:sim_acc}
\end{equation}
and it is the term that enters the combined score of Eq.~\eqref{eq:combined}.
Branches are counted by whether they were reached rather than whether they were
planned: with $\mathcal J$ the branches that the test plans to exercise, the
branches covered by a candidate are
\begin{equation}
  \mathcal J^{\mathrm{hit}}(\theta)=\bigl\{b\in\mathcal J:\;
  b\text{ is reached in }\tau_j\text{ for some }j\bigr\}.
  \label{eq:branch_cov}
\end{equation}

\paragraph{Comparing candidates.}
Two candidates $\theta_A$ and $\theta_B$ face the same frozen test, Skill/SOP
version, observation boundaries, and run budget. Because the environment
responds to the agent, the two may reach different states and conversations
from the same start; this divergence is the behavior under test, and their
dialogues are not forced into a common script. They are compared case by case,
\begin{equation}
  \Delta_{B-A}=\frac{1}{N}\sum_{j=1}^{N}
  \bigl(Y_j(\theta_B)-Y_j(\theta_A)\bigr).
  \label{eq:paired}
\end{equation}
Results obtained on tests compiled from different Skill/SOP versions, or under a
different definition of success, are not a controlled comparison.

\paragraph{Test compilation and scope.}
Whenever the Skill/SOP changes, the scenario compiler regenerates the simulation
test from the latest version automatically: it enumerates the business branches
that the new policy defines and compiles a test case of the form of
Eq.~\eqref{eq:testcase} for each. The compiled set is then frozen, and all
candidates evaluated in that period run against the same static set. A result
therefore supports a bounded conclusion: on these cases, under this Skill/SOP
version, and with this evidence, the candidate passed a given share of the cases. Its validity rests on four conditions: the cases cover the
branches that matter, and many cases do not by themselves imply good branch
coverage; the oracle agrees with the rules in force and never guides the agent
during the run; the simulated user acts only on what it has seen while keeping
a natural range of expression; and the business state changes only through
executed actions and time. The result does not extrapolate to satisfaction on
production traffic, which is why a candidate is also compared online
(Section~\ref{sec:online_eval}); the fidelity of the simulator is discussed in
Section~\ref{sec:discussion}. Execution tiers and experimental controls are
detailed in Appendix~\ref{app:simulation}.

\subsection{Offline inspection test}
\label{sec:offline_inspection}
The offline inspection test takes the traces that a candidate produced in the
simulation test and judges them with the inspection procedure of
Section~\ref{sec:inspection}, in the offline mode that freezes the traces and
the evidence view (Appendix~\ref{app:inspection}). The two stages are
complementary. The simulation oracle checks business outcomes
(Eq.~\eqref{eq:sim_pass}): whether the required steps were taken, whether a
prohibited action occurred, and whether the required state change happened. The
inspection judges every decision along the way, including service quality,
language, and policy compliance in replies, so a trace that reaches the right
outcome can still contain a Bad Case. Because the traces come from the frozen
simulation test, the resulting confirmed-bad rate $\widehat R$ of
Eq.~\eqref{eq:risk}, the \emph{inspection Bad Case rate}, has a fixed
denominator and is comparable across candidates, unlike the rate observed by
online inspection.

\subsection{Online comparison}
\label{sec:online_eval}
A candidate that passes offline evaluation serves part of the production traffic
while the incumbent serves the rest at the same time. The primary online metric
is user satisfaction, compared against the incumbent in the same period; the
escalation rate to human staff is monitored as a secondary signal, since an
appropriate handoff is a correct outcome.

\subsection{Promotion}
\label{sec:gate}
The two offline scores are combined into a single evaluation score,
\begin{equation}
  S(\theta)=\tfrac12\Bigl[\mathrm{Acc}_{\mathrm{sim}}(\theta)
  +\bigl(100-\widehat R(\theta)\bigr)\Bigr],
  \label{eq:combined}
\end{equation}
where $\mathrm{Acc}_{\mathrm{sim}}$ is the simulation accuracy of
Eq.~\eqref{eq:sim_acc} and $\widehat R$ the inspection Bad Case rate, both
percentages measured on the current frozen simulation test,
so that $S$ rewards a high simulation accuracy and a low inspection Bad Case
rate equally. A candidate $\theta'$ replaces the deployed version $\theta_k$ only
if it is at least at parity on the combined score,
\begin{equation}
  S(\theta')\;\ge\;S(\theta_k),
  \label{eq:gate}
\end{equation}
and it is then confirmed online against the incumbent. Using one combined score
rather than two separate conditions accepts a candidate whose small loss on one
score is outweighed by a larger gain on the other, while still requiring the
evaluation as a whole not to regress.
Promotion to production requires
human approval, which makes \care a human-on-the-loop system in the sense of
Section~\ref{sec:related_rsi}, and the previous version remains available as a
fallback. Because the test sets are compiled from the latest Skill/SOP, a
candidate is evaluated against the policy currently in force, not against earlier
policy versions; Section~\ref{sec:discussion} discusses this limitation.

%% file: Technical_Report/figs/simulation.tex
\begin{figure}[t]
\centering
\resizebox{0.96\linewidth}{!}{%
\begin{tikzpicture}[
  font=\small,
  mod/.style={rectangle, rounded corners=3pt, draw=#1!80!black, fill=#1!9,
    line width=0.8pt, align=center, text width=3.5cm, minimum height=1cm,
    inner sep=3pt},
  io/.style={{Stealth[length=2.2mm]}-{Stealth[length=2.2mm]}, line width=0.9pt,
    draw=careink!75},
  setup/.style={-{Stealth[length=2mm]}, line width=0.7pt, draw=caregray,
    dash pattern=on 3pt off 2pt},
  lbl/.style={font=\scriptsize\itshape, text=careink!80, align=center,
    inner sep=1.5pt},
  vlbl/.style={font=\scriptsize, text=careink!85, align=center, inner sep=1.5pt},
]
\node[mod=caregray, text width=10.6cm, minimum height=1.05cm] (scen) at (0,3.45)
  {\textbf{Scenario compiler}\\[1pt]\scriptsize frozen case
   $e_j{=}(\phi_j,x_{j,0},\ctx,b_j,\Omega_j)$, split into one view per module
   (Eq.~\ref{eq:views})};
\node[mod=careorange] (user) at (-5.7,0.55)
  {\textbf{User simulator} $\pi_{\mathrm u}$};
\node[mod=careblue] (orch) at (0,0.55)
  {\textbf{Orchestrator}};
\node[mod=careteal] (agent) at (5.7,0.55)
  {\textbf{Agent under test}};
\node[mod=caregreen] (sys) at (-5.7,-2.45)
  {\textbf{System simulator}};
\node[mod=carepurple] (judge) at (5.7,-2.45)
  {\textbf{Post-run assessment}};

\draw[setup] (-5.0,2.92) -- (-5.0,1.05)
  node[vlbl, right, pos=0.45] {$\phi_j$: goal $g_j$, knowledge $k^0_j$,\\
   gated reactions $(\sigma_\ell,r_\ell)$};
\draw[setup] (0,2.92) -- (orch.north)
  node[vlbl, right, pos=0.45] {case id,\\ budget $L$};
\draw[setup] (5.0,2.92) -- (5.0,1.05)
  node[vlbl, left, pos=0.45] {$\ctx$ only};
\draw[setup] (scen.west) -- (-8.15,3.45) -- (-8.15,-2.45)
  node[vlbl, pos=0.45, rotate=90, above] {$x_{j,0}$ with event schedule}
  -- (sys.west);
\draw[setup] (scen.east) -- (8.15,3.45) -- (8.15,-2.45)
  node[vlbl, pos=0.45, rotate=-90, above] {$b_j,\ \Omega_j$, read after the run}
  -- (judge.east);

\draw[io] (user) -- node[lbl,above] {utterance, op.} node[lbl,below]
  {$o^{\mathrm u}_t$: messages,\\ cards, results} (orch);
\draw[io] (orch) -- node[lbl,above] {$o_t$, $y_t$} node[lbl,below]
  {reply, query,\\ action} (agent);
\draw[io] (orch.south west) -- node[lbl, sloped, above] {requests}
  node[lbl, sloped, below] {outcomes $y_t$, projections} (sys.north east);

\draw[setup] (orch.south east) -- node[lbl, sloped, above] {trace $\tau_j$}
  (judge.north west);
\draw[setup] (sys.east) -- node[lbl, above] {state log, final state $\bar x_j$}
  (judge.west);

\node[font=\scriptsize, text=caregray, align=center] at (0,-3.6)
  {$\mathcal E=(\mathcal X,A_{\mathrm u},A_{\mathrm c},T,O_{\mathrm u},O_{\mathrm c},\Omega)$;\quad
   solid: interaction during the run;\quad dashed: before or after the run};
\end{tikzpicture}}
\caption{The executable simulation environment. Before the run, the frozen case
is split into one view per module (Eq.~\ref{eq:views}): the user simulator
receives the user's goal, initial knowledge, and signal-gated reactions; the
system simulator the initial state and event schedule; the agent only its
operating context; and the assessment the branch and the oracle, which no
participant sees (Eq.~\ref{eq:oracle_blind}). During the run, the orchestrator
mediates every exchange, only the system simulator changes business state, and
each participant receives its own projection of that state. After the run, the
trace and the state log are judged against the oracle, and each case ends in
pass or fail (Eq.~\ref{eq:sim_pass}). The traces are also inspected offline
(Section~\ref{sec:offline_inspection}).}
\label{fig:simulation}
\end{figure}
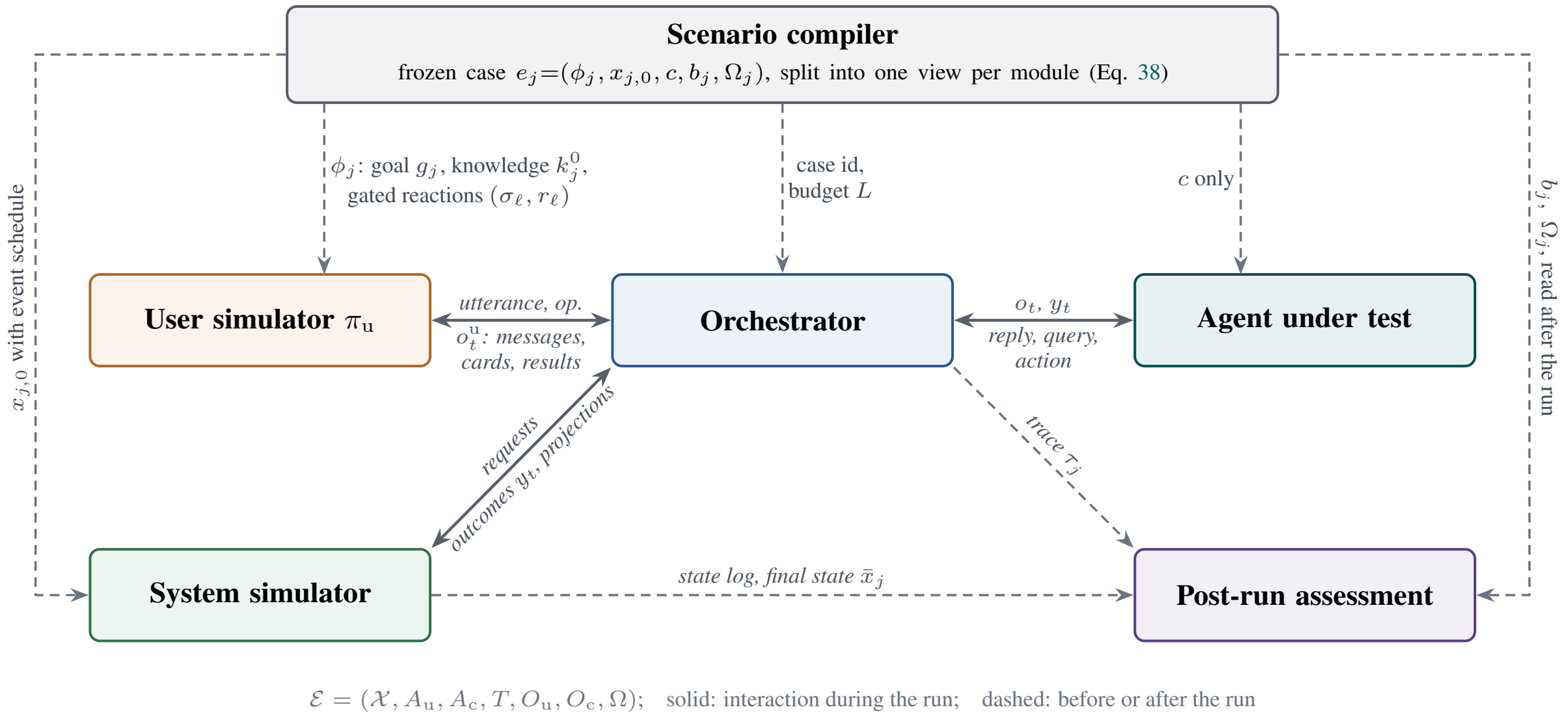

%% file: Technical_Report/sections/08_experiments.tex
\section{Experiments}
\label{sec:experiments}

We evaluate \care over five recursive iterations in a production deployment and
ask five questions. Does the recursive phase keep improving the in-house
model, offline and online (Section~\ref{sec:exp_rsi})? How does the final model
compare with the incumbent and other frontier models
(Section~\ref{sec:exp_compare})? How do the versions fare with real users
(Section~\ref{sec:exp_online})? What does the final model take to serve
(Section~\ref{sec:exp_serving})? And which design choices
matter when learning from repairs
(Sections~\ref{sec:exp_objective}--\ref{sec:exp_loss})?

\subsection{Setup}
\label{sec:exp_setup}
\paragraph{Deployment.}
The agent handles after-sales customer service for a food-delivery platform in
seven regions: Saudi Arabia (SA), Brazil (BR), Hong Kong (HK), and the United
Arab Emirates, Bahrain, Kuwait, and Qatar, which form the GCC4 group. Typical
scenarios include missing or wrong items, cold food and poor taste, spillage
and damage, non-receipt, delivery delays, refunds, and cancellations. The
agent uses Skills and tools for order lookup, compensation, refunds, and handoff
to human staff, and serves several languages.

\paragraph{Models.}
The incumbent is GPT-5.4, which served production before replacement. Our
in-house model, \caremodel, is post-trained from the open-weight
Gemma-4-31B-IT: \caremodel V0 is the cold-start model of
Section~\ref{sec:cold_start}, and V1--V5 are produced by five Phase~II
iterations, each with its own round of incremental repair data.
As references we evaluate GPT-6-Luna, Gemini-3.5-Flash, DeepSeek-V4-Flash,
Qwen3.8-27B, and the untuned Gemma-4-31B-IT, all without task-specific training.
The harness around the model is maintained by engineering and differs between
versions, for instance when a harness defect is fixed, so the absolute results of
a version can reflect harness changes as well as the model. Each change applies
to \caremodel and the incumbent at the same time, so the difference between the
two in the same period is not affected.

\paragraph{Offline evaluation.}
Both offline metrics come from the evaluation system of
Section~\ref{sec:evaluation}, not from the online inspection that supplies
training failures.
\emph{Simulation accuracy} is $\mathrm{Acc}_{\mathrm{sim}}$ of
Eq.~\eqref{eq:sim_acc}, the share of the cases in the frozen simulation test
of Section~\ref{sec:simulation} that pass.
The \emph{inspection Bad Case rate} is the confirmed-bad rate $\widehat R$ of
Eq.~\eqref{eq:risk} obtained by inspecting the same simulation traces offline
(Section~\ref{sec:offline_inspection}); lower is better, and it is a conservative observed
rate rather than a calibrated population error rate. All versions and reference
models are evaluated on the same frozen test of 559 cases.

\paragraph{Online evaluation.}
In all seven regions, production traffic is split between \caremodel and the incumbent,
each serving part of the users at the same time, so the incumbent in the same
period serves as the concurrent control. Each version served one period (P1--P6 in
Table~\ref{tab:rsi}). The primary online metric is \emph{satisfaction}, the
share of 4--5 star ratings among valid 1--5 star ratings of agent-handled
conversations. Conversations are assigned to
periods by their start time in local time. Ratings are sparse, so we report 95\%
Wilson intervals and two-sided two-proportion tests.

\subsection{Recursive improvement over five iterations}
\label{sec:exp_rsi}
\input{Technical_Report/tables/rsi_iterations}
\input{Technical_Report/figs/rsi_curves}

Table~\ref{tab:rsi} and Figure~\ref{fig:rsi_curves} follow the six versions.
Cold start makes the base model competent at the task but not yet safe enough
to replace the incumbent. V0 completes 93.0\% of simulated cases, against 79.0\%
for the untuned Gemma-4-31B-IT, yet its inspection Bad Case rate of 6.42\% is
higher than that of both the untuned base (5.59\%) and the incumbent (4.60\%),
and online it trails the incumbent by 4.6 points of satisfaction ($p=0.03$).
Imitating historical traces teaches the model to act within the business
process, but not to avoid the errors that appear only once it acts on its own,
which is the gap the recursive phase is designed to close.

\paragraph{Offline.}
The first iteration lowers the Bad Case rate from 6.42\% to 4.35\%, V4 reaches
4.11\%, and V5 reaches 1.97\%, 69\% below V0 and less than half of the
incumbent's 4.60\%. Simulation accuracy rises from 93.0\% to a peak of 95.30\% at
V3 and stands at 94.10\% for V5, 2.7 points below the incumbent. The two scores
do not improve in lockstep: the Bad Case rate rises from V1 to V2, simulation
accuracy dips slightly from V3 to V4, and V5 gives up 1.07 points of simulation
accuracy against V4 for a drop of 2.14 points in the Bad Case rate. Each
iteration learns from the failures of the version before it, so its effect
concentrates on newly observed failure families. The combined score of
Eq.~\eqref{eq:combined}, on which promotion is decided, never decreases: it
rises from 93.29 for V0 to 96.07 for V5, against 96.10 for the incumbent.

\paragraph{Online.}
The satisfaction gap closes over the same iterations
(Figure~\ref{fig:rsi_curves}a). The early versions trail the incumbent by 3.1 to
8.0 points; V4 matches it ($+0.34$ points), and V5's point estimate
exceeds it ($+1.73$ points). Neither of the last two differences is statistically
significant ($p=0.85$ and $p=0.47$), so the correct reading is that the
in-house model reaches parity with the incumbent in user satisfaction after
five iterations, with point estimates at or above it.

\subsection{Comparison with other models}
\label{sec:exp_compare}
\input{Technical_Report/tables/main_results}

Table~\ref{tab:main_results} places the final model among the reference models.
With 31B parameters, \caremodel V5 has the lowest inspection Bad Case rate of all
models evaluated, less than half of the incumbent's (1.97\% against 4.60\%). Its
combined score of 96.07 is the second highest, 0.03 below the incumbent and ahead
of every reference model, and its simulation accuracy of 94.10\% is the third
highest, behind the incumbent and Gemini-3.5-Flash (94.30\%) and ahead of
DeepSeek-V4-Flash and GPT-6-Luna. Similarly sized models
without task-specific training complete fewer than 80\% of simulated cases
(Qwen3.8-27B 77.08\%, Gemma-4-31B-IT 79.00\%), so the competence of the
in-house model comes from post-training rather than from its base. Against the
replacement criterion of Definition~\ref{def:replacement}, V5 is below the
incumbent on Bad Case rate, within 2.7 points on simulation accuracy, and at
parity on online satisfaction. Section~\ref{sec:exp_serving} reports the remaining
term of the criterion, the serving cost.

\subsection{Online deployment by scenario}
\label{sec:exp_online}
Both arms vary from period to period because traffic composition, the harness,
and operating context change over time; this is why each version is compared only with the incumbent in the
same period. In
P6, \caremodel matches or exceeds the incumbent's satisfaction in the largest
scenario, meal-handover issues ($+0.97$ points), as well as in late delivery
($+4.33$) and food quality ($+5.87$), while it trails slightly in refunds and
compensation ($-3.29$) and order acceptance ($-1.78$). Per-scenario rating counts are
small, so these differences are indicative only.

\subsection{Serving latency and cost}
\label{sec:exp_serving}
\input{Technical_Report/tables/serving}

Table~\ref{tab:serving} compares \caremodel with the incumbent. V5 replies faster
than the incumbent at every point reported, with a reduction of about 28\% at each
percentile. The mean falls by less, 18.4\%, which suggests slow replies beyond
P95, so the improvement is established only up to P95. Both arms run in the same
harness at the same time, so the gap lies with the model and how it is served.

The serving cost has three parts: the API cost of the traffic the incumbent still
serves, the NVIDIA H20 GPUs that serve \caremodel, and a fixed machine cost that
does not depend on the model. Replacing the incumbent completely lowers the cost
per service order by 70.5\%, which meets the cost term $K(\pith)<K(\pistar)$ of
Definition~\ref{def:replacement}. Model serving alone falls by 92.8\%, and what
remains is mostly the fixed machine cost. The saving does not require full
replacement: half of the traffic already lowers the cost by 35.8\%.

\subsection{Ablation: how to learn from verified repairs}
\label{sec:exp_objective}
A repair tuple can be turned into training signal in several ways
(Table~\ref{tab:views}). We compare OPSD in its deployed configuration with two
objectives that train on fixed texts: SFT, which imitates the reference repair
$a^{+}$, and DPO~\citep{dpo}, which prefers $a^{+}$ over the erroneous output
$a^{-}$. All three start from V4 and use the same repair tuples and training
budget, so the comparison isolates where the learning signal is defined: on
fixed texts, or on the outputs the model actually produces. We do not include
reinforcement learning from outcome rewards, such as GRPO~\citep{grpo}, because
no reward is available that tracks the production objective of user satisfaction
(Section~\ref{sec:why_not_rl}).

\input{Technical_Report/tables/ablation_objective}

Table~\ref{tab:ablate_objective} shows that both fixed-text objectives collapse
on an increment of this size. SFT does not yield a usable model: incremental
training destabilizes it until it emits tool calls continuously, it is stopped
by a one-case smoke test, and it never enters the full test. DPO completes
training but collapses in behavior, handing the conversation to a human in the
first turn: its simulation accuracy is 15.1\%, against 95.17\% for V4 before the
update and 94.1\% for OPSD. Its Bad Case rate of 1.88\% is low only because a
conversation that is handed off at once leaves few decisions to get wrong, which
is why promotion weighs both scores: the combined score of DPO is 56.61, against
96.07 for OPSD.

We read this as a consequence of training on fixed texts at this data scale
rather than of SFT as such. The same negative log-likelihood trains V0 stably in
cold start (Section~\ref{sec:cold_objective}) on a corpus about fifty times
larger than the increment. A few hundred repair tuples cannot constrain a
full-parameter update toward fixed texts: every token of those texts is fitted,
and nothing holds the model's behavior in place elsewhere. In OPSD the target at
every token is the model's own distribution, reshaped only where the privileged
information applies, so the same data corrects decisions without destabilizing
the model. Small increments are the normal operating condition of
the recursive phase, so this stability is a precondition for the loop, not a
secondary benefit.

\subsection{Ablation: OPSD loss}
\label{sec:exp_loss}
The OPSD objective of Section~\ref{sec:opsd} leaves three choices open: the
direction of the KL divergence, the truncation size $K$, and the floor on
negative token losses. Table~\ref{tab:ablate_loss} compares four configurations.
Configuration 1 uses the reverse divergence of Eq.~\eqref{eq:opsd_reverse} over
the top 128 tokens and a tail cell. Configurations 2, 3, and 4 use the forward
partial sum of Eq.~\eqref{eq:opsd_forward}: configuration 2 with $K=128$ and the
floor, configuration 3 with $K=128$ and no floor, and configuration 4 with $K=32$
and the floor. All four start from V4 with V4 as the
frozen teacher, train for three epochs (210 steps) on the same 280 repair
tuples, and are evaluated at the third epoch. The remaining settings are shared
(Appendix~\ref{app:opsd_config}): every loss is backpropagated directly and
averaged over the generated tokens of the batch, with no task reward.

\input{Technical_Report/tables/ablation_loss}

Configuration 4 is the best overall and the one deployed as V5: it matches the
highest simulation accuracy (94.1\%) and has the highest combined score (96.07).
Two comparisons change a single factor. From configuration 2 to 4, $K$ falls from 128 to 32,
which raises the simulation accuracy by 1.1 points and the combined score by
1.00. From configuration 2 to 3, the floor is removed at $K=128$, which raises the simulation
accuracy by 1.1 points and the combined score by 0.73. We do not extrapolate from
these two values of $K$ to a claim that smaller is better. The other two
comparisons change several factors at once and therefore do not support a
single-factor reading: configuration 1 to 4 (+1.51 in combined score) replaces the reverse
top-128 loss with its tail cell by the forward top-32 partial sum and adds the
floor, and configuration 3 to 4 (+0.27) changes both $K$ and the floor. The effect of the KL
direction alone is thus not isolated by this design. Configuration 1 shows the trade-off that
the combined score resolves: it has the lowest Bad Case rate (1.79\%), but also
the lowest simulation accuracy (90.9\%), and with it the lowest combined score
(94.56).

These differences rank the configurations but are not significance claims. The
Bad Case counts are the retained counts of separate review batches: those of
configurations 1 and 2 did not pass final batch acceptance, those of
configurations 3 and 4 were not re-reviewed under later rulings, and configuration
4 has one further undetermined case that is not counted. A gap of a few cases,
such as 10 against 11 between configurations 1 and 4 or 14 against 11 between
configurations 3 and 4, should therefore not be read as a difference between them.

%% file: Technical_Report/tables/rsi_iterations.tex
\begin{table}[t]
\caption{Six versions of \caremodel. V0 is the cold-start model, and V$k$ is obtained
from V$(k{-}1)$ by the $k$-th Phase~II iteration. Offline metrics are measured on
\caremodel in percent, and $S$ is the combined score of
Eq.~\eqref{eq:combined} used for promotion; online, $\Delta$ is the satisfaction of
\caremodel minus that of the incumbent serving concurrently, in percentage points,
in the period in which each version served. V5 is configuration 4
of Table~\ref{tab:ablate_loss}. No evaluation supplies privileged information to
the model.}
\label{tab:rsi}
\centering
\small
\setlength{\tabcolsep}{3.5pt}
\begin{tabular}{@{}llcccr@{}}
\toprule
& & \multicolumn{3}{c}{Offline} & Online satisfaction \\
\cmidrule(lr){3-5}\cmidrule(l){6-6}
Version & Period & Sim.\ acc.\ $\uparrow$ & Bad Case $\downarrow$ & Combined $S$ $\uparrow$ & $\Delta$ $\uparrow$ \\
\midrule
V0 & P1 (Jul 30--Aug 11) & 93.00 & 6.42 & 93.29 & $-4.57$ \\
V1 & P2 (Aug 11--18) & 93.12 & 4.35 & 94.39 & $-8.02$ \\
V2 & P3 (Aug 18--27) & 94.58 & 4.72 & 94.93 & $-4.87$ \\
V3 & P4 (Aug 27--Sep 4) & 95.30 & 4.47 & 95.42 & $-3.13$ \\
V4 & P5 (Sep 10--24) & 95.17 & 4.11 & 95.53 & $+0.34$ \\
\textbf{V5} & P6 (Oct 5--15) & 94.10 & \textbf{1.97} & 96.07 & $+1.73$ \\
\midrule
GPT-5.4 & -- & \textbf{96.80} & 4.60 & \textbf{96.10} & -- \\
\bottomrule
\end{tabular}
\end{table}

%% file: Technical_Report/figs/rsi_curves.tex
\begin{figure}[t]
\centering
\pgfplotsset{
  careaxis/.style={
    axis x line*=bottom, axis y line*=left,
    tick label style={font=\scriptsize}, label style={font=\small},
    major tick length=2pt, ymajorgrids, grid style={draw=caregray!20},
  },
  carecurve/.style={careteal, line width=1.3pt, mark=*, mark size=2pt},
}
\tikzset{
  refline/.style={dashed, line width=0.7pt},
  reflbl/.style={font=\scriptsize, anchor=south east, inner sep=1.5pt},
}
\begin{subfigure}[t]{\linewidth}
\centering
\begin{tikzpicture}
\begin{axis}[careaxis, width=0.92\linewidth, height=5.4cm,
  xmin=-0.4, xmax=5.4, xtick={0,1,2,3,4,5},
  xticklabels={{P1\\V0},{P2\\V1},{P3\\V2},{P4\\V3},{P5\\V4},{P6\\V5}},
  xticklabel style={align=center},
  ymin=-12, ymax=20, hide y axis, ymajorgrids=false,
  legend style={font=\small, draw=none, fill=none, at={(0.02,0.97)},
    anchor=north west}, legend cell align=left]
\addplot[carecurve, error bars/.cd, y dir=both, y explicit,
  error bar style={careteal, line width=0.8pt}, error mark options={rotate=90,
  mark size=2.2pt, careteal}]
  coordinates {
    (-0.06,-4.42) += (0,2.86) -= (0,2.76)
    (0.94,-1.18) += (0,7.93) -= (0,7.38)
    (1.94,-1.50) += (0,6.68) -= (0,6.28)
    (2.94,3.57) += (0,5.49) -= (0,5.33)
    (3.94,5.91) += (0,2.86) -= (0,2.83)
    (4.94,11.81) += (0,3.71) -= (0,3.73)
  };
\addlegendentry{\caremodel (in-house)}
\addplot[caregray!80!black, line width=1.3pt, mark=square*, mark size=1.8pt,
  error bars/.cd, y dir=both, y explicit,
  error bar style={caregray, line width=0.8pt}, error mark options={rotate=90,
  mark size=2.2pt, caregray}]
  coordinates {
    (0.06,0.15) += (0,2.85) -= (0,2.78)
    (1.06,6.84) += (0,1.79) -= (0,1.78)
    (2.06,3.37) += (0,1.86) -= (0,1.84)
    (3.06,6.70) += (0,1.27) -= (0,1.27)
    (4.06,5.57) += (0,2.02) -= (0,2.00)
    (5.06,10.08) += (0,2.79) -= (0,2.79)
  };
\addlegendentry{GPT-5.4 (incumbent, concurrent)}
\end{axis}
\end{tikzpicture}
\caption{Online satisfaction, with 95\% Wilson intervals}
\end{subfigure}

\vspace{4pt}
\begin{subfigure}[t]{0.49\linewidth}
\centering
\begin{tikzpicture}
\begin{axis}[careaxis, width=\linewidth, height=5.2cm,
  xmin=-0.3, xmax=5.3, xtick={0,1,2,3,4,5},
  xticklabels={V0,V1,V2,V3,V4,V5},
  ymin=90.8, ymax=97.6, ylabel={Simulation accuracy (\%)}]
\draw[refline, caregray] (axis cs:-0.3,96.8) -- (axis cs:5.3,96.8)
  node[reflbl, text=careink] {GPT-5.4};
\draw[refline, carepurple!80] (axis cs:-0.3,94.3) -- (axis cs:5.3,94.3);
\node[reflbl, anchor=south west, text=carepurple!80!black] at (axis cs:-0.25,94.3)
  {Gemini-3.5-Flash};
\draw[refline, careblue!80] (axis cs:-0.3,93.71) -- (axis cs:5.3,93.71)
  node[reflbl, anchor=north east, text=careblue!80!black] {DeepSeek-V4-Flash};
\draw[refline, careorange!80!black] (axis cs:-0.3,91.6) -- (axis cs:5.3,91.6)
  node[reflbl, text=careorange!80!black] {GPT-6-Luna};
\addplot[carecurve] coordinates {(0,93.0) (1,93.12) (2,94.58) (3,95.3) (4,95.17) (5,94.1)};
\end{axis}
\end{tikzpicture}
\caption{Offline simulation accuracy $\uparrow$}
\end{subfigure}\hfill
\begin{subfigure}[t]{0.49\linewidth}
\centering
\begin{tikzpicture}
\begin{axis}[careaxis, width=\linewidth, height=5.2cm,
  xmin=-0.3, xmax=5.3, xtick={0,1,2,3,4,5},
  xticklabels={V0,V1,V2,V3,V4,V5},
  ymin=1.5, ymax=6.9, ylabel={Inspection Bad Case rate (\%)}]
\draw[refline, carepurple!80] (axis cs:-0.3,6.44) -- (axis cs:5.3,6.44)
  node[reflbl, text=carepurple!80!black] {Gemini-3.5-Flash};
\draw[refline, careorange!80!black] (axis cs:-0.3,5.59) -- (axis cs:5.3,5.59)
  node[reflbl, text=careorange!80!black] {Gemma-4-31B-IT (zero-shot)};
\draw[refline, careblue!80] (axis cs:-0.3,4.97) -- (axis cs:5.3,4.97)
  node[reflbl, text=careblue!80!black] {DeepSeek-V4-Flash};
\draw[refline, caregray] (axis cs:-0.3,4.60) -- (axis cs:5.3,4.60)
  node[reflbl, anchor=north east, text=careink] {GPT-5.4};
\addplot[carecurve] coordinates {(0,6.42) (1,4.35) (2,4.72) (3,4.47) (4,4.11) (5,1.97)};
\end{axis}
\end{tikzpicture}
\caption{Offline inspection Bad Case rate $\downarrow$}
\end{subfigure}
\caption{Recursive improvement of \caremodel over six versions. (a) Online satisfaction
of each version against the incumbent serving concurrently in the same period;
the vertical scale is omitted.
(b, c) Offline metrics of \caremodel, with reference models as dashed
lines; Gemma-4-31B-IT (zero-shot, 79.00\%) and Qwen3.8-27B (77.08\%) fall below
the range of (b), and GPT-6-Luna (5.55\%) and Qwen3.8-27B (9.17\%) are omitted
from (c) for legibility.}
\label{fig:rsi_curves}
\end{figure}
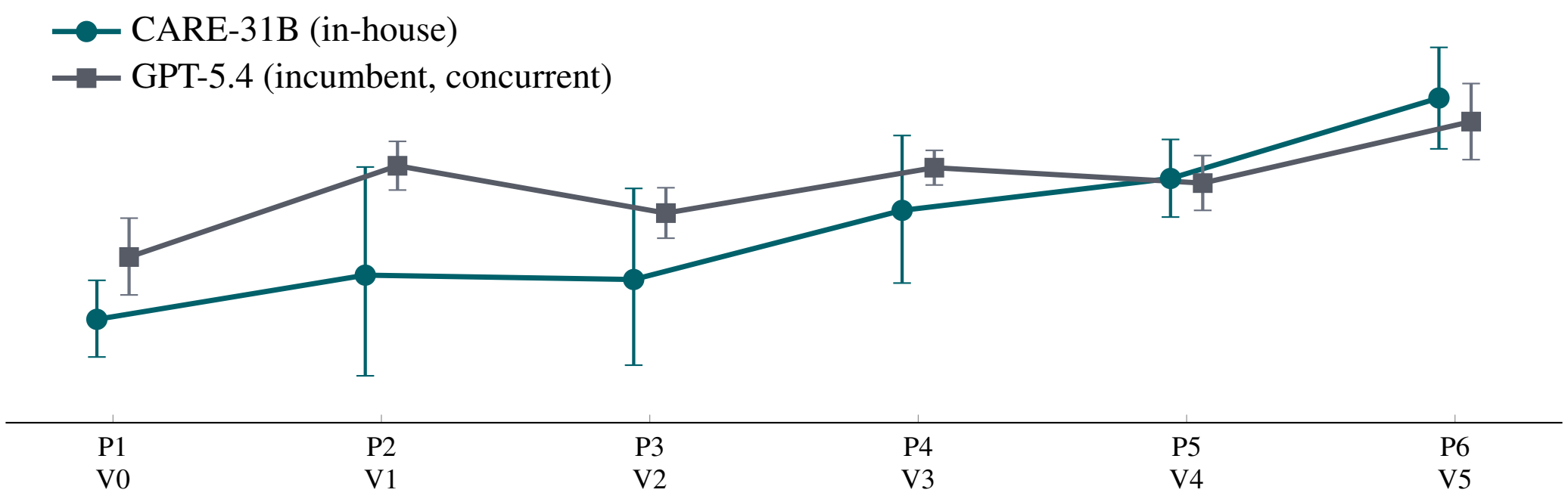
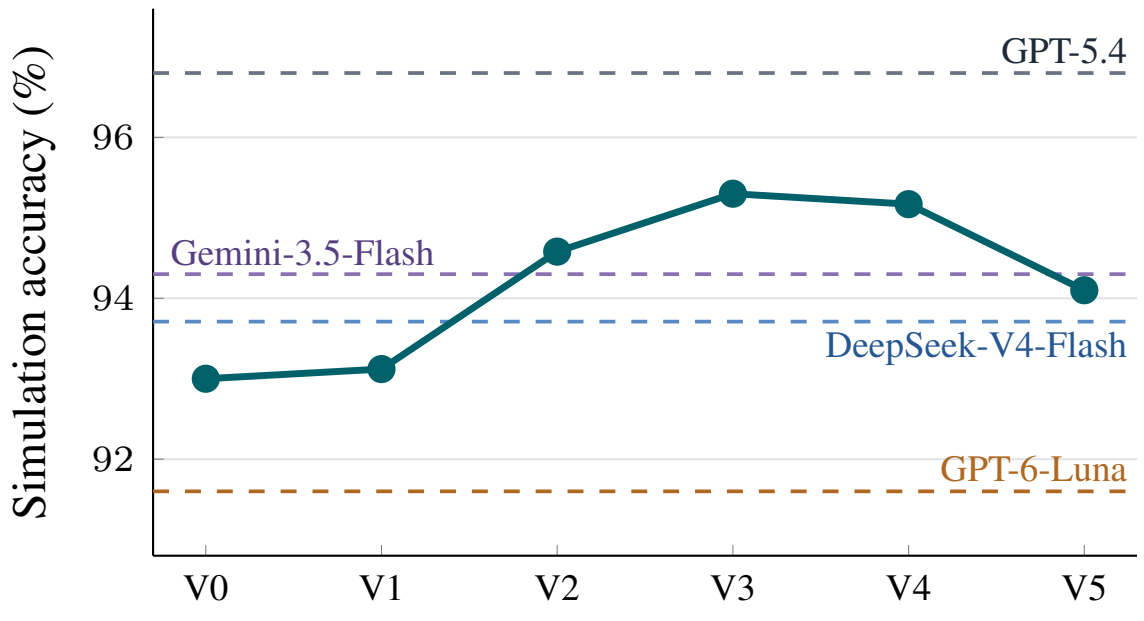
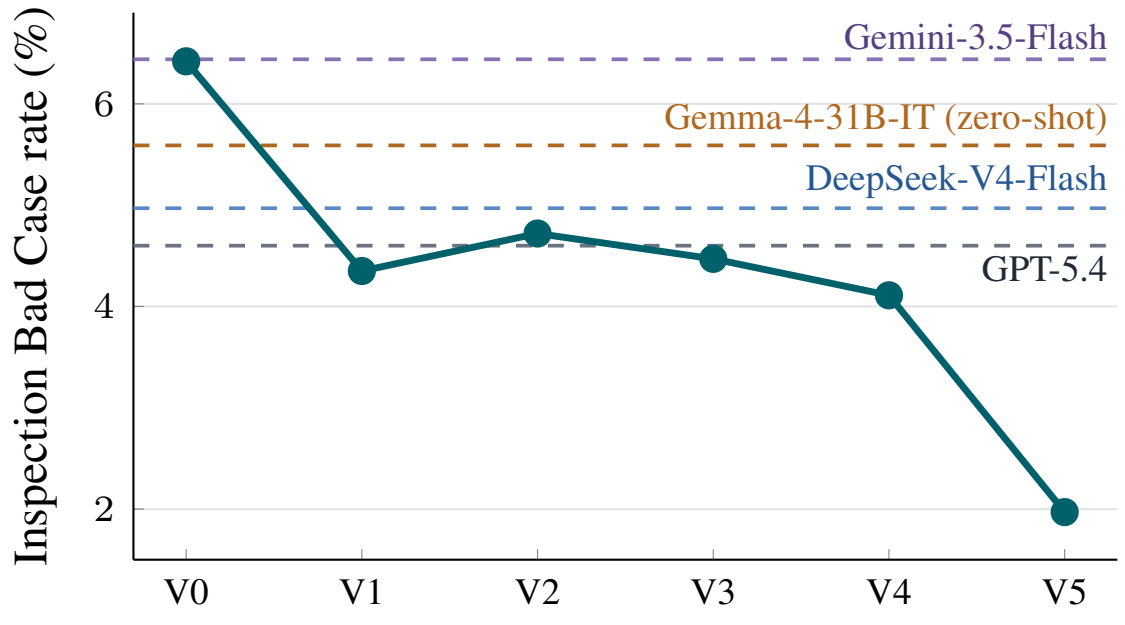

%% file: Technical_Report/tables/main_results.tex
\begin{table}[t]
\caption{Comparison with other models. Simulation accuracy is the share of
simulated cases that pass; the inspection Bad Case rate is the
confirmed-bad rate of Eq.~\eqref{eq:risk}; $S$ is the combined score of
Eq.~\eqref{eq:combined}. Reference models run without task-specific training.}
\label{tab:main_results}
\centering
\small
\begin{tabular}{@{}llccc@{}}
\toprule
Model & Role & Sim.\ accuracy (\%) $\uparrow$ & Bad Case rate (\%) $\downarrow$ & Combined $S$ $\uparrow$ \\
\midrule
GPT-5.4 & incumbent & \textbf{96.80} & 4.60 & \textbf{96.10} \\
\midrule
GPT-6-Luna & reference & 91.60 & 5.55 & 93.03 \\
Gemini-3.5-Flash & reference & 94.30 & 6.44 & 93.93 \\
DeepSeek-V4-Flash & reference & 93.71 & 4.97 & 94.37 \\
Qwen3.8-27B & reference & 77.08 & 9.17 & 83.96 \\
Gemma-4-31B-IT & base, zero-shot & 79.00 & 5.59 & 86.71 \\
\midrule
\caremodel V0 & cold start & 93.00 & 6.42 & 93.29 \\
\caremodel V5 & five iterations & 94.10 & \textbf{1.97} & 96.07 \\
\bottomrule
\end{tabular}
\end{table}

%% file: Technical_Report/tables/serving.tex
\begin{table}[t]
\caption{Reduction of \caremodel against the incumbent, in percent. \emph{Reply
latency}: online reply latency over successful replies, with V5 and the incumbent
serving at the same time. \emph{Cost per service order}: by the share of the
traffic served by the in-house model.}
\label{tab:serving}
\centering
\small
\begin{tabular}{@{}lccccccc@{}}
\toprule
& \multicolumn{4}{c}{Reply latency} & \multicolumn{3}{c}{Cost per service order} \\
\cmidrule(lr){2-5}\cmidrule(l){6-8}
& Mean & P50 & P90 & P95 & 30\% & 50\% & 100\% \\
\midrule
Reduction (\%) $\uparrow$ & 18.4 & 27.7 & 28.0 & 28.9 & 21.7 & 35.8 & \textbf{70.5} \\
\bottomrule
\end{tabular}
\end{table}

%% file: Technical_Report/tables/ablation_objective.tex
\begin{table}[t]
\caption{Ablation of the learning signal. All objectives start from V4 and use
the same repair tuples and training budget. \emph{On-policy}: the loss is defined
on the model's own samples. $S$ is the combined score of
Eq.~\eqref{eq:combined}. A dash means unavailable, not zero: SFT collapsed during
training and never entered the full test. The OPSD row is configuration 4 of
Table~\ref{tab:ablate_loss}.}
\label{tab:ablate_objective}
\centering
\small
\setlength{\tabcolsep}{4pt}
\begin{tabular}{@{}llcccc@{}}
\toprule
Objective & Learning signal & On-policy & Sim.\ acc.\ (\%) $\uparrow$ & Bad Case (\%) $\downarrow$ & $S$ $\uparrow$ \\
\midrule
SFT & imitate $a^{+}$ & $\times$ & -- & -- & -- \\
DPO & prefer $a^{+}$ over $a^{-}$ & $\times$ & 15.1 & \textbf{1.88} & 56.61 \\
\textbf{OPSD (\care)} & match self-teacher given $I$ & \checkmark & \textbf{94.1} & 1.97 & \textbf{96.07} \\
\bottomrule
\end{tabular}
\end{table}

%% file: Technical_Report/tables/ablation_loss.tex
\begin{table}[t]
\caption{Ablation of the OPSD loss. \emph{Bad Case}: confirmed Bad Cases over all cases of the report,
with their number in parentheses. $S$ is the combined score of Eq.~\eqref{eq:combined}.}
\label{tab:ablate_loss}
\centering
\small
\setlength{\tabcolsep}{3.5pt}
\begin{tabular}{@{}clcccccc@{}}
\toprule
Config. & Direction & $K$ & Tail & Floor & Sim.\ acc.\ (\%) $\uparrow$ & Bad Case (\%) $\downarrow$ & $S$ $\uparrow$ \\
\midrule
1 & reverse, Eq.~\eqref{eq:opsd_reverse} & 128 & \checkmark & -- & 90.9 & \textbf{1.79} (10) & 94.56 \\
2 & forward, Eq.~\eqref{eq:opsd_forward} & 128 & -- & \checkmark & 93.0 & 2.87 (16) & 95.07 \\
3 & forward, Eq.~\eqref{eq:opsd_forward} & 128 & -- & -- & \textbf{94.1} & 2.50 (14) & 95.80 \\
\textbf{4} & forward, Eq.~\eqref{eq:opsd_forward} & 32 & -- & \checkmark & \textbf{94.1} & 1.97 (11) & \textbf{96.07} \\
\bottomrule
\end{tabular}
\end{table}

%% file: Technical_Report/sections/09_discussion.tex
\section{Discussion}
\label{sec:discussion}

\subsection{RSI as an industrial learning workflow}
The contribution of \care is to operationalize RSI within a production
customer-service domain, connecting production feedback, failure attribution,
verified repair, and model updates under domain-specific business rules.
Failures encountered by the deployed model become learning tasks only after
evidence review and attribution. Model-attributable failures feed OPSD with the
model itself as teacher, conditioned on privileged failure information, while
component defects are routed to repair. This division of responsibility directs
improvement to the part of the service that should change. Evaluation and
promotion connect each update back to the operating requirements of the
deployed system.

The multilingual food-delivery deployment provides evidence for this workflow
in a live business setting. Across five iterations, the in-house 31B model's
inspection Bad Case rate fell below the incumbent's, and online user
satisfaction reached parity. The recurring process
turns production experience into persistent model updates, extending
improvement beyond the one-time transfer of incumbent traces used for cold
start. These results support RSI as an operational approach to improving a
deployed domain-specific agent.

\subsection{Evolving with business policies}
Customer-service policies change independently of the learner, so continued
improvement requires adaptation to the Skill/SOP currently supplied to the
agent. \care separates changes to business instructions from changes to model
behavior: component owners revise Skills/SOPs, while model-attributable
failures feed the learning loop. Training mixes failures from historical SOP
versions, each judged against the policy in force when it occurred, so that
following the given policy is the learning objective. For promotion, simulation
tests are compiled from the latest SOP and frozen; offline inspection evaluates
the resulting traces, and the combined-score gate precedes online comparison
and human approval. This coupling of policy-conditioned learning and
SOP-derived evaluation is designed to shorten the turnaround from a business
change to a validated model update.

\subsection{Why grounding keeps the loop from collapsing}
Recursive training on a model's own outputs is known to erode the tails of its
distribution~\citep{collapse}, and loops in which a model judges itself tend to
reinforce its own errors~\citep{rsi_taxonomy}. \care faces both risks because
it trains on its own rollouts and uses itself as the teacher. Four design
choices counter these risks. First, failures are established through executed
state and reviewed evidence, never through the model's opinion of its own
output. Second, learning uses the model's own distribution, reshaped only where
a verified diagnosis applies, as its target, rather than raw self-generated
text. Third, allocation in cold start and capped selection in Phase~II
deliberately preserve low-frequency regions and scenarios. Finally, promotion
depends on fixed test sets that are separate from the traffic used for learning
and evaluated without the diagnosis.
Together, these choices keep the loop in the regime that recent analyses
describe as productive: bounded improvement against an external anchor.

The ablations show the same effect from the other side. On a small increment,
objectives that train on fixed texts lose the model's behavior altogether: SFT
collapses during training, and DPO collapses into handing conversations to a
human in the first turn (Section~\ref{sec:exp_objective}). OPSD trains on the same
data and stays stable, which we attribute to its target being the model's own
distribution.

These design choices also define the method's limits. The self-teacher remains
limited by what it can articulate from the privileged information. A verified
reference repair narrows this gap, but an external model writes that repair in
our implementation, and external models also drive inspection and attribution, so
the loop is not self-contained (Section~\ref{sec:loop_scope}).
Grounding comes from the verifier, not the proposer. Repairs are also local to
a single decision. Failures that reflect missing knowledge rather than an
incorrect decision are better addressed through component repair (for example,
a Skill revision), new cold-start data, or human-written targets.

\subsection{Limitations}
The simulator is a modeled world, not a digital twin. Its state model, user
behavior, and rules can be wrong, and we have not yet measured its fidelity
against production in terms of user behavior, conversation length, tool paths,
and failure families. Work on L2 dialogue simulation compares generated and
human dialogue distributions while accounting for speakers' language
backgrounds~\citep{gao-etal-2025-llms}. Adapting such comparisons to
customer-service interaction patterns is a direction for validating our
simulator.

The confirmed-bad rate depends on screening recall and is
therefore a conservative observed rate, not a calibrated population estimate.
Its calibration against multilingual human annotation remains to be reported.
The online results have few satisfaction ratings in the early periods, while
the offline metrics are reported without confidence intervals. The serving cost does not include the research, storage, network
traffic, or additional operations staff of the in-house model, so it understates
the full cost of ownership. Harness changes between versions overlap with the model updates, so
the trend across versions cannot be assigned to the model alone, although the
comparison with the incumbent within a period is not affected. The Bad Case
counts behind the loss ablation come from separate
review batches that were not uniformly finalized, and the ablation changes
several factors at once in some comparisons, so it supports a choice of
configuration rather than the effect of each factor. Attribution is itself a
model-assisted judgment, so its errors
propagate into training. Controlled replay and an explicit unknown category
mitigate this risk, but attribution accuracy is not reported here.

Evaluation is also bound to a single policy version. The simulation test is
compiled from the latest Skill/SOP and then frozen; offline inspection evaluates
the resulting simulation traces. A candidate is therefore measured against the
policy currently in force, not against earlier versions. Although training
mixes failures from several SOP versions, we do not yet test whether a candidate
still follows an earlier SOP correctly when given that SOP as input. The benefit
of mixing versions is therefore argued rather than measured. How well a deployed
version follows a revision made after its training, and how long a revision takes
to reach a validated update, have not yet been quantified. Since the
simulation test is already compiled automatically from a given SOP, a natural
next step is to compile tests for several SOP versions and evaluate each
candidate across them. We plan this cross-version simulation test as future work.

\subsection{Risks and safeguards}
\begin{itemize}[itemsep=2pt]
  \item \textbf{Evaluator gaming.} Repeated optimization against the same judge
  or the same exposed cases can raise scores without improving service. Cases
  used to construct incremental tasks are excluded from evaluation, judgments
  are evidence-blinded, and boundary and exploration samples continue to identify
  new failure families.
  \item \textbf{Privileged leakage.} If the diagnosis were available to the
  student or during evaluation, measured gains would overstate what the model
  learned. By construction, the diagnosis is confined to the teacher context.
  \item \textbf{Forgetting.} Training on new failures can erase behavior that
  earlier versions handled well. Static test sets detect such regressions only
  for the current SOP, making cross-version evaluation a priority for future
  work.
  \item \textbf{Privacy and authority.} Production traces contain personal and
  order information. Collection, linking of human conversations, and
  publication require explicit authorization, de-identification, least-privilege
  access, and audit logs; credentials and customer identifiers never enter
  prompts or reports. Promotion is a consequential action approved by an
  accountable owner.
\end{itemize}

\subsection{Toward a more closed loop}
\care is deliberately human-on-the-loop and relies on an external frontier model
for retrospection and construction. Several extensions would further close the
improvement loop. Having the current model propose its own repairs under the same
verifier would remove the external proposer from Phase~II; inspection, attribution,
and verification would remain external, because the grounding depends on their
independence from the model being improved.
Allowing the student to continue the conversation in the simulator after the
repaired decision would extend on-policy learning from single decisions to
multi-step recovery. Compiling simulation tests for several Skill/SOP versions
would allow each candidate to be evaluated across policy revisions rather than
only against the latest one. Rubric-based reinforcement
learning~\citep{rubrics} could complement self-distillation for failures whose
correction is easier to verify than to articulate, provided that hard business
constraints cannot be offset by soft rubric scores. Finally, co-evolving the
evaluator with the agent~\citep{rqgm}, under the same grounding discipline,
would let the retrospection stage improve alongside the model it inspects.

%% file: Technical_Report/sections/10_conclusion.tex
\section{Conclusion}
\label{sec:conclusion}
We presented \care, a production framework for recursive self-improvement in
customer service. It aims to replace a frontier incumbent with an in-house
mid-sized model that learns from its own experience. A one-time cold
start on governed production traces made a 31B open-weight model competent but
left it with more confirmed failures than the incumbent. Five recursive
iterations then reduced its inspection Bad Case rate below the incumbent's and
brought online user satisfaction to parity with the incumbent. On small
increments of repairs, training on fixed target texts collapsed, whereas
self-distillation remained stable, which makes it the learning signal that the
recursive phase depends on.

In each iteration, the model serves production traffic and its own failures
provide the next learning signal.
A 24/7 online inspection process establishes failures from reviewed evidence,
and attribution separates model errors from surrounding-system defects.
Model-attributable failures yield verified repairs for on-policy
self-distillation: the same model serves as teacher, with attribution and repair
available as privileged information withheld from the student.

To support changing business requirements, Skills/SOPs enter through the model's
input. Training mixes failures from several SOP versions, each judged against
the policy in force when it occurred, to teach the model to follow the policy it
is given. Evaluation uses a simulation test compiled from the latest SOP and
then frozen, followed by offline inspection of its traces. The combined score
must match or exceed that of the deployed version before an online comparison
with the incumbent and promotion under human approval.

\care thus makes continuing model evolution a practical production process,
designed to keep pace with business changes and shorten the path from a revised
policy to a validated update.

%% file: Technical_Report/sections/appendix.tex
\section{Cold-Start Details}
\label{app:data}

\subsection{Evolution of the data recipe}
\label{app:data_history}
The data engine of Section~\ref{sec:cold_start} is the result of several
revisions, listed in Table~\ref{tab:data_history}. Early recipes concatenated
complete human conversations (D2) or inserted human turns into agent
trajectories (D3--D6). Insertion aimed to improve tone but risked introducing
steps that the agent never took and facts that only a human operator could
know, which motivated protecting business actions during insertion (D4), the
precursor of Eq.~\eqref{eq:protect}, and, eventually, the current rule of rewriting only existing turns
without adding messages. Rebalancing (D5) anticipated the hierarchical
allocation of Section~\ref{sec:allocation}, and the language repair of D7
anticipated the language admission checks. The current engine replaces
insertion with state-aligned refinement and moves all screening before
allocation.

\input{Technical_Report/tables/data_versions}

\subsection{Semantic screening states}
Table~\ref{tab:screening_states} defines the three verdicts of the semantic
screening step. Screening must cover the full structurally qualified pool with
one-to-one identity matching; missing results, duplicate identities, and
inconsistent assignments are treated as failures rather than passes. Disabling
screening yields a structurally qualified pool, not a pool that has passed
semantic screening.

\begin{table}[H]
\caption{Semantic screening verdicts and their downstream effect.}
\label{tab:screening_states}
\centering
\small
\begin{tabularx}{\linewidth}{@{}l Y Y@{}}
\toprule
Verdict & Meaning & Effect \\
\midrule
\texttt{no\_problem} & No supported execution defect; the decision does not depend on missing runtime evidence & Admitted to allocation \\
\texttt{problem} & Visible evidence supports a specific execution defect & Quarantined \\
\texttt{inconclusive} & No defect found, but the decision depends on missing decisive runtime evidence & Quarantined separately \\
\bottomrule
\end{tabularx}
\end{table}

\subsection{Training configuration}
\label{app:cold_config}
Table~\ref{tab:cold_config} lists the configuration used to train V0 on $D_0$
with the objective of Eq.~\eqref{eq:cold}. The maximum sequence length matches
the admission check of Section~\ref{sec:admission}: conversations longer than
16,384 tokens are excluded from $D_0$ rather than truncated.

\begin{table}[H]
\caption{Cold-start SFT configuration.}
\label{tab:cold_config}
\centering
\small
\begin{tabularx}{\linewidth}{@{}>{\raggedright\arraybackslash}p{0.27\linewidth}Y@{}}
\toprule
Item & Setting \\
\midrule
Base model & Gemma-4-31B-IT \\
Objective & masked negative log-likelihood on assistant tokens (Eq.~\ref{eq:cold}) \\
Maximum sequence length & 16,384 tokens \\
Epochs and steps & one epoch; 467 steps \\
Batch size & global batch 32; micro-batch 1 per GPU \\
Learning rate & peak $1\times10^{-5}$; warmup over 5\% of steps (23 steps), then cosine decay \\
Precision & BF16 \\
Hardware & 32 NVIDIA A100 80GB GPUs \\
\bottomrule
\end{tabularx}
\end{table}

\section{Simulation Environment Details}
\label{app:simulation}

\paragraph{Execution tiers.}
Simulation results are produced in three tiers with increasing scope
(Table~\ref{tab:sim_tiers}). An isolated execution error may be retried without
changing the experiment identity, and retries preserve attempt identity so that
failed executions are not silently dropped; global provider, version, or
aggregation failures stop the run.

\begin{table}[H]
\caption{Simulation execution tiers.}
\label{tab:sim_tiers}
\centering
\small
\begin{tabularx}{\linewidth}{@{}l Y Y@{}}
\toprule
Tier & Purpose & Supported conclusion \\
\midrule
Fixture smoke & Check assets, protocols, state transitions, and the offline loop with a substitute agent & Internal consistency of the environment \\
Live smoke & Exercise a small slice with the real model, routes, and tools & Integration evidence for that slice \\
Live end-to-end & Run the real chain on a frozen scenario set & Acceptance within the selected scope \\
\bottomrule
\end{tabularx}
\end{table}

\paragraph{Experimental controls.}
Controlled comparisons freeze the scenario set, capability versions, simulator
capabilities, route identity, user model, agent lifecycle, and judgment
contract. Version bindings, a compiled test plan, and a run lock retain these
identities, and detected drift stops execution instead of substituting a
runnable alternative with different semantics.

\paragraph{Evidence and privacy.}
An attempt is the basic unit that joins user generation, system actions, and
agent calls into one local trace. Persistent reports retain redacted
identities, statuses, and references; raw prompts, full requests and business
state, credentials, and sensitive customer identifiers are excluded.

\section{Inspection Details}
\label{app:inspection}

\paragraph{Inspection modes.}
The inspection system exposes several modes that share prompts and evidence
schemas but keep independent queues, denominators, and side effects
(Table~\ref{tab:inspection_modes}). This separation prevents a local sample from
becoming a population denominator and a diagnostic screen from being reported
as a confirmed-bad rate.

\begin{table}[H]
\caption{Inspection modes.}
\label{tab:inspection_modes}
\centering
\small
\begin{tabularx}{\linewidth}{@{}l Y Y@{}}
\toprule
Mode & Input and purpose & Boundary \\
\midrule
Online continuous & Mature production sessions, for ongoing discovery & Persistent source, ranked pool, admitted queue \\
Custom online & A selected experiment, market, window, or user segment & Independent of the main online queue \\
Offline check & A frozen report or export with a fixed test set & Frozen cases and evidence view \\
Screening only & Fast candidate diagnosis & No deep analysis or publication \\
Technical check & Tool-call, log, and infrastructure anomalies & Technical signal only; no semantic verdict \\
\bottomrule
\end{tabularx}
\end{table}

\paragraph{Stage contract.}
Each stage of Figure~\ref{fig:retrospection} produces a required artifact and
can stop a case; a later stage cannot manufacture evidence that an earlier one
lacked (Table~\ref{tab:inspection_stages}). Publication to the shared quality
dashboard is incremental by test-set version, requires a passing audit and
explicit confirmation, and is complete only after the written record has been
read back and matched against the intended payload.

\begin{table}[H]
\caption{Inspection stage contract.}
\label{tab:inspection_stages}
\centering
\small
\begin{tabularx}{\linewidth}{@{}l Y Y@{}}
\toprule
Stage & Required artifact & Stop condition \\
\midrule
Freeze & Source, case identities, evidence view & Identity or snapshot not fixed \\
Screen & Candidate score, evidence anchors, risk tier & Screening failure \\
Admit & Ranked pool and admitted queue & Capacity or strata unavailable \\
Deep analysis & Structured case record with claim ledger & Unsupported claims \\
Review & Independent verdict and correction history & Conflict, missing evidence, malformed output \\
Gate and audit & Terminal verdict, counts, denominator & Missing field, count mismatch, leakage \\
\bottomrule
\end{tabularx}
\end{table}

\paragraph{Evaluation dimensions.}
Business correctness covers routing, tool and action validity, process
progression, consistency between replies and actions, coverage of available
solutions, and appropriate escalation. Service quality covers language
consistency, emotion recognition, specific empathy, ownership, and the absence
of blaming or defensive language. Hard business violations are reported
separately from soft quality, and no aggregate allows pleasant language to
compensate for an invalid business action.

\subsection{Two inspected production cases}
\label{app:inspection_cases}
We follow two production conversations through the procedure of
Section~\ref{sec:inspection}: one that the inspection judged correct and one it
confirmed as a Bad Case. Both are refund inquiries from consumer traffic in
Saudi Arabia on 1 August 2026 (UTC), served by the in-house model during period
P1 (V0) under the same refund-consult Skill, and both were inspected in the
custom online mode of Table~\ref{tab:inspection_modes}. Dialogue is translated
from Arabic and condensed; merchants and orders are replaced by placeholders,
with \emph{Merchant~A} the merchant the customer means and \emph{Merchant~B}
the merchant on the order card the system suggests.

Two rules of the Skill decide both cases. When no order has been selected, the
agent calls a context resolver; if it returns no order ID, the only allowed
reply is \texttt{ORDER\_CONFIRMATION}, asking the customer to pick the order. If
the customer says the current order is not the one they mean, the agent must
ask for the new order and must not reuse the previous order's result.
Table~\ref{tab:case_summary} summarizes the two cases.

\begin{table}[H]
\caption{The two inspected cases at a glance.}
\label{tab:case_summary}
\centering
\small
\begin{tabularx}{\linewidth}{@{}l Y Y@{}}
\toprule
 & Case 1: expected behavior & Case 2: confirmed Bad Case \\
\midrule
Customer issue & Refund for an order cancelled because the courier was late & Charged, but the order was never confirmed \\
Retained window & 9 entries over 45\,s & 15 entries over 5.5\,min \\
Tool calls & 6: Skill load, context resolver & 8: Skill load, context resolver, order detail \\
Review & First review FAIL; reconciled; second review PASS as expected behavior, high confidence & First review PASS as confirmed Bad Case, high confidence \\
Gate and outcome & Upload not allowed; closed as no issue & Notification gate PASS; uploaded as a Bad Case \\
Inspection time & 26\,min & 18.5\,min \\
\bottomrule
\end{tabularx}
\end{table}

\paragraph{Case 1: a suspicious-looking conversation judged correct.}
Table~\ref{tab:case_good} lists the dialogue. The customer is frustrated, the
agent never states a refund status, and the order card it shows belongs to a
different merchant, all of which a judge relying on weak signals could read as
a failure. The inspection did not. In every turn after the Skill was loaded,
the resolver returned an empty order ID, so \texttt{ORDER\_CONFIRMATION} was the
only path the Skill allowed; the agent neither guessed the order nor claimed a
refund state. The card on G09 is the order picker's initial suggestion,
awaiting the customer's confirmation, and the review found that showing such a
suggestion does not by itself violate the picker's design; the retained window
ends before the customer confirms any order. The first review failed the analysis; after
reconciliation, the second review passed it as expected behavior with no
missing evidence. Under Eq.~\eqref{eq:feedback}, an expected verdict of this
kind becomes a negative example for the screener.

\begin{table}[H]
\caption{Case 1 dialogue (expected behavior). Replies the model generated
for G02, G05, and G06 were not delivered to the customer and are omitted.}
\label{tab:case_good}
\centering
\footnotesize
\begin{tabularx}{\linewidth}{@{}l l Y >{\raggedright\arraybackslash}p{0.27\linewidth}@{}}
\toprule
\# & Speaker & Message (translated, condensed) & Action and tools \\
\midrule
G01 & Agent & Welcome. Is your issue ``When will I get my money back?'' & Entry FAQ \\
G02--03 & User & No. \;/\; The Merchant~A order. & \\
G04 & Agent & Do you mean an order from Merchant~A? Please tell me the issue, e.g.\ a missing item, a wrong order, cold food, or a late delivery. & \texttt{INTENT\_GUIDE} \\
G05--07 & User & I cancelled it and the money has not been returned. \;/\; The driver was late. \;/\; It has been over an hour and the money has not been returned. & \\
G08 & Agent & Since the order was cancelled because the driver was late, please confirm the order in the list and I will check its refund status. & Skill loaded; resolver returns no order ID; \texttt{ORDER\_CONFIRMATION} \\
G09 & Card & Suggested order: Merchant~B, delivered, 50.40~SAR, awaiting confirmation. & \\
\bottomrule
\end{tabularx}
\end{table}

\paragraph{Case 2: a stale order kept after an explicit correction.}
Table~\ref{tab:case_bad} lists the dialogue. The customer names Merchant~A in
B05, yet in B06 the agent looks up and reports on the Merchant~B order, a
delivered order from three days earlier with no refund record. After the
explicit correction in B07, the reply B08 acknowledges in text that the
customer means a different order, but the confirmation card B09 still carries
the stale Merchant~B order. The customer confirms that card (B10--B12), and the
conversation ends in a transfer to a human agent (B13--B15) without the
intended order ever having been checked. The review confirmed a wrong-entity
defect, the stale order and card after the correction, with high confidence
and no missing evidence; the gate passed and the case was uploaded. The
first erroneous turn is B06. The resolver also returned the stale order ID in
B06, B08, and B13, so the card shown in B09 reflects runtime state as well as
the model's choice. Separating the two is the task of attribution
(Section~\ref{sec:attribution}), which decides whether such a case becomes
training data.

\begin{table}[H]
\caption{Case 2 dialogue (confirmed Bad Case). Rows marked $\times$ contain
the defect.}
\label{tab:case_bad}
\centering
\footnotesize
\begin{tabularx}{\linewidth}{@{}l l Y >{\raggedright\arraybackslash}p{0.27\linewidth}@{}}
\toprule
\# & Speaker & Message (translated, condensed) & Action and tools \\
\midrule
B01 & Agent & Welcome. Is your issue ``Why did the payment fail?'' & Entry FAQ \\
B02 & User & I placed an order; the money was taken, but the order was not confirmed. & \\
B03 & Agent & Please select the order you are asking about, and I will check its payment or refund status. & Skill loaded; resolver returns no order ID; \texttt{ORDER\_CONFIRMATION} \\
B04 & Card & Suggested order: Merchant~B, delivered, 34.68~SAR, awaiting confirmation. & \\
B05 & User & I ordered from Merchant~A and you took the money. Why was the order not confirmed? & \\
B06\,$\times$ & Agent & I checked the Merchant~B order: it was completed and delivered, and has no refund request. If this is not the right order, please give the order number. & Resolver returns the Merchant~B order; order detail: completed on 29 July, no after-sale record; \texttt{GUIDE} \\
B07 & User & My order is not Merchant~B; my new order is desserts from Merchant~A. & \\
B08 & Agent & The order shown to me is from Merchant~B, but you mean the new dessert order from Merchant~A. Please select or confirm it. & Resolver returns the Merchant~B order again; \texttt{ORDER\_CONFIRMATION} \\
B09\,$\times$ & Card & Order card: the same Merchant~B order, 34.68~SAR, marked confirmed. & \\
B10--12 & User & Clicks confirm on the card; ``Yes.''; the system records the Merchant~B order as confirmed. & \\
B13 & Agent & Sorry for the confusion. Since you confirmed this completed order, please contact a specialist, who can check the new Merchant~A order directly. & Resolver and order detail on the Merchant~B order again; \texttt{TRANSFER\_TO\_HUMAN} \\
B14--15 & Agent, User & Contact-a-specialist entry; the customer transfers to a human agent. & \\
\bottomrule
\end{tabularx}
\end{table}

\section{Repair Construction Details}
\label{app:construction}

\paragraph{Sources and identity.}
Each confirmed failure is frozen together with its inspection record and
diagnosis and four trace streams: the requests sent to the model, the agent's
outputs and tool calls, state persistence, and the content rendered to the
user. Cases are keyed by both case and conversation identity, and file
manifests are hashed. The persistence and rendering streams support attribution
but never enter the student input, and neither do other cases from the same
conversation, neighboring requests, or a policy revision newer than the
incident.

\paragraph{Anchors.}
The first erroneous decision is anchored by the request identity together with
the exact logged output and its content hash. Ordinal positions (``the third
assistant turn''), the first occurrence of a repeated request identity, and
earlier repair artifacts are not accepted as anchors, although earlier
artifacts may help to locate a candidate that is then re-verified against the
frozen sources.

\paragraph{Verification axes.}
Table~\ref{tab:verifier_axes} lists the axes of the independent verifier. A
tuple is admitted only if every axis takes its passing value; ineligible,
unrepaired, undetermined, malformed, and pending outcomes are all excluded.

\begin{table}[H]
\caption{Axes of the independent repair verifier.}
\label{tab:verifier_axes}
\centering
\small
\begin{tabularx}{\linewidth}{@{}l Y l@{}}
\toprule
Axis & Question & Passing value \\
\midrule
Disposition & Is the case repaired by the proposed decision? & repaired \\
Eligibility & May the pair be used for training at all? & eligible \\
Fault relation & Is the anchored output the model's own fault? & model fault \\
Symptom & Is the inspected symptom corrected? & corrected \\
Boundary & Is the change confined to the anchored decision? & corrected \\
Policy & Does the repair follow the incident-time SOP? & compliant \\
Format & Does it respect the Skill, tool, and field contract? & compliant \\
\bottomrule
\end{tabularx}
\end{table}


\paragraph{Training admission.}
A verified tuple is a candidate, not yet training data. Before the OPSD runs of
Section~\ref{sec:experiments}, every selected tuple passed the following checks:
(i) \emph{request fidelity}: the student messages equal the frozen request and
its content hash matches, with tools retained and the prompt ending where the
next assistant turn begins; (ii) \emph{privilege isolation}: the teacher text is
non-empty and absent from the student request, verified by reading every tuple
back through the actual training dataset and collator; (iii) \emph{visual
evidence}: every image resolves to a frozen file whose hash, size, and decoding
are checked, and tuples with more than four images are excluded whole; (iv)
\emph{length}: student and teacher contexts are encoded with the target
tokenizer, chat template, tool schemas, and image processor, and must fit the
32K context together with a 2,048-token response budget, with truncation
treated as an error; and (v) a one-step run on the full training layout,
covering rollout, teacher scoring, update, checkpointing, export, and
independent readback, after which training restarts from the initial
checkpoint. Replaying the incident model at $x$, to test whether the original
error reproduces, is a separate check.

\section{OPSD Training Configuration}
\label{app:opsd_config}
Table~\ref{tab:opsd_config} lists the OPSD configuration of the final model
(configuration 4 of Table~\ref{tab:ablate_loss}). Configurations 1, 2, and 3 of
Section~\ref{sec:exp_loss} share every setting except the objective row, which
Table~\ref{tab:ablate_loss} gives for each of them. SFT and DPO in
Section~\ref{sec:exp_objective} use the same initialization, repair tuples, and
training budget.

\begin{table}[H]
\caption{OPSD training configuration of the final model.}
\label{tab:opsd_config}
\centering
\small
\begin{tabularx}{\linewidth}{@{}>{\raggedright\arraybackslash}p{0.27\linewidth}Y@{}}
\toprule
Item & Setting \\
\midrule
Student initialization & V4, the previous version; full-parameter update, vision tower trainable \\
Teacher & V4, frozen for the whole run, not a moving average; scores the student's rollout token by token; targets detached and kept in FP32 \\
Privileged information & appended to the teacher's system message; never in the student context \\
Training set & 280 repair tuples after micro-scenario capping (Section~\ref{sec:replay}); 281 candidates, one excluded for exceeding four images per tuple \\
Objective & forward KL partial sum over the teacher's top 32 tokens, floored at zero per token (Eq.~\ref{eq:opsd_forward}); backpropagated directly; loss coefficient 1.0; mean over all generated tokens of the batch (Eq.~\ref{eq:opsd_batch}) \\
Not used & task reward, policy gradient, tail cell, renormalization within the top-$K$ set, probability floor or smoothing, auxiliary SFT or DPO loss \\
Student sampling & one sample per tuple and update; temperature 1.0, top-$p$ 1.0, no top-$k$; single turn; at most 2,048 response tokens \\
Context limits & prompt $\le$ 30,719 tokens; prompt plus response $\le$ 32,768 tokens for both student and teacher contexts; no truncation \\
Optimizer & AdamW, $\beta=(0.9,0.999)$, weight decay 0.01, gradient-norm clipping 1.0, fresh state \\
Learning rate & peak $5\times10^{-6}$, linear warmup over 5\% of steps (10 steps), cosine decay \\
Batch and schedule & global batch 4, three epochs of 70 steps; 210 steps; the third epoch is the final model \\
Precision and parallelism & BF16; 16 GPUs, sequence parallel 8 $\times$ data parallel 2, one sequence per micro-batch \\
Checkpoints & end of every epoch; each exported and independently read back \\
\bottomrule
\end{tabularx}
\end{table}

%% file: Technical_Report/tables/data_versions.tex
\begin{table}[H]
\caption{Historical revisions of the cold-start data recipe.}
\label{tab:data_history}
\centering
\small
\begin{tabularx}{\linewidth}{@{}l Y@{}}
\toprule
Revision & Transformation \\
\midrule
D1 Baseline & Production incumbent trajectories, no human data \\
D2 Concat & Append complete human-support conversations \\
D3 Style & Filter, rewrite, and insert human turns \\
D4 Protect & Preserve business actions and metadata during insertion \\
D5 Rebalance & Adjust the region and scenario distribution \\
D6 EvalPatch & Target inspected failures while preserving structure \\
D7 LangFix & Repair assistant language only \\
\bottomrule
\end{tabularx}
\end{table}